\documentclass[10pt]{article} 
\usepackage[preprint]{tmlr}

\usepackage{amsmath,amsfonts,bm}

\def\eqref#1{(\ref{#1})}

\def\1{\bm{1}}

\DeclareMathAlphabet{\mathsfit}{\encodingdefault}{\sfdefault}{m}{sl}
\SetMathAlphabet{\mathsfit}{bold}{\encodingdefault}{\sfdefault}{bx}{n}

\usepackage{hyperref}
\usepackage{url}

\title{On the Convergence of Vanilla Adam under Heavy-Tailed Noise and Index Sampling}

\author{\name Kanata Oowada \email kanata7527@gmail.com \\
      \addr Meiji University
      \AND  
      \name Hideaki Iiduka \email iiduka@cs.meiji.ac.jp\\
      \addr Meiji University}

\def\month{MM}  
\def\year{YYYY} 
\def\openreview{\url{https://openreview.net/forum?id=XXXX}} 

\usepackage{bm}
\usepackage{algorithm, algorithmic, comment, amsthm, graphicx,mathtools,empheq, amsmath,caption,wrapfig, bbm,amssymb,makecell}

\newtheorem*{assumption*}{Assumption}
\newtheorem{lemma}{Lemma}
\newtheorem{proposition}{Proposition}
\newtheorem{corollary}{Corollary}

\newtheorem{theorem}{Theorem}
\newtheorem{condition}{Condition}

\DeclareMathOperator*{\diag}{diag}

\usepackage{enumitem,hyperref}
\usepackage{etoolbox}
\AtBeginEnvironment{algorithmic}{\let\AND\relax}

\begin{document}

\maketitle

\begin{abstract}
Understanding how stochastic gradient noise affects the convergence of Adaptive Moment Estimation (Adam) is an important problem in stochastic optimization, particularly when the noise exhibits heavy-tailed behavior. In this paper, we investigate the convergence of vanilla Adam in both the expected risk and the empirical risk minimization, focusing on the effects of heavy-tailed population noise and index-sampling noise. For the empirical risk minimization, we establish convergence guarantees under H\"older smoothness, showing that heavy-tailed population noise can affect the convergence bounds through dataset-dependent constants that may increase with the dataset size. For the expected risk minimization, we firstly establish both asymptotic and non-asymptotic convergence guarantees for vanilla Adam under H\"older smoothness and bounded $\mathfrak{p}$-th moment assumptions, allowing for infinite-variance stochastic gradient noise. In particular, we derive explicit convergence rates in terms of the expected squared gradient norm and show that the convergence rates deteriorate as $\mathfrak{p}$ decreases. Numerical experiments further support these theoretical findings and provide empirical evidence for the relevance of the sufficient conditions underlying our convergence analysis.
\end{abstract}

\section{Introduction}
Stochastic gradient descent (SGD) \citep{10.1214/aoms/1177729586} and its variants are fundamental optimization methods for large-scale machine learning. Among them, Adaptive Moment Estimation (Adam) \citep{kingma2015adam} has become one of the most widely used optimization algorithms in deep learning, with successful applications across a broad range of tasks \citep{NIPS2017_3f5ee243,jacob2019bert,ICLR2024_292b9cfd}. Adam combines momentum-based updates with adaptive, coordinate-wise learning rates determined by estimates of the first and second moments of stochastic gradients. Its adaptive mechanism is closely related to earlier methods such as RMSProp \citep{rmsprop}, while its momentum component builds on ideas underlying classical methods such as the heavy-ball method \citep{polyak1964}. Despite its practical success, understanding the convergence behavior of Adam remains an important topic in stochastic optimization. \citet{j.2018on} demonstrated that Adam can fail to converge even in simple convex optimization problems and proposed AMSGrad to address this issue. Subsequent studies have established convergence guarantees under various assumptions on the objective function, stochastic gradients, and algorithmic parameters.

Nevertheless, understanding how stochastic gradient noise, particularly heavy-tailed noise, affects the convergence of Adam remains an important challenge. Although bounded variance is a common assumption \citep{zaheer2018adam_convanal,wang2023closinggap,jin2026adambeatsgd} in stochastic optimization, stochastic gradients arising in deep learning have been reported to exhibit heavy-tailed behavior \citep{simsekli2019hevy,garg2021heavy,battash2024noisy}. Such noise can be modeled by assuming only a bounded $\mathfrak{p}$-th moment for some $\mathfrak{p}\in(1,2]$, which allows the second moment to be infinite when $\mathfrak{p}<2$. While gradient clipping is known to be effective in mitigating the effects of heavy-tailed noise in SGD \citep{nguyen2023clip,shestakov2026gap_expempi}, we investigate whether vanilla Adam can converge under such noise without gradient clipping.

A key distinction arises between the expected risk and the empirical risk minimization when considering heavy-tailed stochastic gradient noise. In the expected risk minimization, stochastic gradients are computed from data sampled directly from the underlying population distribution $\mathcal{D}$, which may give rise to infinite-variance noise. In contrast, the empirical risk minimization is based on a finite dataset sampled from $\mathcal{D}$, and stochastic gradients are subsequently computed by sampling indices from the fixed dataset. Consequently, while the noise induced by index sampling has finite moments for any fixed dataset and evaluation point, the corresponding moment bounds may depend on the dataset and grow as its size increases \citep{brown2021datahev,yang2025datahev,shestakov2026gap_expempi} (see Section \ref{subsec:prelimi_asm_algo}). Thus, although index sampling over a fixed finite dataset cannot itself produce infinite-variance noise, the heavy-tailed nature of the underlying population distribution may still affect the convergence behavior of Adam through dataset-dependent quantities. This distinction motivates analyzing the two settings separately and clarifying the relationship between their stochastic gradient noise assumptions and convergence guarantees.

\subsection{Related Works and Contributions}
\begin{table}[tpb]
\footnotesize
\centering
\renewcommand{\arraystretch}{1.8}
\begin{tabular}{|l|c|c|c|c|c|c|}
\hline
Reference & \makecell{\textbf{Vanilla}\\\textbf{Adam}} &  \textbf{Noise Asm.}& \textbf{Smoothness}& \textbf{LR} & \textbf{Beta} & \textbf{Rate} \\ \hline
\cite{defossez2022simpleconv} & No & A.s. bounded SG & Lipschitz & $\alpha=\frac{1}{\sqrt{\tilde{K}}}$ & \makecell{$\beta_1 \in [0, 1),$\\$\beta_2=1-\frac{1}{\tilde{K}}$} & $\mathcal{O}_{\tilde{K}} \left( \frac{\log \tilde{K}}{\sqrt{\tilde{K}}} \right)$ \\ \hline
\citet{Liang2025convadam} & No & A.s. bounded SG & Lipschitz & $\alpha_k = \frac{1}{\sqrt{k}}$ & \makecell{$\beta_1 \in [0, \beta)$,\\$\beta_{2,k}=1-\frac{1}{k}$} & $O \left( \frac{\log K}{\sqrt{K}} \right)$ \\ \hline
\cite{zou2019assuficient} & No & \makecell{Bounded 2nd\\ moment of SG} & Lipschitz & $\alpha_k=\frac{1}{\sqrt{k}}$ & \makecell{$\beta_1 \in [0, \beta)$,\\$\beta_{2,k}=1-\frac{1}{k}$} & $O \left( \frac{\log K}{\sqrt{K}} \right)$ \\ \hline
\cite{wang2023closinggap} & No & Bounded variance & Lipschitz & $\alpha=\frac{1}{\sqrt{\tilde{K}}}$ & \makecell{$\beta_1 =\sqrt{\beta_2},$\\$\beta_2=1-\frac{1}{\tilde{K}}$} & $\mathcal{O}_{\tilde{K}} \left( \frac{\log \tilde{K}}{\sqrt[4]{\tilde{K}}} \right)$ \\ \hline
\cite{jin2025comp} & No & ABC condition & Lipschitz & $\alpha_k = \frac{1}{\sqrt{k}}$ & \makecell{$\beta_1 \in [0, 1)$,\\$\beta_{2,k}=1-\frac{1}{k}$} & $O\left( \frac{\log^2 K}{\sqrt{K}} \right)$ \\ \hline
Theorem \ref{thm:holder_bcg_conv} & Yes & A.s. bounded SG & H\"older & $\alpha_k = \frac{1}{k^{1/(1+\mathfrak{h})}}$ & \makecell{$\beta_{1,k} \to 0$,\\$\beta_2=1-1/k^s$} &$O \left( \frac{\log K}{K^{\mathfrak{h}/(1+\mathfrak{h})}} \right)$ \\ \hline
\citet{liu2026adamconv} & Yes & \makecell{Bounded\\ $\mathfrak{p}$-th moment} & Well-behaved & $\alpha_k \sim \text{Exp}(1)$ & \makecell{$\beta_1 =1-\frac{1}{\tilde{K}^{u}}$,\\$\beta_{2}=1-\frac{1}{\tilde{K}^{u}}$} & $\mathcal{O}_{\tilde{K}} \left( \frac{1}{\tilde{K}^{v}}\right)$ \\ \hline
Theorem \ref{thm:heavy_conv} & Yes & \makecell{Bounded\\ $\mathfrak{p}$-th moment} & H\"older & $\alpha_k = \frac{1}{k^{1/(1+\mathfrak{h})}}$ & \makecell{$\beta_{1,k} \to 0$,\\$\beta_{2,k}=1-1/k^{s'}$} &$O\left(\left(\frac{\log K}{K^{\mathfrak{h}/(1+\mathfrak{h})}}\right)^{w}\right)$ \\ \hline
\end{tabular}
\caption{Comparison of existing convergence results with our theoretical guarantees, where $s >\frac{\mathfrak{h}}{1+\mathfrak{h}},s'>\frac{2}{1+\mathfrak{h}}$, $u=\frac{8\mathfrak{p}}{19\mathfrak{p}-12}$, $v=\frac{2(3\mathfrak{p}-4)}{19\mathfrak{p}-12}$, and $w=\frac{\mathfrak{p}}{1+\mathfrak{p}}$. $\mathfrak{p}\in(1,2]$ is the tail index. $\mathfrak{h}\in(0,1]$ is the H\"older index. The Vanilla Adam column indicates whether bias correction is applied. Except for \citet{liu2026adamconv}, the convergence rates are with respect to the expected squared norm. $\mathcal{O}_{\tilde{K}}$ denotes the finite horizon rates. LR represents learning rate. The condition ``Well-behaved'' is defined by $f(\bm{x})-f(\bm{y})=\int_0^1\langle\nabla f(\bm{y}+t(\bm{x}-\bm{y})),\bm{x}-\bm{y}\rangle dt$ for any $\bm{x},\bm{y}\in\mathbb{R}^d$.}\label{tab_1}
\end{table}
\textbf{Convergence analysis of Adam.} Several studies have established convergence guarantees for Adam and its variants under Lipschitz smoothness, as summarized in Table \ref{tab_1}. Their assumptions on stochastic gradients differ in strength. \citet{defossez2022simpleconv} and \citet{Liang2025convadam} assume almost surely bounded stochastic gradients, which is stronger than the bounded second moment condition employed by \citet{zou2019assuficient}. \citet{wang2023closinggap} considers the weaker bounded variance condition, while \citet{jin2025comp} employs the more general ABC condition, which allows the second moment of stochastic gradients to depend on the objective-function gap and full gradient norm. Although these assumptions differ in strength, they all require finite second-moment control and therefore do not directly accommodate infinite-variance stochastic gradient noise. The $\mathcal{O}_{\tilde{K}}$ rates reported in Table \ref{tab_1} are finite horizon convergence rates, some of which are obtained using parameters that depend on the prescribed iteration horizon $\tilde{K}$. Regarding the asymptotic big-O rates, $O(\log K/\sqrt{K})$ provides a tighter upper bound than $O(\log^2 K/\sqrt{K})$. However, the latter rate, established by \citet{jin2025comp}, was obtained under the weaker ABC condition on stochastic gradients. Our rate $O(\log K/K^{\mathfrak{h}/(1+\mathfrak{h})})$ in Corollary \ref{cor:holder_bcg_convrate} recovers $O(\log K/\sqrt{K})$ when $\mathfrak{h}=1$, corresponding to Lipschitz smoothness, while extending the convergence analysis to the broader class of H\"older-smooth objectives.

\textbf{Adam under heavy-tailed noise.} \citet{liu2026adamconv} analyzed vanilla Adam under heavy-tailed noise ($\mathfrak{p}\in(4/3,2)$) using randomly scaled learning rates and horizon-dependent momentum parameters. For nonsmooth nonconvex optimization, they established a finite-horizon convergence rate of $\mathcal{O}_{\tilde{K}}(\tilde{K}^{-\frac{2(3\mathfrak{p}-4)}{19\mathfrak{p}-12}})$ in terms of $(\lambda,\varepsilon)$-$L_1$-stationarity, a generalized stationarity criterion. Under the additional assumption of H\"older smoothness, their result can be translated into a rate of $\mathcal{O}_{\tilde{K}}(\tilde{K}^{-\frac{2\mathfrak{h}(3\mathfrak{p}-4)}{19\mathfrak{p}-12}})$ in terms of the expected squared gradient norm. In contrast, we establish an asymptotic convergence rate of $O(\frac{\log^{\mathfrak{p}} K}{K^{\mathfrak{hp}/\{(1+\mathfrak{h})(1+\mathfrak{p})\}}})$ under H\"older smoothness and heavy-tailed noise. This result relies on the condition $1+\mathfrak{h}\leq\mathfrak{p}$ and thus does not guarantee convergence under infinite-variance noise when $\mathfrak{h}=1$, corresponding to Lipschitz smoothness. Even in this setting, our empirical-risk analysis demonstrates that the effects of heavy-tailed population noise can manifest through dataset-dependent constants. On the other hand, since $\frac{\mathfrak{p}\mathfrak{h}}{(1+\mathfrak{p})(1+\mathfrak{h})}>\frac{2\mathfrak{h}(3\mathfrak{p}-4)}{19\mathfrak{p}-12}$ holds in the parameter regime where the assumptions of both analyses are satisfied, our rate is faster than the rate implied by their result under H\"older smoothness, even when our result is expressed as a finite-horizon rate. Nevertheless, their analysis addresses the more general nonsmooth setting. 

\textbf{Contributions.} Our main contributions are summarized as follows. The code used for the numerical experiments is available at \url{https://github.com/KanataOowada/adam_preprint_2026}.
\begin{enumerate}
    \item Convergence analysis under index sampling. We establish a convergence of $O(\frac{\log K}{K^{\mathfrak{h}/(1+\mathfrak{h})}})$ for vanilla Adam in the empirical risk minimization under H\"older smoothness and almost surely bounded stochastic gradients with respect to index sampling. The bounds in Theorems \ref{thm:holder_bcg_conv} and \ref{thm:bvariance_conv} reveal that the convergence guarantees may deteriorate as the dataset size $N$ increases through dataset-dependent constants.
    \item Convergence analysis under heavy-tailed noise. We establish asymptotic and non-asymptotic convergence guarantees for vanilla Adam in the expected risk minimization under H\"older smoothness and bounded $\mathfrak{p}$-th moment assumptions. In particular, we obtain a convergence rate of $O(\frac{\log^{\mathfrak{p}} K}{K^{\mathfrak{hp}/\{(1+\mathfrak{h})(1+\mathfrak{p})\}}})$ for the expected squared gradient norm, which, to the best of our knowledge, is the first such result for vanilla Adam under heavy-tailed noise. Corollary \ref{cor:heavy_convrate} further shows that vanilla Adam converges under heavy-tailed noise, although the convergence rate deteriorates as $\mathfrak{p}$ decreases. The numerical experiments in Section \ref{sec:num_heavy} further support this finding and provide empirical evidence for the relevance of the sufficient conditions underlying our convergence analysis.
\end{enumerate} 

\section{Preliminary and Heavy-Tailed Noise}\label{subsec:prelimi_asm_algo}
For $\bm{x},\bm{y} \in \mathbb{R}^d$, we define $\langle \bm{x},\bm{y} \rangle \coloneq \bm{x}^\top \bm{y}$ and $\| \bm{x}\| \coloneq \sqrt{\langle \bm{x}, \bm{x} \rangle}$. Let $\mathbb{N}$ be the set of positive integers, $\mathbb{N}_n\coloneq\{1,\ldots,n\}$, and let $C^1(\mathbb{R}^d)$ be the set of continuously differentiable functions $f:\mathbb{R}^d\to\mathbb{R}$. For square matrices $X, Y \in \mathbb{R}^{d \times d}$, we write $X \preceq Y$ if $Y-X$ is positive semidefinite; $X \odot Y$ denotes the Hadamard product. For a distribution $\mathcal{D}$ on a measurable space $(\Omega,\mathcal{F})$ and a measurable function $f:\mathbb{R}^d\times \Omega\to\mathbb{R}$, we define $\mathbb{E}_{\xi}[f(x;\xi)]\coloneq \mathbb{E}_{\xi\sim\mathcal{D}}[f(x;\xi)]\coloneq\int_{\Omega}f(x;z)\mathcal{D}(dz)$ and denote the total expectation for $k$ steps as $\mathbb{E}\coloneq\mathbb{E}_{(\xi_1,\ldots,\xi_k)\sim\mathcal{D}^{\otimes k}}=\mathbb{E}_{\xi_1}\ldots\mathbb{E}_{\xi_k}$. Let $L^1(\Omega)$ be the set of measurable functions on $(\Omega,\mathcal{F})$ satisfying $\int_{\Omega}|f|d\mathcal{D}<\infty$. For sequences $(x_k)_{k\in\mathbb{N}}$ and $(y_k)_{k\in\mathbb{N}}\subset[0,\infty)$, denote $y_k=o(x_k)$ if $\lim_{k\to\infty}y_k/x_k=0$, i.e., $\forall c> 0, \exists k_0,\forall k\in\mathbb{N}:k_0\leq k\implies y_k\leq cx_k$; denote $y_k=O(x_k)$ if $\exists c> 0, \exists k_0,\forall k\in\mathbb{N}:k_0\leq k\implies y_k\leq cx_k$. Here, we introduce $\mathcal{O}$ notation for finite horizon rates: denote $y_k=\mathcal{O}_{\tilde{k}}(x_{\tilde{k}})$ if $\exists c>0,\exists k_0,\forall k\in\mathbb{N}:k_0\vee \tilde{k}\leq k\implies y_k\leq c x_{\tilde{k}}$. A relation $y_k=o(x_k)\implies y_k=O(x_k)\implies y_k=\mathcal{O}_{\tilde{k}}(x_{\tilde{k}})$ holds.

\begin{algorithm}[htpb] 
\caption{Vanilla Adam \citep{kingma2015adam}} 
\label{algo:adam} 
\begin{algorithmic}
\REQUIRE
$\bm{x}_{1} \in\mathbb{R}^d$ (initial point), $\alpha_k \in \mathbb{R}$ (learning rate), $\beta_{1,k}, \beta_{2,k} \in [0, 1)$ (momentum parameters), $\epsilon > 0$ (stabilization constant).
\ENSURE
$\bm{x}_{K}$
\FOR{$k = 1, \ldots, K$}
\STATE
$\bm{g}(\bm{x}_k;\xi_k)
\gets \nabla F(\bm{x}_k;\xi_k)$
\STATE
$\bm{m}_k \gets (1 - \beta_{1,k})\bm{g}(\bm{x}_k;\xi_k) + \beta_{1,k}\bm{m}_{k-1}$
\STATE
$\hat{\bm{m}}_k \gets (1-\prod_{j=1}^k{\beta}_{1, j})^{-1}\bm{m}_k$
\STATE
$\bm{v}_k \gets (1-\beta_{2,k}) \bm{g}(\bm{x}_k;\xi_k) \odot \bm{g}(\bm{x}_k;\xi_k) + \beta_{2,k}\bm{v}_{k-1}$
\STATE
$\hat{\bm{v}}_k \gets (1-\prod_{j=1}^k{\beta}_{2, j})^{-1}\bm{v}_k$
\STATE
$H_k \gets \diag(\sqrt{\hat{v}_{k,i}} + \epsilon)$
\STATE
$\bm{x}_{k+1} \gets \bm{x}_k - \alpha_k H_k^{-1}\hat{\bm{m}}_k$
\ENDFOR
\end{algorithmic}
\end{algorithm}
We consider the expected risk minimization problem:
\begin{align*}
\text{minimize}\ f(\bm{x}) \coloneq  \mathbb{E}_{\xi\sim\mathcal{D}}[F(\bm{x};\xi)] \quad \text{subject to } \bm{x}\in\mathbb{R}^d,
\end{align*}
where $\mathcal{D}$ is a distribution on a measurable space $(\Omega,\mathcal{F})$, and we suppose that $F(\cdot;\xi) \in C^1(\mathbb{R}^d)$ $\mathcal{D}$-almost surely and $f<\infty$; for example, $F(\bm{x};\cdot)\in L^1(\Omega)$ ensures the latter condition. $f$ can be only accessed through the estimator $F$ because the true data distribution $\mathcal{D}$ is unknown. To solve the problem, \citet{kingma2015adam} proposed a groundbreaking algorithm named Adam as in Algorithm \ref{algo:adam}. $\xi_{k}$ is an i.i.d. sample from the distribution $\mathcal{D}$ at the $k$-th step, and we note that it is independent of $\{\bm{x}_t\}_{t=1}^k$ generated by Algorithm \ref{algo:adam}. For the convergence analysis of Algorithm \ref{algo:adam}, we consider following conditions:
\begin{condition}
Let $(\bm{x}_k)_{k\in \mathbb{N}} \subset \mathbb{R}^d$ be a sequence generated by Algorithm \ref{algo:adam}.
\begin{enumerate}[label=(C\arabic*)]
\item \label{asm:lowerbounded_objective} The objective function has a finite lower bound: $f_{\star}\coloneq \inf_{k\in\mathbb{N}}f(\bm{x}_k)\in\mathbb{R}$.
\item \label{asm:holdersmooth_gradient} Let $\mathfrak{h}\in(0,1]$. Then, the objective function is $L_{\mathfrak{h}}$-H\"{o}lder smooth: $f$ is differentiable and \[\exists L_{\mathfrak{h}},\forall \bm{x},\forall \bm{y}\in\mathbb{R}^d:\|\nabla f(\bm{x})-\nabla f(\bm{y})\|\leq L_{\mathfrak{h}}\|\bm{x}-\bm{y}\|^{\mathfrak{h}}.\]
\item \label{asm:bounded_gradient} The full gradient norm is bounded: \[\exists G>0,\forall k\in\mathbb{N}:\|\nabla f(\bm{x}_k)\|\leq G.\]
\item \label{asm:unbiasedestimator_sg} The stochastic gradient is an unbiased estimator of the full gradient with respect to $\mathcal
{D}$: \[\forall k\in\mathbb{N}:\mathbb{E}_{\xi_k\sim\mathcal{D}}[\bm{g}(\bm{x}_k;\xi_k)]=\nabla f(\bm{x}_k).\]
\item \label{asm:boundedpthmoment_sg} Let $\mathfrak{p}\in(1,2]$. Then, the $\mathfrak{p}$-th moment of the stochastic gradient is bounded: \[\exists \sigma\geq 0,\forall k\in\mathbb{N}:\mathbb{E}_{\xi_k\sim\mathcal{D}}[\|\bm{g}(\bm{x}_k;\xi_k)-\nabla f(\bm{x}_k)\|^{\mathfrak{p}}]\leq \sigma^{\mathfrak{p}}.\]
\end{enumerate}
\end{condition}
Taking the cross entropy loss as $F$ provides an example in which $f$ satisfies \ref{asm:lowerbounded_objective} with $f_{\star}=0$. Cross entropy loss is widely used for training DNNs, including large language models and models for image processing \citep{jacob2019bert,raffel2020tfive,tom2020lm_fewshot,kaplan2020scalinglow,alec2021learningtransfe}. Moreover, this condition is also satisfied by many other standard loss functions used in machine learning. Therefore, \ref{asm:lowerbounded_objective} is consistent with the properties of loss functions commonly used in practical learning problems and constitutes a natural condition in our setting. \ref{asm:holdersmooth_gradient} \citep{hoelder1882holdersmooth} generalizes Lipschitz smoothness, which is a standard condition in convergence analyses for nonconvex objectives \citep{zou2019assuficient,defossez2022simpleconv, wang2023closinggap}, and reduces to the Lipschitz-smooth case when $\mathfrak{h}=1$. Furthermore, \ref{asm:holdersmooth_gradient} immediately implies that $f\in C^1(\mathbb{R}^d)$. In many convergence analyses, an alternative form implied by \ref{asm:holdersmooth_gradient} \citep{yuyuan2023holder_descent} is useful: $\forall \bm{x},\forall \bm{y}\in\mathbb{R}^d:f(\bm{y})
\leq f(\bm{x})+\langle\nabla f(\bm{x}),\bm{y}-\bm{x}\rangle+\frac{L_{\mathfrak{h}}}{1+\mathfrak{h}}\|\bm{y}-\bm{x}\|^{1+\mathfrak{h}}$. 

\ref{asm:bounded_gradient} is weaker than the bounded stochastic gradient norms appearing in \citet{zaheer2018adam_convanal, defossez2022simpleconv, Liang2025convadam}. \ref{asm:bounded_gradient} can be reasonable in practice, as discussed below.
\begin{proposition}\label{prop:bound_grad}
If the objective function $f$ satisfies \ref{asm:lowerbounded_objective}, \ref{asm:holdersmooth_gradient}, and $f^{\star}\coloneq\sup_{k\in\mathbb{N}}f(\bm{x}_k)\in\mathbb{R}$, then, it holds that $\forall k\in\mathbb{N}:\|\nabla f(\bm{x}_k)\|\leq((1+\mathfrak{h})L_{\mathfrak{h}}^{\frac{1}{\mathfrak{h}}}(f^{\star}-f_{\star})/\mathfrak{h})^{\mathfrak{h}/(1+\mathfrak{h})}.$
\end{proposition}
\begin{figure}
\centering
\includegraphics[width=0.60\linewidth]{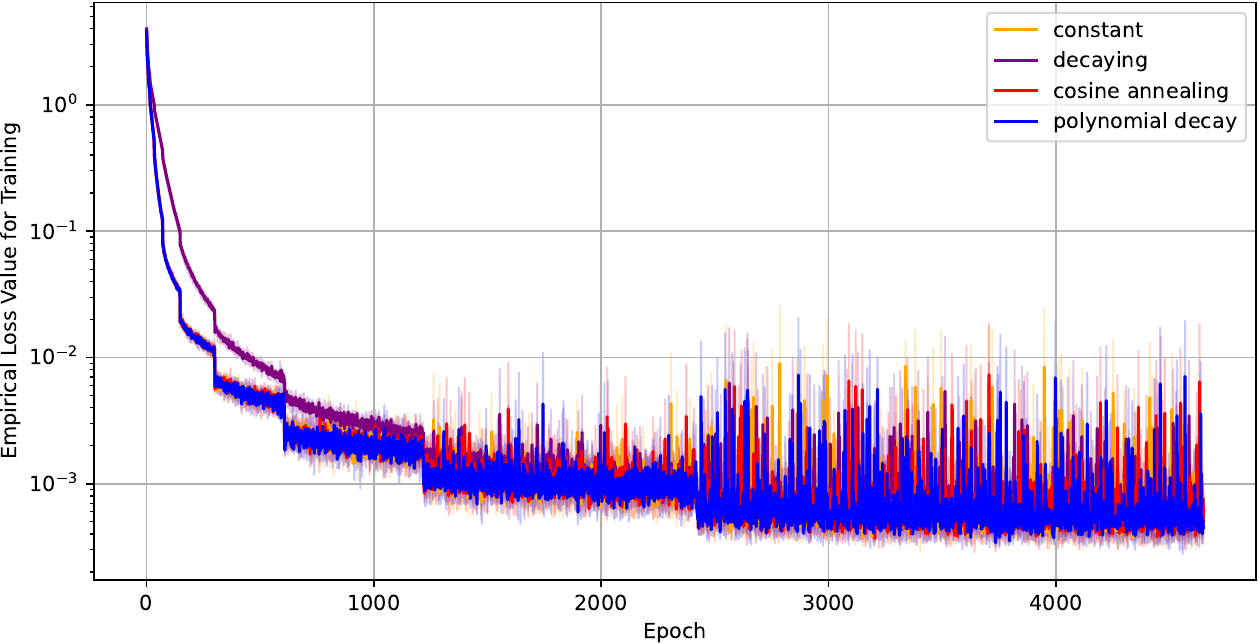}
\caption{Empirical loss versus number of epochs to train ResNet-18 on the CIFAR-100 dataset \citep{Krizhevsky2009LearningML}.}
\label{fig:loss_for_boundedgradassum}
\end{figure}
The proof can be found in Appendix \ref{appdix:prf_boundedgrad_andobj}. In the context of machine learning, we have $f_{\star} = 0$ and that the maximum training loss typically occurs around the initial point and it rarely grows without bound after that. For example, when training on CIFAR-100 with the cross-entropy loss as the objective function, suppose that, at the initial point, each class is equally likely to be output. In this case, $f_i(\bm{x}_1) = -\log p_i = -\log 1/100 \approx4.60517$. Therefore, the assumption that the loss has both upper and lower bounds appears to be a natural one. Figure \ref{fig:loss_for_boundedgradassum} presents the empirical loss values at each epoch using various learning rate schedulers. Figure \ref{fig:loss_for_boundedgradassum} shows $f(\bm{x}_1) = f^\star \approx 4.07\ldots$ and $f(\bm{x}_k)$ decreasing for epochs. Accordingly, the above discussion, together with \ref{asm:holdersmooth_gradient}, leads to the finding that, for all $k\in\mathbb{N}$, $\| \nabla f(\bm{x}_k)\|\lesssim((1+\mathfrak{h})L^{\frac{1}{\mathfrak{h}}}f(\bm{x}_1)/\mathfrak{h})^{\mathfrak{h}/(1+\mathfrak{h})}$. Since the exact gradient of the expected risk generally requires an expectation over the entire data distribution, it is typically unavailable in practice. We therefore use the gradient of a randomly sampled loss as a stochastic estimator of the true gradient. \ref{asm:unbiasedestimator_sg} ensures that this stochastic gradient coincides with the true gradient on average.  For example, \ref{asm:lowerbounded_objective}, \ref{asm:holdersmooth_gradient}, \ref{asm:bounded_gradient}, and \ref{asm:boundedpthmoment_sg} yield \ref{asm:unbiasedestimator_sg}; indeed, for any $\bm{d}\in\mathbb{R}^d$,
\begin{align*}
\langle \nabla f(\bm{x}_k),\bm{d}\rangle
&=\lim_{h\downarrow0}\frac{f(\bm{x}_k+h\bm{d})-f(\bm{x}_k)}{h}
=\lim_{h\downarrow0}\mathbb{E}_{\xi_k}\left[\frac{F(\bm{x}_k+h\bm{d};\xi_k)-F(\bm{x}_k;\xi_k)}{h}\right]\\
&=\mathbb{E}_{\xi_k}\left[\lim_{h\downarrow0}\frac{F(\bm{x}_k+h\bm{d};\xi_k)-F(\bm{x}_k;\xi_k)}{h}\right]
=\mathbb{E}_{\xi_k}[\langle \bm{g}(\bm{x}_k;\xi_k),\bm{d}\rangle]
=\langle \mathbb{E}_{\xi_k}[\bm{g}(\bm{x}_k;\xi_k)],\bm{d}\rangle,
\end{align*}
where the third last equality is from the Vitali convergence theorem, and we considered that Jensen's inequality and Cauchy-Schwartz inequality imply the uniform integrability
\begin{align*}
&\sup_{h>0}\mathbb{E}_{\xi_k}\left[\left\|\frac{F(\bm{x}_k+h\bm{d};\xi_k)-F(\bm{x}_k;\xi_k)}{h}\right\|\right]
=\sup_{h>0} \frac{1}{h}\mathbb{E}_{\xi_k}\left[\left\|\int_{0}^1\langle \bm{g}(\bm{x}_k+th\bm{d};\xi_k), th\bm{d}\rangle dt\right\|^{\mathfrak{p}\frac{1}{\mathfrak{p}}}\right]\\
&\leq\sup_{h>0} \frac{1}{h}\mathbb{E}_{\xi_k}\left[\left(\int_{0}^1\left\|\langle \bm{g}(\bm{x}_k+th\bm{d};\xi_k), th\bm{d}\rangle\right\|^{\mathfrak{p}} dt\right)^{\frac{1}{\mathfrak{p}}}\right]
\leq \sup_{h>0} C\left\|\bm{d}\right\|\left(\int_{0}^1 t^{\mathfrak{p}} dt\right)^{\frac{1}{\mathfrak{p}}}
=\frac{C\left\|\bm{d}\right\|}{(\mathfrak{p}+1)^{1/\mathfrak{p}}}
<\infty.
\end{align*}
\ref{asm:boundedpthmoment_sg} generalizes the standard bounded variance condition, corresponding to $\mathfrak{p}=2$ \citep{zou2019assuficient, wang2023closinggap}, and allows for heavier-tailed stochastic gradients that have been reported to arise in DNN training \citep{battash2024noisy}. In particular, it covers heavy-tailed settings in which stochastic gradients may exhibit extreme outliers, including cases where $\sup_{k}\mathbb{E}_{\xi_k\sim\mathcal{D}}[\|\bm{g}(\bm{x}_k;\xi_k)-\nabla f(\bm{x}_k)\|^{\mathfrak{p}}]<\infty$ while $\mathbb{E}_{\xi_k\sim\mathcal{D}}[\|\bm{g}(\bm{x}_k;\xi_k)-\nabla f(\bm{x}_k)\|^{2}]= \infty$ for some $k\in\mathbb{N}$ (as mentioned later).

We now formulate the empirical risk minimization and its mini-batch stochastic gradient. Let $\{\bm{\xi}_j\}_{j=1}^N$ be a dataset drawn i.i.d. from $\mathcal{D}$ on $(\mathbb{R}^m,\mathcal{B}(\mathbb{R}^m))$. The empirical risk can be expressed as $f_N(\bm{x})=N^{-1}\sum_{j=1}^NF(\bm{x};\bm{\xi}_j)$. For index sampling, we consider the uniform distribution $\mathcal{U}$ on $\mathbb{N}_N$. At iteration $k$, let $\iota_{k,1},\ldots,\iota_{k,b_k}$ be i.i.d. samples from $\mathcal{U}$ and define $\bm{\iota}_k\coloneq(\iota_{k,1},\ldots,\iota_{k,b_k})$. The mini-batch stochastic gradient with batch size $b_k$ is given by $\bm{g}_{B_k}(\bm{x}_k)\coloneq N^{-1}\sum_{j=1}^{b_k}\bm{g}_{\iota_{k,j}}(\bm{x}_k)\coloneq N^{-1}\sum_{j=1}^{b_k}\bm{g}(\bm{x}_k;\xi_{\iota_{k,i}})$. By replacing $\bm{g}(\bm{x}_k;\xi_k)$ with $\bm{g}_{B_k}(\bm{x}_k)$ in Algorithm \ref{algo:adam}, the updating rule is rewritten for the empirical risk minimization. Here, we prepare the alternative conditions for the empirical risk minimization:
\begin{proposition}\label{prop:unbiasedestimator_sg_empi}
The stochastic gradient is an unbiased estimator of the full gradient: \[\forall k\in\mathbb{N},\forall i\in\mathbb{N}_{b_k}:\mathbb{E}_{\iota_{k,i}\sim\mathcal{U}}[\bm{g}_{\iota_{k,j}}(\bm{x}_k)]=\nabla f_N(\bm{x}_k).\]
\end{proposition}
\begin{condition}
\begin{enumerate}[label=(C\arabic*')]
\setcounter{enumi}{4}
\item \label{asm:boundedpthmoment_sg_empi} Let $\mathfrak{p}\in(1,2]$. Then, the $\mathfrak{p}$-th moment of the stochastic gradient is bounded: \[\exists \sigma_N\geq 0,\forall k\in\mathbb{N}, \forall i\in\mathbb{N}_{b_k}:\mathbb{E}_{\iota_{k,i}\sim\mathcal{U}}[\|\bm{g}_{\iota_{k,j}}(\bm{x}_k)-\nabla f_N(\bm{x}_k)\|^{\mathfrak{p}}]\leq \sigma_N^{\mathfrak{p}}.\]
\end{enumerate}
\end{condition}
Proposition \ref{prop:unbiasedestimator_sg_empi} holds because $\mathbb{E}_{\iota_{k,i}\sim\mathcal{U}}[\bm{g}_{\iota_{k,i}}(\bm{x}_k)]=\sum_{j=1}^N \bm{g}_{j}(\bm{x}_k)\mathcal{U}(\{j\})=N^{-1}\sum_{j=1}^N \bm{g}_{j}(\bm{x}_k)=\nabla f_N(\bm{x}_k)$.

Finally, we mention to the relation between the expected risk and the empirical risk on \ref{asm:boundedpthmoment_sg} and \ref{asm:boundedpthmoment_sg_empi}. The key point is to distinguish the population distribution $\mathcal{D}$ from the index-sampling distribution $\mathcal{U}$ and to evaluate the moments of stochastic gradient noise with respect to the appropriate distribution. In the expected risk minimization, stochastic gradients are computed using data sampled directly from the population distribution $\mathcal{D}$. In contrast, in the empirical risk minimization, a finite dataset is first constructed by drawing $N$ i.i.d. samples from $\mathcal{D}$. Once this dataset is fixed, stochastic gradients are computed by sampling indices from the index set $\mathbb{N}_N$ according to the distribution $\mathcal{U}$. In particular, by the strong law of large numbers, under $F(\bm{x};\cdot)\in L^1(\Omega)$, the empirical loss converges almost surely to the expected loss as $N\to\infty$ for each fixed $\bm{x}$: namely, for $\{\bm{\xi}_j\}_{j=1}^{\infty}$ i.i.d. sampled from $\mathcal{D}$, $\lim_{N\to\infty}N^{-1}\sum_{j=1}^N F(\bm{x};\bm{\xi}_j)=\mathbb{E}_{\bm{\xi}\sim\mathcal{D}}[F(\bm{x};\bm{\xi})]$ a.s. In this sense, the expected risk can be interpreted as the population limit of the empirical risk. Here, we enumerate the primary properties and show them in order:
\begin{proposition}
\begin{enumerate}[label=(P\arabic*)]
    \item\label{parprop:hev_exp} Let $c\in(1,2)$. It is possible for both $\sup_{k\in\mathbb{N}}\mathbb{E}_{\xi_k\sim\mathcal{D}}[\|\bm{g}(\bm{x}_k;\xi_k)-\nabla f(\bm{x}_k)\|^{\mathfrak{p}}]<\infty$ for any $\mathfrak{p}\in(1,c)$ and $\mathbb{E}_{\xi_k\sim\mathcal{D}}[\|\bm{g}(\bm{x}_k;\xi_k)-\nabla f(\bm{x}_k)\|^{\mathfrak{q}}]=\infty$ for some $k\in\mathbb{N}$ and any $\mathfrak{q}\in[c,2]$ to hold.
    \item\label{parprop:hev_fullbound} It is possible for both \ref{parprop:hev_exp} and \ref{asm:bounded_gradient}.
    \item\label{parprop:hev_asbound} It is impossible for both \ref{parprop:hev_exp} and $\sup_{k\in\mathbb{N}}\|\bm{g}(\bm{x}_k;\xi_k)\|<\infty$ $\mathcal{D}$-a.s. to hold.
    \item $\sup_{k\in\mathbb{N},i\in\mathbb{N}_{b_k}}\mathbb{E}_{\iota_{k,i}\sim\mathcal{U}}[\|\bm{g}_{\iota_{k,i}}(\bm{x}_k)-\nabla f_N(\bm{x}_k)\|^{\mathfrak{p}}]<\infty$ for any $\mathfrak{p}\in(1,2)$ is equivalent to $\sup_{k\in\mathbb{N},i\in\mathbb{N}_{b_k}}\mathbb{E}_{\iota_{k,i}\sim\mathcal{U}}[\|\bm{g}_{\iota_{k,i}}(\bm{x}_k)-\nabla f_N(\bm{x}_k)\|^{2}]<\infty$; thus, the strict heavy-tailed noise with respect to the index sampling distribution does not arise.
    \item\label{parprop:hev_poplimit} However, increasing the number of data points $N$ can make the upper bound of the variance diverge without diverging the $\mathfrak{p}(\in(1,2))$-th moment; let $c\in(1,2)$; it is possible for $\lim_{N\to\infty}\sup_{k\in\mathbb{N},i\in\mathbb{N}_{b_k}}\mathbb{E}_{\iota_{k,i}\sim\mathcal{U}}[\|\bm{g}_{\iota_{k,i}}(\bm{x}_k)-\nabla f_N(\bm{x}_k)\|^{\mathfrak{p}}]<\infty$ for any $\mathfrak{p}\in(1,c)$ and $\lim_{N\to\infty}\mathbb{E}_{\iota_{k,i}\sim\mathcal{U}}[\|\bm{g}_{\iota_{k,i}}(\bm{x}_k)-\nabla f_N(\bm{x}_k)\|^{\mathfrak{q}}]=\infty$ for some $k\in\mathbb{N},i\in\mathbb{N}_{b_k}$ and any $\mathfrak{q}\in[c,2]$ to hold.
    \item\label{parprop:hav_asbound_empi} It is possible for both \ref{parprop:hev_poplimit} and \ref{asm:asboundedsg}.
\end{enumerate}
\end{proposition}
For example, setting $(\Omega,\mathcal{F})=(\mathbb{R},\mathcal{B}(\mathbb{R})), \xi\sim\mathcal{D}\coloneq\mathsf{t}_{\nu}, x\in\mathbb{R}$, and $F(x;\xi)\coloneq \log(1+e^{x\xi})$ imply those two conditions under $1<\mathfrak{p}<\nu<2$, where $\mathsf{t}_{\nu}$ is the Student's $\mathsf{t}$ distribution with the degrees of freedom $\nu$; indeed, we have $g(x_k;\xi_k)=\xi_k e^{x_k\xi_k}/(1+e^{x_k\xi_k}), |g(x_k;\xi_k)|^{\mathfrak{p}}\leq |\xi_k|^{\mathfrak{p}}$, and $|\nabla f(x_k)|^{\mathfrak{p}}\leq \mathbb{E}_{\xi_k}[|g(x_k;\xi_k)|^{\mathfrak{p}}]\leq\mathbb{E}_{\xi_k}[|\xi_k|^{\mathfrak{p}}]$, which leads to $\sup_k\mathbb{E}_{\xi_k}[|g(x_k;\xi_k)-\nabla f(x_k)|^{\mathfrak{p}}]\leq2^{\mathfrak{p}-1}\sup_k\mathbb{E}_{\xi_k}[|g(x_k;\xi_k)|^{\mathfrak{p}}+|\nabla f(x_k)|^{\mathfrak{p}}]\leq 2^{\mathfrak{p}-1}\sup_k\{\mathbb{E}_{\xi_k}[|g(x_k;\xi_k)|^{\mathfrak{p}}]+\mathbb{E}_{\xi_k}[|\xi_k|^{\mathfrak{p}}]\}\leq 2^{\mathfrak{p}}\sup_k \mathbb{E}_{\xi_k}[|\xi_k|^{\mathfrak{p}}]=2^{\mathfrak{p}}\mathbb{E}_{\xi}[|\xi|^{\mathfrak{p}}]<\infty$, where we considered that $\mathbb{E}_{\xi\sim\mathsf{t}_{\nu}}[|\xi|^{\mathfrak{p}}]<\infty$ holds for $1\leq\mathfrak{p}<\nu<2$. Next, since $\mathbb{E}_{\xi\sim\mathsf{t}_{\nu}}[|\xi|^{\mathfrak{q}}]=\infty$ holds for $1<\nu\leq\mathfrak{q}$, the density of $\mathtt{t}_{\nu}$ is even, and $a^{\mathfrak{q}}+b^{\mathfrak{q}}\geq (a+b)^{\mathfrak{q}}/2^{\mathfrak{q}-1}$ holds for $a,b\geq0$, we obtain 
\begin{align}\label{eq:calc_logit}
&\mathbb{E}_{\xi_k}[|g(x_k;\xi_k)|^{\mathfrak{q}}]
=\mathbb{E}_{\xi_k}\left[|\xi_k|^{\mathfrak{q}} \left(\frac{e^{x_k\xi_k}}{1+e^{x_k\xi_k}}\right)^{\mathfrak{q}}\right]
=\frac{1}{2}\left\{\mathbb{E}_{\xi_k}\left[|\xi_k|^{\mathfrak{q}} \left(\frac{e^{x_k\xi_k}}{1+e^{x_k\xi_k}}\right)^{\mathfrak{q}}\right]+\mathbb{E}_{\xi_k}\left[|\xi_k|^{\mathfrak{q}} \left(\frac{e^{-x_k\xi_k}}{1+e^{-x_k\xi_k}}\right)^{\mathfrak{q}}\right]\right\}\nonumber\\
&\geq \frac{1}{2}\mathbb{E}_{\xi_k}\left[|\xi_k|^{\mathfrak{q}} \frac{1}{2^{\mathfrak{q}-1}}\left(\frac{e^{x_k\xi_k}}{1+e^{x_k\xi_k}}+\frac{e^{-x_k\xi_k}}{1+e^{-x_k\xi_k}}\right)^{\mathfrak{q}}\right]
=\frac{1}{2^{\mathfrak{q}}}\mathbb{E}_{\xi_k}[|\xi_k|^{\mathfrak{q}}]
=\infty,
\end{align}
which, together with $(a-b)^{\mathfrak{q}}\geq a^{\mathfrak{q}}/2^{\mathfrak{q}-1}-b^{\mathfrak{q}}$, yields $\mathbb{E}_{\xi_k}[|g(\bm{x}_k;\xi_k)-\nabla f(\bm{x}_k)|^{\mathfrak{q}}]\geq \mathbb{E}_{\xi_k}[|g(\bm{x}_k;\xi_k)|^{\mathfrak{q}}/2^{\mathfrak{q}-1}-|\nabla f(\bm{x}_k)|^{\mathfrak{q}}]\geq 2^{1-^{\mathfrak{q}}}\mathbb{E}_{\xi_k}[|g(\bm{x}_k;\xi_k)|^{\mathfrak{q}}]-G^{\mathfrak{q}}=\infty$. Another example can be found in Section \ref{sec:num_heavy}. We note that this situation is possible under \ref{asm:bounded_gradient} but not under the $\mathcal{D}$-almost surely bounded stochastic gradient condition which is, for any $k\in\mathbb{N}$, $\|\bm{g}(\bm{x}_k;\cdot)\|\leq \bar{g}$ $\mathcal{D}$-a.s. The latter claim is obvious because $\sup_k\mathbb{E}_{\xi_k}[\|\bm{g}(\bm{x}_k;\xi_k)-\nabla f(\bm{x}_k)\|^{\mathfrak{p}}]\leq 2\mathbb{E}_{\xi_k}[\|\bm{g}(\bm{x}_k;\xi_k)\|^{\mathfrak{p}}+\|\nabla f(\bm{x}_k)\|^{\mathfrak{p}}]\leq 2(\bar{g}^{\mathfrak{p}}+G^{\mathfrak{p}})<\infty$ holds for any $\mathfrak{p}>1$. The example $f$ mentioned above satisfies \ref{asm:bounded_gradient} due to $|\nabla f(x_k)|\leq \mathbb{E}[|g(x_k;\xi_k)|]\leq\mathbb{E}[|\xi_k|]<\infty$ but does not the $\mathtt{t}_{\nu}$-almost surely bounded stochastic gradient condition, which gives an example for the former claim. Since the support of $\mathsf{t}_{\nu}$ is $\mathbb{R}$, $\mathsf{t}_{\nu}(|\xi|>c)>0$ holds for any $c\in\mathbb{R}$. Moreover, $|g(x_k;\xi)|+|g(x_k;-\xi)|=|\xi|$ is obtained from the similar calculation in \eqref{eq:calc_logit}, which implies $\{|\xi|>2c\}\subset\{|g(x;\xi)|>c\}\cup\{|g(x;-\xi)|>c\}$. Since the density $p_{\nu}$ of $\mathsf{t}_{\nu}$ is even,
\begin{align*}
\mathsf{t}_{\nu}(|g(x_k;\xi)|>c)
&=\mathbb{E}_{\xi\sim\mathsf{t}_{\nu}}[\mathbbm{1}_{\{|g(x_k;\xi)|>c\}}]
=\int_{-\infty}^{\infty}\mathbbm{1}_{\{|g(x_k;\xi)|>c\}}p_{\nu}(\xi)d\xi
=\int_{-\infty}^{\infty}\mathbbm{1}_{\{|g(x_k;-\xi)|>c\}}p_{\nu}(-\xi)d\xi\\
&=\int_{-\infty}^{\infty}\mathbbm{1}_{\{|g(x_k;-\xi)|>c\}}p_{\nu}(\xi)d\xi
=\mathbb{E}_{\xi\sim\mathsf{t}_{\nu}}[\mathbbm{1}_{\{|g(x_k;-\xi)|>c\}}]
=\mathsf{t}_{\nu}(|g(x_k;-\xi)|>c).
\end{align*}
Combining these facts, for any $c\in\mathbb{R}$, we have $0<\mathsf{t}_{\nu}(|\xi|>c)=\mathsf{t}_{\nu}(\{|g(x_k;\xi)|>c\}\cup\{|g(x_k;-\xi)|>c\})\leq \mathsf{t}_{\nu}(|g(x_k;\xi)|>c)+\mathsf{t}_{\nu}(|g(x_k;-\xi)|>c)=2\mathsf{t}_{\nu}(|g(x_k;\xi)|>c)$. Therefore, the stochastic gradient of $f$ is not $\mathtt{t}_{\nu}$-almost surely bounded.

Accordingly, although \ref{asm:boundedpthmoment_sg} and \ref{asm:boundedpthmoment_sg_empi} both impose boundedness of the $\mathfrak{p}$-th moment of stochastic gradient noise, they differ in the distribution with respect to which the expectation is taken. In particular, since $\mathcal{U}$ is a distribution over a finite set, for any fixed dataset, bounded variance of stochastic gradient noise with respect to $\mathcal{U}$ is equivalent to boundedness of its $\mathfrak{p}$-th moment for $\mathfrak{p}\in(1,2)$; indeed, the monotonicity of $L^p$ and $\ell^{p}$ norms implies
\begin{align*}
(\mathbb{E}_{\iota_{k,i}\sim\mathcal{U}}[\|\bm{g}_{\iota_{k,i}}(\bm{x}_k)-\nabla f_N(\bm{x}_k)\|^{\mathfrak{p}}])^{\frac{2}{\mathfrak{p}}}
\leq \mathbb{E}_{\iota_{k,i}\sim\mathcal{U}}[\|\bm{g}_{\iota_{k,i}}(\bm{x}_k)-\nabla f_N(\bm{x}_k)\|^{2}]
\leq N^{\frac{2}{\mathfrak{p}}-1}(\mathbb{E}_{\iota_{k,i}\sim\mathcal{U}}[\|\bm{g}_{\iota_{k,i}}(\bm{x}_k)-\nabla f_N(\bm{x}_k)\|^{\mathfrak{p}}])^{\frac{2}{\mathfrak{p}}}.
\end{align*}
However, when $\mathfrak{p}<2$, the rightmost bound contains a coefficient that diverges as the dataset size $N$ increases. Consequently, the equivalence of these moment conditions for each fixed dataset does not guarantee a variance bound that is uniform in $N$. Indeed, \citet{shestakov2026gap_expempi} showed that, when the population stochastic gradients are heavy-tailed, the empirical variance can increase with the dataset size under suitable conditions. This phenomenon can also be illustrated using the example introduced in the discussions of \ref{parprop:hev_exp}, \ref{parprop:hev_fullbound}, and \ref{parprop:hev_asbound}; let $\{\xi_i\}_{i=1}^N$ be a finite dataset consisting of i.i.d. samples from $\mathtt{t}_{\nu}$, where $1<\mathfrak{p}<\nu<2$, and consider the corresponding empirical loss $f_N(x)=N^{-1}\sum_{j=1}^N\log(1+e^{x\xi_j})$. The strong law of large numbers, and $|g_{\iota}(x_k)|\leq |\xi_{\iota}|$ yield 
\begin{align*}
&\lim_{N\to\infty}\sup_{k\in\mathbb{N},i\in\mathbb{N}_{b_k}}\mathbb{E}_{\iota_{k,i}\sim\mathcal{U}}[|g_{\iota_k}(x_k)-\nabla f_N(x_k)|^{\mathfrak{p}}]
=\lim_{N\to\infty}\sup_{k}\frac{1}{N}\sum_{j=1}^N|g_{j}(x_k)-\nabla f_N(x_k)|^{\mathfrak{p}}\\
&\leq 2^{\mathfrak{p}-1}\lim_{N\to\infty}\sup_{k}\left\{\frac{1}{N}(\sum_{j=1}^N|g_{j}(x_k)|^{\mathfrak{p}})+|\nabla f_N(x_k)|^{\mathfrak{p}}\right\}
\leq 2^{\mathfrak{p}}\lim_{N\to\infty}\sup_{k}\frac{1}{N}\sum_{j=1}^N|g_{j}(x_k)|^{\mathfrak{p}}
\leq 2^{\mathfrak{p}}\lim_{N\to\infty}\sup_{k}\frac{1}{N}\sum_{j=1}^N|\xi_j|^{\mathfrak{p}}\\
&=2^{\mathfrak{p}}\lim_{N\to\infty}\frac{1}{N}\sum_{j=1}^N|\xi_j|^{\mathfrak{p}}
=2^{\mathfrak{p}}\mathbb{E}_{\xi\sim\mathtt{t}_{\nu}}[|\xi|^{\mathfrak{p}}]<\infty,
\end{align*}
where we considered that Jensen's inequality yields $|\nabla f_N(x_k)|^{\mathfrak{p}}=|N^{-1}\sum_{j=1}^Ng_j(x_k)|^{\mathfrak{p}}=|\mathbb{E}_{\iota\sim\mathcal{U}}[g_{\iota}(x_k)]|^{\mathfrak{p}}\leq \mathbb{E}_{\iota}[|g_{\iota}(x_k)|^{\mathfrak{p}}]=N^{-1}\sum_{j=1}^N|g_{j}(x_k)|^{\mathfrak{p}}$. Recall $1<\nu\leq\mathfrak{q}$. When $x_k=0$, $\mathbbm{1}_{\{x_k\xi\geq0\}}=1$ yields $\mathbb{E}_{\xi\sim\mathtt{t}_{\nu}}[|\xi|^{\mathfrak{q}}\mathbbm{1}_{\{x_k\xi\geq0\}}]=\mathbb{E}_{\xi\sim\mathtt{t}_{\nu}}[|\xi|^{\mathfrak{q}}]=\infty$; when $x_k\gtrless0$, $\mathbbm{1}_{\{x_k\xi\geq0\}}=\mathbbm{1}_{\{\xi\gtreqless 0\}}$ yields $\mathbb{E}_{\xi\sim\mathtt{t}_{\nu}}[|\xi|^{\mathfrak{q}}\mathbbm{1}_{\{x_k\xi\geq0\}}]=\mathbb{E}_{\xi\sim\mathtt{t}_{\nu}}[|\xi|^{\mathfrak{q}}\mathbbm{1}_{\{\xi\gtreqless0\}}]=(1/2)\mathbb{E}_{\xi\sim\mathtt{t}_{\nu}}[|\xi|^{\mathfrak{q}}]=\infty$, where the inequality signs are taken in the corresponding order; thus, $\mathbb{E}_{\xi\sim\mathtt{t}_{\nu}}[|\xi|^{\mathfrak{q}}\mathbbm{1}_{\{x_k\xi\geq0\}}]=\infty$ holds. From $|g_j(x_k)|^{\mathfrak{q}}\geq |g_j(x_k)|^{\mathfrak{q}}\mathbbm{1}_{\{x_k\xi_j\geq0\}}\geq|\xi_j|^{\mathfrak{q}}\mathbbm{1}_{\{x_k\xi_j\geq0\}}/(1+e^{-x_k\xi_j})^{\mathfrak{q}}\geq|\xi_j|^{\mathfrak{q}}\mathbbm{1}_{\{x_k\xi_j\geq0\}}/2^{\mathfrak{q}}$, the strong law of large numbers implies
\begin{align*}
&\lim_{N\to\infty}\mathbb{E}_{\iota_{k,i}\sim\mathcal{U}}[|g_{\iota_{k,i}}(x_k)-\nabla f_N(x_k)|^{\mathfrak{q}}]
\geq\lim_{N\to\infty}\left\{\frac{1}{2^{\mathfrak{q}-1}}\mathbb{E}_{\iota_{k,i}\sim\mathcal{U}}[|g_{\iota_k}(x_k)|^{\mathfrak{q}}]-|\nabla f_N(x_k)|^{\mathfrak{q}}\right\}\\
&=\lim_{N\to\infty}\left\{\frac{1}{2^{\mathfrak{q}-1}N}\sum_{j=1}^N|g_{j}(x_k)|^{\mathfrak{q}}-\left|\frac{1}{N}\sum_{j=1}^N|\xi_j|\right|^{\mathfrak{q}}\right\}
\geq \lim_{N\to\infty}\left\{\frac{1}{2^{2\mathfrak{q}-1}N}\sum_{j=1}^N|\xi_j|^{\mathfrak{q}}\mathbbm{1}_{\{x_k\xi_j\geq0\}}-\left|\frac{1}{N}\sum_{j=1}^N|\xi_j|\right|^{\mathfrak{q}}\right\}\\
&=\frac{1}{2^{2\mathfrak{q}-1}}\mathbb{E}_{\xi\sim\mathtt{t}_{\nu}}[|\xi|^{\mathfrak{q}}\mathbbm{1}_{\{x_k\xi\geq0\}}]-|\mathbb{E}_{\xi\sim\mathtt{t}_{\nu}}[|\xi|]|^{\mathfrak{q}}
=\infty.
\end{align*}
This example implies that, although the variance is finite for every fixed finite dataset, it diverges almost surely as $N$ increases, while the $\mathfrak{p}$-th moment remains uniformly bounded. Thus, the bounded variance for a fixed dataset alone is insufficient to capture the heavy-tailed nature of stochastic gradient noise at the population level. These observations motivate convergence analyses under bounded $\mathfrak{p}$-th moment assumptions in both the expected risk and the empirical risk minimization. Such analyses not only accommodate infinite-variance stochastic gradient noise at the population level but also account for stochastic gradient noise whose variance may grow unboundedly with the dataset size in the empirical risk minimization. The derivation of \ref{parprop:hav_asbound_empi} is given in Section \ref{sec:anal_bcg}.

\section{Analysis for Expected Risk Under Bounded $\mathfrak{p}$-th Moment Noise}\label{sec:gen_case}
We consider the expected risk minimization in this section. Theorem \ref{thm:heavy_conv} is the convergence analysis of vanilla Adam \ref{algo:adam} under the general conditions: H\"{o}lder smoothness and bounded $\mathfrak{p}$-th moment. The proofs of Theorem \ref{thm:heavy_conv} and Corollary \ref{cor:heavy_convrate} are in Appendix \ref{appdix:prf_heavy_conv}.
\begin{theorem}\label{thm:heavy_conv}
Set $(\beta_{1,k})_k$ and $(\alpha_k)_k$ satisfying $\bar{\beta}_1\coloneq \sup_{k\in\mathbb{N}}\beta_{1,k}<1$, $\alpha_k \in [0, \alpha_{\max}] \subset[0,1)$, and $\alpha_{\max}>0$ respectively. Then, under \ref{asm:lowerbounded_objective}--\ref{asm:boundedpthmoment_sg} and $1+\mathfrak{h}\leq\mathfrak{p}$, a sequence $(\bm{x}_k)_{k\in\mathbb{N}}$ generated by vanilla Adam \ref{algo:adam} satisfies the following properties.  
\begin{enumerate}[label=1-(\roman*)]
\item for any $K\in\mathbb{N}$,
\begin{align*}
&\underset{1\leq k\leq K}{\min}\mathbb{E}_{(\xi_1,\ldots,\xi_K)\sim\mathcal{D}^{\otimes K}}[\|\nabla f(\bm{x}_k)\|^2]
\leq \frac{1}{\sum_{k=1}^{K}\alpha_k}\sum_{k=1}^K\alpha_{k}\mathbb{E}_{(\xi_1,\ldots,\xi_K)\sim\mathcal{D}^{\otimes K}}[\|\nabla f(\bm{x}_k)\|^2]\\
&\leq Q_1\left(\frac{1}{\sum_{k=1}^{K}\alpha_k}\right)^{\frac{\mathfrak{p}}{1+\mathfrak{p}}}+ Q_2\left(\frac{\sum_{k=1}^K\beta_{1,k}}{\sum_{k=1}^{K}\alpha_k}\right)^{\frac{\mathfrak{p}}{1+\mathfrak{p}}} +Q_3\left(\frac{\sum_{k=1}^K\alpha_k^{1+\mathfrak{h}}}{\sum_{k=1}^{K}\alpha_k}\right)^{\frac{\mathfrak{p}}{1+\mathfrak{p}}}+Q_4\left(\frac{\sum_{k=1}^{K}\alpha_k(1 - \beta_{2,k})^{\mathfrak{h}/2}}{\sum_{k=1}^{K}\alpha_k}\right)^{\frac{\mathfrak{p}}{1+\mathfrak{p}}},
\end{align*}
where $Q_1,Q_2,Q_3$, and $Q_4$ are independent of $K$. 
\item If $(\alpha_k)_k, (\beta_{1,k})_k$ and $(\beta_{2,k})_k$ satisfy $\sum_{k=1}^{\infty}\alpha_k=\infty,\ \sum_{k=1}^{\infty}\alpha_k^{1+\mathfrak{h}}<\infty,\ \sum_{k=1}^{\infty}\beta_{1,k}^2<\infty$, and $\sum_{k=1}^{\infty}\alpha_k(1 - \beta_{2,k})^{\mathfrak{h}/2}<\infty$, then $\min_{1\leq k\leq K}\mathbb{E}[\|\nabla f_N(\bm{x}_k)\|^2]=O(1/\sum_{k=1}^{K-1}\alpha_k)$ holds, which implies $\liminf_{k\to\infty}\mathbb{E}[\|\nabla f(\bm{x}_k)\|^2]=0$.
\end{enumerate}
\end{theorem}
If $\alpha_{k}\leq\alpha_{k-1}$ and $\beta_k\leq \beta_{k-1}$ are additionally assumed, we can replace $\sum_{k=1}^K\beta_{1,k}$ with $\sum_{k=1}^K\alpha_k\beta_{1,k}$. An example of $(\beta_{1,k})_{k\in\mathbb{N}}$ satisfying $\sum_{k=1}^{\infty}\beta_{1,k}<\infty$ is $\beta_{1,k}\coloneq\beta_{1,\max}\zeta^{\lfloor k/E\rfloor}$, where $\beta_{1,\max}\in[0,1),E\in\mathbb{N}$, and $\zeta\in(0,1)$. Examples of $(\alpha_k)_{k\in\mathbb{N}}$ and $(\beta_{2,k})_{k\in\mathbb{N}}$ satisfying $\sum_{k=1}^{\infty}\alpha_k=\infty, \sum_{k=1}^{\infty}\alpha_k^{1+\mathfrak{h}}<\infty$, and $\sum_{k=1}^{\infty}\alpha_k(1 - \beta_{2,k})<\infty$ are $\alpha_k \coloneq\alpha_{\max}k^{-r}$ and $\beta_{2,k}\coloneq 1-\tilde{\beta}_{2}/k^{s}$, where $r\in(1/(1+\mathfrak{h}),1], s\in(2(1-r)/\mathfrak{h},\infty)$, and $\tilde{\beta}_2\in(0,1)$. Applying these examples to Theorem \ref{thm:heavy_conv}, we derive the concrete convergence rates as in Corollary \ref{cor:heavy_convrate}.
\begin{corollary}\label{cor:heavy_convrate}
We assume \ref{asm:lowerbounded_objective}--\ref{asm:boundedpthmoment_sg} and $1+\mathfrak{h}\leq\mathfrak{p}$. Set $\beta_{1,\max}\in[0,1), E\in\mathbb{N}, \tilde{\beta}_{2},\zeta\in(0,1), r\in(0,1], s>2(1-r)/\mathfrak{h}, \beta_{1,k}\coloneq\beta_{1,\max}\zeta^{\lfloor k/M\rfloor},\beta_{2,k}=1-\tilde{\beta}_{2}/k^s$, and $\alpha_k \coloneq\alpha_{\max}k^{-r}$. Then, it holds that
\begin{align*}
  \underset{1\leq k\leq K}{\min}\mathbb{E}[\|\nabla f(\bm{x}_k)\|^2]=
  \begin{dcases}
    O\left(K^{-\frac{(1-r)\mathfrak{p}}{1+\mathfrak{p}}}\right) & \text{if}\quad r\in\left(\frac{1}{1+\mathfrak{h}},1\right),\\
    O\left((\log K)^{-\frac{\mathfrak{p}}{1+\mathfrak{p}}}\right) & \text{if}\quad r=1\\
    O\left(\left(\frac{\log K}{K^{\frac{\mathfrak{h}}{1+\mathfrak{h}}}}\right)^{\frac{\mathfrak{p}}{1+\mathfrak{p}}}\right) & \text{if}\quad r=\frac{1}{1+\mathfrak{h}}.
  \end{dcases}
\end{align*}
\end{corollary}
When $0<r<1/(1+\mathfrak{h})$ and $s>2r$, we can derive the convergence rate of $O(1/K^{\frac{r\mathfrak{h}\mathfrak{p}}{1+\mathfrak{p}}})$. Combining Theorem \ref{thm:heavy_conv} and Corollary \ref{cor:heavy_convrate} yields both asymptotic and non-asymptotic convergence result for vanilla Adam under the heavy-tailed noise. A limitation of this result is that it does not cover heavy-tailed noise when $\mathfrak{h}=1$, corresponding to Lipschitz smoothness. The convergence rate for $r=1/(1+\mathfrak{h})$ leads to those for $r\in(0,1]\backslash\{1/(1+\mathfrak{h})\}$, but those for $r\in(0,1]\backslash\{1/(1+\mathfrak{h})\}$ do not; thus, the convergence rate for $r=1/(1+\mathfrak{h})$ is the best among those for $r\in(0,1]$. In particular, when $1+\mathfrak{h}=\mathfrak{p}$, the convergence rate become $O(\{(\log K)^{\mathfrak{p}}/K^{\mathfrak{p}-1}\}^{1/(1+\mathfrak{p})})$. Since the convergence rates are the best among $\mathfrak{p}\in(1,2]$ when the variance is bounded, i.e., when $\mathfrak{p}=2$, Corollary \ref{cor:heavy_convrate} claims that heavy-tailed noise worsens the convergence rates of vanilla Adam. The convergence guarantees in Theorem \ref{thm:heavy_conv} and Corollary \ref{cor:heavy_convrate} depend on the condition $1+\mathfrak{h}\leq\mathfrak{p}$. Since $\mathfrak{h}$ cannot be reliably estimated in practice, verifying the condition $1+\mathfrak{h}\leq\mathfrak{p}$ on real-world tasks would be infeasible. In Section \ref{sec:num_heavy}, we therefore conduct numerical experiments on a synthetic function for which $\mathfrak{h}$ and $\mathfrak{p}$ can be fully controlled.

\subsection{Numerical Experiments under Controlled Heavy-Tailed Noise}\label{sec:num_heavy}
\begin{figure}[htbp]
\begin{minipage}[ht]{0.39\linewidth}
\centering
\includegraphics[width=0.95\linewidth]{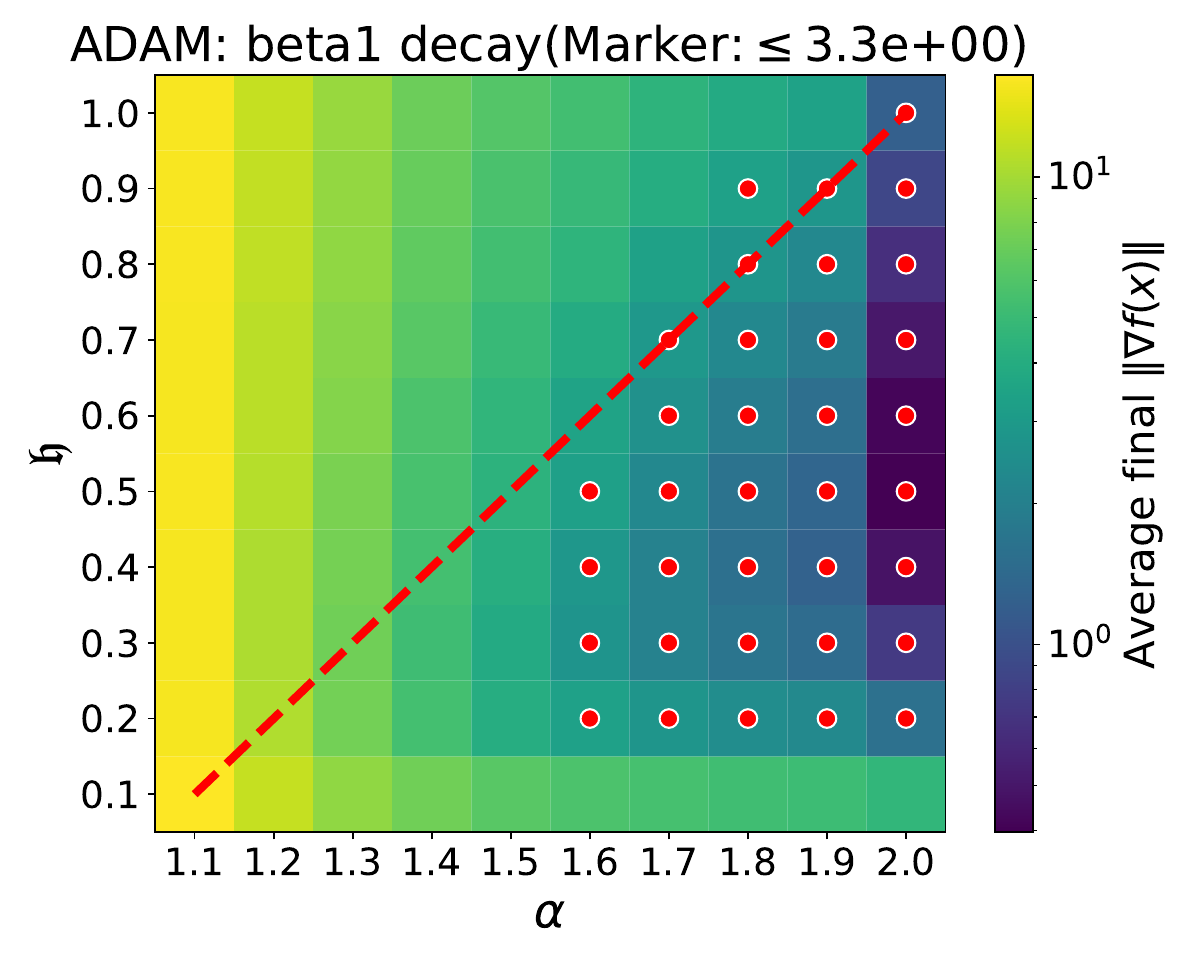}
\caption{Convergence performance across various $(\mathfrak{h},\alpha)$ pairs. The heatmaps show the average final gradient norms after 4,000 steps of vanilla Adam. The top 35 performing combinations are indicated by red markers. Below the red dotted line, the theoretical
condition by our convergence analysis $1+\mathfrak{h}< \alpha$ is satisfied.}
\label{fig:heat_decbeta}
\end{minipage}
\begin{minipage}[ht]{0.58\linewidth}
\centering
\includegraphics[width=0.48\linewidth]{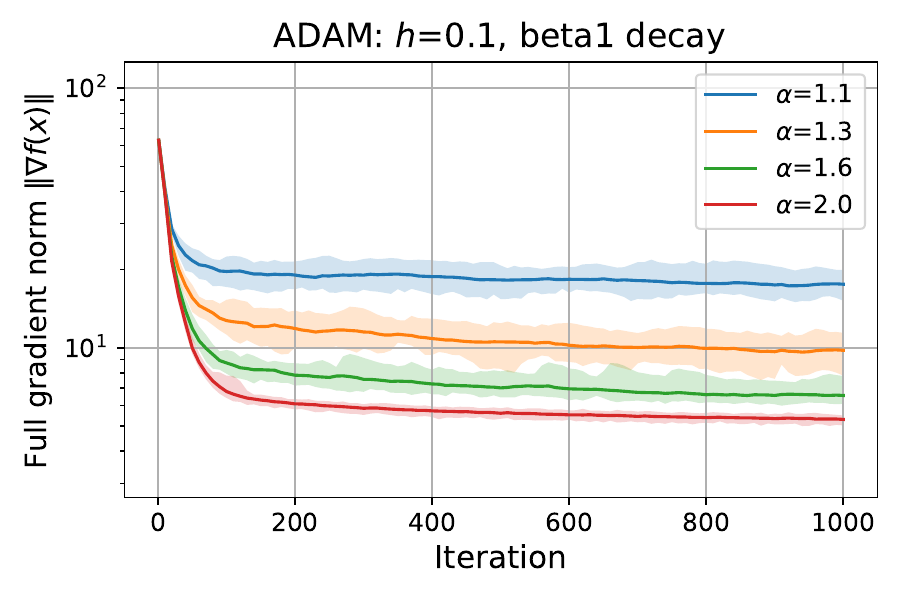}
\includegraphics[width=0.48\linewidth]{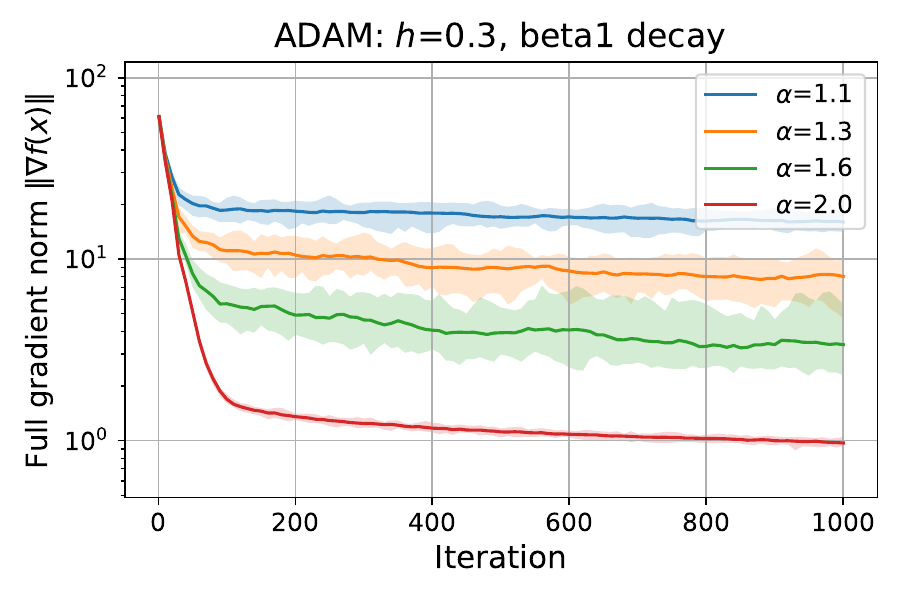}
\includegraphics[width=0.48\linewidth]{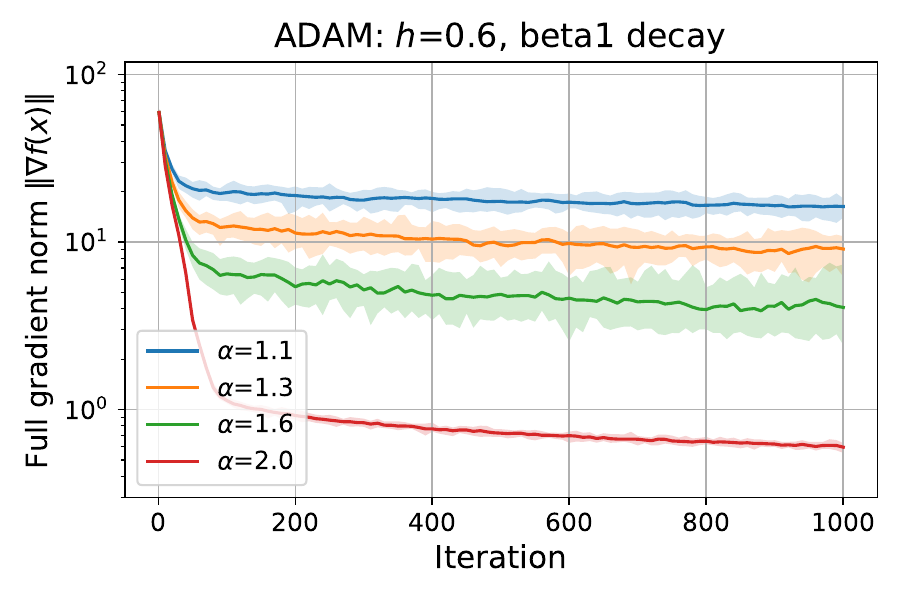}
\includegraphics[width=0.48\linewidth]{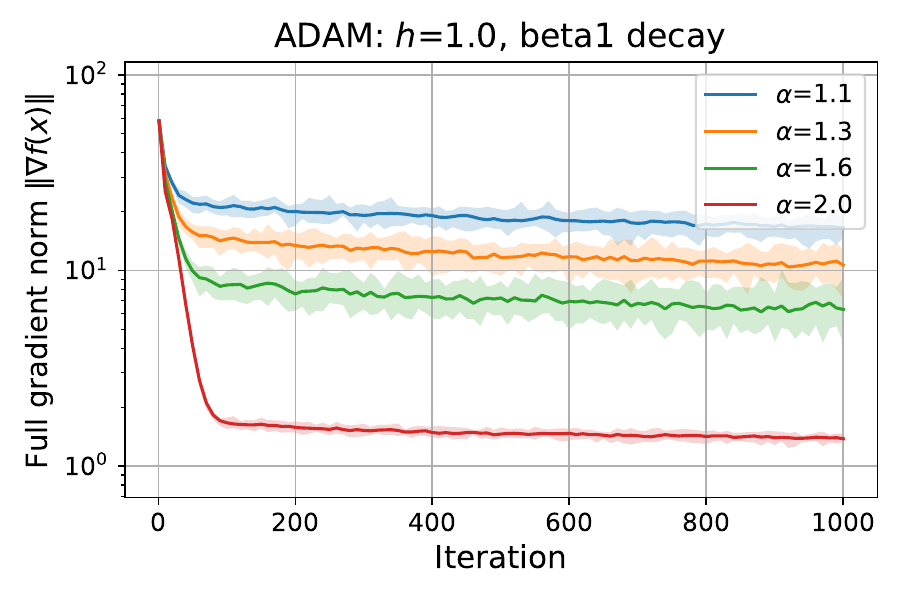}
\caption{Convergence trajectories of vanilla Adam. The plots illustrate the evolution of gradient norms over 1,000 steps for the functions defined
in \eqref{eq:defobj_heavy}. Solid lines represent the average of 20 independent trials, while the lightly shaded regions indicate the range
between the minimum and maximum values.}
\label{fig:conv_traj}   
\end{minipage}
\end{figure}
Based on \citet{yamada2026heavysgd}, we conduct numerical experiments to verify $1+\mathfrak{h}\leq \mathfrak{p}$ and support the convergence analysis. The objective function $f$ considered in our experiments is defined as
\begin{equation}\label{eq:defobj_heavy}
f(\boldsymbol{x})\coloneq \mathbb{E}_{\bm{\xi}\sim\mathcal{D}}[F(\bm{x};\bm{\xi})],
\ F(\bm{x};\bm{\xi})\coloneq\sum_{j=1}^{d}
\left\{\int_{0}^{x_j}\text{sign}(s)\min\left\{|s|^{\mathfrak{h}},1\right\}ds
-\frac{1}{2}\cos(2\pi x_j)+\frac{1}{2}\right\}+\lambda\langle\bm{x},\bm{\xi}\rangle,
\end{equation}
where $\bm{x}\coloneq (x_1,\ldots,x_d)^{\top}\in\mathbb{R}^d$, $d$ is set to $500$ in all experiments, and $\mathfrak{h}$ can be chosen arbitrarily from $(0,1]$. Our experiment treats the two-sided Pareto distribution \citep{fatkhullin2025can} as $\mathcal{D}$: $\bm{\xi}_k=(\xi_{k,1},\ldots,\xi_{k,d})^{\top}$ such that $\xi_{k,j}\coloneq s_{k,j} u_{k,j}^{-1/\alpha}$, where $s_{k,j}\sim \mathcal{U}(\{\pm1\}),u_{k,j}\sim \mathcal{U}(0,1)$, and $\alpha\in(1,2)$. For the case $\alpha=2$, we set $\bm{\xi}_k$ to standard Gaussian noise. Under this $\mathcal{D}$, we can derive $\bm{g}(\bm{x};\bm{\xi})=\nabla f(\bm{x})+\lambda\bm{\xi}$. The $\mathfrak{p}$-th moment of the stochastic gradient is bounded for any $\mathfrak{p}\in(1,\alpha)$, but not for any $\mathfrak{p}\in[\alpha,2)$. When $\alpha=2$, the variance is bounded, namely, heavy-tailed noise does not arise. The objective function satisfies \ref{asm:lowerbounded_objective}, \ref{asm:holdersmooth_gradient}, \ref{asm:bounded_gradient}, and \ref{asm:unbiasedestimator_sg}. The derivations of the properties discussed above can be found in Appendix \ref{appdix:prf_boundedgrad_andobj}. $\lambda$ is set to $0.1$; the learning rate is set to $\alpha_k\coloneq \alpha_{\max}/k^{1/(1+\mathfrak{h})}$, where $\alpha_{\max}\in\{1.0,0.5,0.25,0.05\}$; the momentum parameters are set to $\beta_{1,k}=0.9(0.99)^{\lfloor k/E\rfloor},\beta_2=1-0.75/k^{2.01/(1+\mathfrak{h})}$, where $E=400$ in Figure \ref{fig:heat_decbeta}, and $E=100$ in Figures \ref{fig:conv_traj}.

We empirically investigate the condition $1+\mathfrak{h}\leq\mathfrak{p}$ used to establish the convergence guarantees for vanilla Adam. To this end, we consider 100 combinations of $\mathfrak{h}\in\{0.1,\ldots,1.0\}$ and $\alpha\in\{1.1,\ldots,2.0\}$ and run vanilla Adam for 4,000 iterations for each combination. Figure \ref{fig:heat_decbeta} presents heatmaps of the results averaged over 20 independent trials, with red markers highlighting the 35 best-performing combinations. Since \ref{asm:boundedpthmoment_sg} holds for $\mathfrak{p}<\alpha$, the condition $1+\mathfrak{h}\leq\mathfrak{p}$ can be satisfied when $1+\mathfrak{h}<\alpha$, which corresponds to the lower-right triangular region in Figure \ref{fig:heat_decbeta}. As shown in the figure, this region contains almost all markers indicating high performance, suggesting that the observed results are broadly consistent with the theoretical condition $1+\mathfrak{h}<\alpha$ implied by our convergence analysis. Moreover, Adam exhibits degraded performance as $\alpha$ decreases, providing empirical evidence that heavier-tailed noise adversely affects its convergence performance. This observation is also consistent with the convergence-rate bound established in Corollary \ref{cor:heavy_convrate}, which deteriorates as $\mathfrak{p}$ decreases.

Next, we examine the convergence behavior in more detail. Figure \ref{fig:conv_traj} presents convergence curves for $\alpha\in\{1.1,1.3,1.6,2.0\}$ at four fixed values of $\mathfrak{h}$, which increase from the upper-left to the upper-right, lower-left, and lower-right panels. As $\mathfrak{h}$ increases, the condition $1+\mathfrak{h}<\alpha$ is violated successively, first by the blue curve in the upper-right panel and subsequently by the orange curve. Compared with the upper-left panel, the convergence performance of these curves deteriorates progressively as the condition is violated. In contrast, the red curve, which satisfies $1+\mathfrak{h}\leq\mathfrak{p}$ in all four panels, maintains its convergence performance throughout. These observations highlight the importance of the relationship between the smoothness parameter $\mathfrak{h}$ and the tail parameter $\alpha$, suggesting that considering the broader class of H\"older-smooth objectives, rather than restricting the analysis to the conventional Lipschitz smooth setting, is beneficial for understanding Adam under heavy-tailed noise.

\section{Analysis for Empirical Risk Under Almost Surely Bounded Stochastic Gradients}\label{sec:anal_bcg}
\begin{condition}
\begin{enumerate}[label=(C\arabic*')]
  \setcounter{enumi}{2}
  \item\label{asm:asboundedsg} For a fixed dataset, the norm of the stochastic gradient is almost surely bounded with respect to the index sampling $\iota_{k,i}\sim\mathcal{U}$: \[\exists G_N>0,\forall k\in\mathbb{N},\forall i\in\mathbb{N}_{b_k}:\|\bm{g}_{\iota_{k,i}}(\bm{x}_k)\|\leq G_N\quad \mathcal{U}\text{-a.s.}\] 
\end{enumerate}
\end{condition}
We consider the empirical risk minimization under \ref{asm:asboundedsg} in this section. \ref{asm:asboundedsg} \citep{zaheer2018adam_convanal,defossez2022simpleconv,Liang2025convadam} is tighter than \ref{asm:bounded_gradient} for $f_N$, thereby yielding some strong properties. For example, for any $\mathfrak{p}\in(1,2]$, the $\mathfrak{p}$-th moment of the stochastic gradient is upper bounded; namely, under \ref{asm:asboundedsg}, \ref{asm:boundedpthmoment_sg_empi} always holds for any $\mathfrak{p}\in(1,2]$ as
\begin{align*}
\mathbb{E}_{\iota_{k,i}\sim\mathcal{U}}[\|\bm{g}_{\iota_{k,i}}(\bm{x}_k)-\nabla f_N(\bm{x}_k)\|^{\mathfrak{p}}]\leq (2G_N)^{\mathfrak{p}}.
\end{align*}
Recall the example $f_N(x)=N^{-1}\sum_{j=1}^N\log(1+e^{x\xi_j}),\ \xi_j\sim\mathtt{t}_\nu$ for $1<\mathfrak{p}<\nu<2$ satisfies \ref{parprop:hev_poplimit}. Since $|g_j(\bm{x}_k)|\leq |\xi_j|$, we can take $G_N^{\mathfrak{p}}=\max_{j\in\mathbb{N}_N}|\xi_j|^{\mathfrak{p}}$, which implies \ref{parprop:hav_asbound_empi}. Thus, even in the empirical risk under \ref{asm:asboundedsg}, the heavy-tailed behavior CAN be observed by increasing the dataset size $N$. Note that $G_N$ is random with respect to the sampling of the dataset from $\mathcal{D}$, but deterministic with respect to the index sampling from $\mathcal{U}$. As mentioned in \citet{shestakov2026gap_expempi}, the effects of heavy-tailed noise as the dataset size $N$ increases are reflected in dataset-dependent constants of the loss function, such as $G_N$ and $L_{\mathfrak{h},N}$. For instance, in the preceding example, $G_N$ grows as $N$ increases, and diverges $\mathtt{t}_{\nu}$-a.s. The former is obvious due to $A\subset B\implies \max_A\leq\max_B$. Since the support of $\mathtt{t}_{\nu}$ is $\mathbb{R}$, $\mathtt{t}_{\nu}(|\xi|>c)>0$ holds for any $c\in\mathbb{R}$. Thus, by considering what $\{\xi_j\}_{j=1}^N$ is independent, for any $c\in\mathbb{R}$, $N\to\infty$ leads to
\begin{align*}
\mathtt{t}_{\nu}(G_N\leq c)
&=\mathtt{t}_{\nu}(\max_{j\in\mathbb{N}_N}|\xi_j|\leq c)
=\mathtt{t}_{\nu}\left(\bigcap_{j=1}^N\{|\xi_j|\leq c\}\right)
=\prod_{j=1}^N\mathtt{t}_{\nu}\left(|\xi_j|\leq c\right)\\
&=\mathtt{t}_{\nu}\left(|\xi|\leq c\right)^N
=(1-\mathtt{t}_{\nu}\left(|\xi|> c\right))^N
\to 0.
\end{align*}
Since $G_N$ is monotonically increasing, $\mathbbm{1}_{\{\sup_{N\in\mathbb{N}} G_N\leq c\}}=\lim_{N\to\infty} \mathbbm{1}_{\{G_N\leq c\}}$ holds. From these equations and the monotone convergence theorem, we have
\begin{align*}
\mathtt{t}_{\nu}\left(\lim_{N\to\infty}G_N\leq c\right)
&=\mathtt{t}_{\nu}\left(\sup_{N\in\mathbb{N}}G_N\leq c\right)
=\int_{\mathbb{R}}\mathbbm{1}_{\{\sup_{N\in\mathbb{N}} G_N\leq c\}}(x)\mathtt{t}_{\nu}(dx)
=\int_{\mathbb{R}}\lim_{N\to\infty} \mathbbm{1}_{\{G_N\leq c\}}(x)\mathtt{t}_{\nu}(dx)\\
&=\lim_{N\to\infty}\int_{\mathbb{R}} \mathbbm{1}_{\{G_N\leq c\}}(x)\mathtt{t}_{\nu}(dx)
=\lim_{N\to\infty}\mathtt{t}_{\nu}\left(G_N\leq c\right)
=0,
\end{align*}
which implies $\forall c\in\mathbb{R}:\lim_{N\to\infty}G_N> c$ $\mathtt{t}_{\nu}$-a.s. Therefore, we obtain $\lim_{N\to\infty}G_N=\infty$ $\mathtt{t}_{\nu}$-a.s. The upper bounds derived in this section concern gradient norms in expectation with respect to the index sampling distribution $\mathcal{U}$ for a fixed dataset. As discussed above, the dataset-dependent constant $G_N$ or $L_{\mathfrak{h},N}$ can increase with the dataset size $N$, potentially worsening these upper bounds. Moreover, taking expectations on both sides of the bounds with respect to the random dataset drawn from $\mathcal{D}$ may yield bounds that diverge as $N\to\infty$, due to the growth of $G_N$ or $L_{\mathfrak{h},N}$. In this regard, the upper bounds established in this section are meaningful because they explicitly capture how the effects of heavy-tailed noise can manifest through dataset-dependent constants.

Another benefit of \ref{asm:asboundedsg} is that \ref{asm:asboundedsg} gives $H_k^{-1}$ a lower bound. Apart from the fact that $\mathfrak{p}$ no longer needs to be considered directly, this property is the main difference from the analysis of Adam in Section \ref{sec:gen_case}. From Lemma \ref{lem:appdix_boundgradnorm} and the update rule of Algorithm \ref{algo:adam}, we have $\sqrt{\hat{v}_{k,i}} = \sqrt{v_{k,i}/(1-\beta_{2,k}^k)} \leq G_N/\sqrt{1-\bar{\beta}_2}$, which implies \[\mu\coloneq\frac{\sqrt{1-\bar{\beta}_2}}{G_N + \epsilon\sqrt{1-\bar{\beta}_2}} \leq \frac{1}{\sqrt{\hat{v}_{k,i}} + \epsilon} \leq \frac{1}{\epsilon}\eqcolon\nu,\] where $\bar{\beta}_2 \coloneq \sup_k \beta_{2,k}$. Hence, $\mu I \preceq H_k^{-1} \preceq \nu I$ holds. As mentioned in the proof of \ref{thm:holder_bcg_upperbound_beta2}, one can take a $\mu$ independent of $\bar{\beta}_2$.
\begin{proposition}\label{prop:bounds_paramofadam}
In Algorithm \ref{algo:adam}, under \ref{asm:asboundedsg}, there exist $\mu,\nu>0$ such that $\forall k\in\mathbb{N}:\mu I \preceq H_k^{-1} \preceq \nu I$ $\mathcal{U}$-a.s.
\end{proposition}

\subsection{H\"older Smooth Objectives}\label{sec:holder_anal}
\begin{theorem}\label{thm:holder_bcg_conv}
Set $(\beta_{1,k})_k$ and $(\alpha_k)_k$ satisfying $\bar{\beta}_1\coloneq \sup_{k\in\mathbb{N}}\beta_{1,k}<1$, $\alpha_k \in [0, \alpha_{\max}] \subset[0,1)$, and $\alpha_{\max}>0$ respectively. Let $S(K)\coloneq\sum_{j=1}^K b_j$. Then, under \ref{asm:lowerbounded_objective} and \ref{asm:holdersmooth_gradient} for $f_N$, and \ref{asm:asboundedsg}, a sequence $(\bm{x}_k)_{k\in\mathbb{N}}$ generated by vanilla Adam \ref{algo:adam} satisfies the following properties.
\begin{enumerate}[label=2-(\roman*)]
\item\label{thm:holder_bcg_upperbound_beta2} For any $K\in\mathbb{N}$,
    \begin{align*}
    &\underset{1\leq k\leq K}{\min}\mathbb{E}_{(\bm{\iota}_1,\ldots,\bm{\iota}_K)\sim\mathcal{U}^{\otimes S(K)}}[\|\nabla f_N(\bm{x}_k)\|^2]
    \leq \frac{1}{\sum_{k=1}^{K}\alpha_k}\sum_{k=1}^{K}\alpha_{k}\mathbb{E}_{(\bm{\iota}_1,\ldots,\bm{\iota}_K)\sim\mathcal{U}^{\otimes S(K)}}[\|\nabla f_N(\bm{x}_k)\|^2]\\
    &\leq Q_1\frac{1}{\sum_{k=1}^{K}\alpha_k}+Q_2\frac{\sum_{k=1}^{K}\alpha_k\beta_{1,k}^2}{\sum_{k=1}^{K}\alpha_k}+Q_3\frac{\sum_{k=1}^{K}\alpha_k(1 - \beta_{2,k})}{\sum_{k=1}^{K}\alpha_k}+Q_4\frac{\sum_{k=1}^{K}\alpha_k^{1+\mathfrak{h}}}{\sum_{k=1}^{K}\alpha_k},
    \end{align*}
    where $Q_1=2(f(\bm{x}_1)-f_{\star})/\mu, Q_2=8\nu G_N^2/\mu^2, Q_3=G_N^4/\{\epsilon^4\mu^2(1-\beta_{2,1})\}$, and $Q_4=2L_{\mathfrak{h},N}\nu^{1+\mathfrak{h}}G_N^{1+\mathfrak{h}}/\{(1+\mathfrak{h})\mu\}$.
\item \label{thm:holder_bcg_small_o}If $(\alpha_k)_k, (\beta_{1,k})_k$ and $(\beta_{2,k})_k$ satisfy $\sum_{k=1}^{\infty}\alpha_k=\infty,\ \sum_{k=1}^{\infty}\alpha_k^{1+\mathfrak{h}}<\infty,\ \sum_{k=1}^{\infty}\beta_{1,k}^2<\infty$, and $\sum_{k=1}^{\infty}\alpha_k(1 - \beta_{2,k})<\infty$ (or $\sum_{k=1}^{K}b_k^{-1}<\infty$), then $\min_{1\leq k\leq K}\mathbb{E}[\|\nabla f_N(\bm{x}_k)\|^2]=o(1/\sum_{k=1}^{K-1}\alpha_k)$ holds, which implies $\liminf_{k\to\infty}\mathbb{E}[\|\nabla f_N(\bm{x}_k)\|^2]=0$.
\end{enumerate}
\end{theorem}
The proof of Theorem \ref{thm:holder_bcg_conv} is in Appendix \ref{appdix:prf_holder_bcg_conv}. By replacing $Q_3$ with $2\nu G_N^2/\mu^2$, we can replace $\sum_{k=1}^{K}\alpha_k(1 - \beta_{2,k})$ with $\sum_{k=1}^{K}\alpha_k/b_k$. Corollary \ref{cor:holder_bcg_convrate} can be shown by the similar calculations for Corollary \ref{cor:heavy_convrate}. If $G_N$ or $L_{\mathfrak{h},N}$ grows by increasing $N$, the upper bounds in Theorem \ref{thm:holder_bcg_conv} worsen. Since \ref{parprop:hev_poplimit} i.e., heavy-tailed noise can arise by $N\to\infty$, heavy-tailed noise can worsen the upper bounds. An example of $(\beta_{1,k})_{k\in\mathbb{N}}$ has already given in Section \ref{sec:gen_case}. Examples of $(\alpha_k)_{k\in\mathbb{N}}$ and $(\beta_{2,k})_{k\in\mathbb{N}}$ satisfying $\sum_{k=1}^{\infty}\alpha_k=\infty, \sum_{k=1}^{\infty}\alpha_k^{1+\mathfrak{h}}<\infty$, and $\sum_{k=1}^{\infty}\alpha_k(1 - \beta_{2,k})<\infty$ are $\alpha_k \coloneq\alpha_{\max}k^{-r}$ and $\beta_{2,k}\coloneq 1-\tilde{\beta}_{2}/k^{s}$, where $r\in(1/(1+\mathfrak{h}),1], s\in(1-r,\infty)$, and $\tilde{\beta}_2\in(0,1)$. Since an increasing batch size is mentioned in Section \ref{sec:lip_special}, this section focus only on the case of varying $\beta_{2,k}$. Combining Theorem \ref{thm:holder_bcg_conv} and Corollary \ref{cor:holder_bcg_convrate} yields both asymptotic and non-asymptotic convergence results for vanilla Adam under H\"older smooth assumption.
\begin{corollary}\label{cor:holder_bcg_convrate}
We assume the same assumptions with Theorem \ref{thm:holder_bcg_conv}. Set $\beta_{1,\max}\in[0,1), \tilde{\beta}_2\in(0,1), E\in\mathbb{N}, \zeta\in(0,1), \beta_{1,k}\coloneq\beta_{1,\max}\zeta^{\lfloor k/E\rfloor}, r\in(0,1], s>1-r, \beta_{2,k}\coloneq 1-\tilde{\beta}_2/k^s,b_k=b$, and $\alpha_k \coloneq\alpha_{\max}k^{-r}$. Then, it holds that
\begin{align*}
  \underset{1\leq k\leq K}{\min}\mathbb{E}[\|\nabla f_N(\bm{x}_k)\|^2]=
  \begin{dcases}
    o\left(\frac{1}{K^{1-r}}\right) & \text{if}\quad r\in\left(\frac{1}{1+\mathfrak{h}},1\right), \\
    o\left(\frac{1}{\log K}\right) & \text{if}\quad r=1,\\
    O\left(\frac{\log K}{K^{\frac{\mathfrak{h}}{1+\mathfrak{h}}}}\right) & \text{if}\quad r=\frac{1}{1+\mathfrak{h}}.
  \end{dcases}
\end{align*}
\end{corollary}
When $0<r<1/(1+\mathfrak{h})$ and $s>r\mathfrak{h}$, we can also derive the convergence rate of $O(1/K^{r\mathfrak{h}})$. For any $1/(1+\mathfrak{h})<r<1$, $O(\log K/K^{\mathfrak{h}/(1+\mathfrak{h})})=o(1/K^{1-r})$ holds, but $o(1/K^{1-r})=O(\log K/K^{\mathfrak{h}/(1+\mathfrak{h})})$ does not; thus, the convergence rate for $r=1/(1+\mathfrak{h})$ is the best among those for $r\in[0,1]\backslash\{1/(1+\mathfrak{h})\}$. For sufficiently small $\tau>0$, setting $r=\tau+1/(1+\mathfrak{h})$ yields the convergence rate $o(1/K^{\mathfrak{h}/(1+\mathfrak{h})-\tau})$, which shows that taking a smaller $\tau$ improves this convergence rate and matches the situation. The convergence rate $O(\log K/K^{\mathfrak{h}/(1+\mathfrak{h})})$ coincides with that of SGD under H\"older smoothness \citep{yamada2026heavysgd}.
 
\subsection{The Lipschitz Smooth Special Case}\label{sec:lip_special}
\begin{condition}
\begin{enumerate}[label=(C\arabic*')]
  \setcounter{enumi}{1}
  \item\label{asmmodi:lipsmooth_gradient} $f_N$ is $L_{1,N}$-smooth; namely, $f_N$ satisfies \ref{asm:holdersmooth_gradient} with $\mathfrak{h}=1$. Let $L_N$ be $L_{1,N}$.
  \end{enumerate}
\end{condition}
In this section, we give an upper bound under the tighter smooth condition \ref{asmmodi:lipsmooth_gradient}. The main difference from Theorem~\ref{thm:holder_bcg_conv} is that the summability condition $\sum_{k=1}^{\infty}\alpha_k^{1+\mathfrak{h}}<\infty$, which was required for convergence in that theorem, is no longer needed. This is because the remainder term in the descent lemma is now quadratic rather than of order $1+\mathfrak{h}$, allowing the term corresponding to $\alpha_k^{1+\mathfrak{h}}$ in Theorem \ref{thm:holder_bcg_conv} to be absorbed into the squared gradient-norm term used as the convergence measure. However, this relaxation comes at the cost of imposing an upper bound on the learning rate, which was not required in Theorem \ref{thm:holder_bcg_conv}. The replacement allows us to choose from a wider variety of learning rate schedules.
\begin{theorem}\label{thm:bvariance_conv}
Set $(\beta_{1,k})_k$ and $(\alpha_k)_k$ satisfying $\bar{\beta}_1\coloneq \sup_{k\in\mathbb{N}}\beta_{1,k}<1$, $\alpha_k \in [0, \alpha_{\max}] \subset [0, \mu/(2L_N\nu^2))$, and $\alpha_{\max}>0$ respectively. Then, under \ref{asm:lowerbounded_objective} for $f_N$, \ref{asmmodi:lipsmooth_gradient}, and \ref{asm:asboundedsg}, a sequence $(\bm{x}_k)_{k\in \mathbb{N}}$ generated by vanilla Adam \ref{algo:adam} satisfies the following properties.
\begin{enumerate}[label=3-(\roman*)]
\item \label{thm:bvariance_upboud}For any $K\in\mathbb{N}$, 
    \begin{align*}
    &\min_{1\leq k\leq K}\mathbb{E}[\|\nabla f_N(\bm{x}_k)\|^2]
    \leq \frac{1}{\sum_{k=1}^{K}\alpha_{k}}\sum_{k=1}^{K}\alpha_{k}\mathbb{E}[\|\nabla f_N(\bm{x}_k)\|^2]\\
    &\leq Q_1\frac{1}{\sum_{k=1}^{K}\alpha_{k}}+Q_2\frac{\sum_{k=1}^{K}\alpha_k\beta_{1,k}^2}{\sum_{k=1}^{K}\alpha_{k}}+Q_3\frac{\sum_{k=1}^{K}\alpha_k/b_k}{\sum_{k=1}^{K}\alpha_{k}}.
    \end{align*}
    where $Q_1=2(f_N(\bm{x}_1)-f_{N,\star})/(\mu-L_N\nu^2\alpha_{\max}), Q_3= 2\nu G_N^2(\mu^{-1}+2L_N\nu\alpha_{\max})/(\mu-L_N\nu^2\alpha_{\max}), Q_2=4Q_3$.
\item If $(\alpha_k), (\beta_{1,k})_k$, and $(b_k)_k$ satisfy $\sum_{k=1}^{\infty}\alpha_k = \infty, \ \sum_{k=1}^{\infty}\beta_{1,k}^2 < \infty$, and $ \sum_{k=1}^{\infty}\alpha_k/b_k < \infty$, then $\min_{1\leq k\leq K}\mathbb{E}[\|\nabla f_N(\bm{x}_k)\|^2]
    =O(1/\sum_{k=1}^{K}\alpha_k)$ holds, which implies $\liminf_{k\to\infty}\mathbb{E}[\|\nabla f_N(\bm{x}_k)\|^2]=0$.
\end{enumerate}
\end{theorem}
The proof of Theorem \ref{thm:bvariance_conv} can be found at Appendix \ref{appdix:prf_variance_conv}. If $G_N$ or $L_{N}$ grows by increasing $N$, the upper bound in Theorem \ref{thm:bvariance_conv} worsens. Since \ref{parprop:hev_poplimit} i.e., heavy-tailed noise can arise by $N\to\infty$, heavy-tailed noise can worsen the upper bound. Furthermore, heavy-tailed noise can reduce the upper bound on the admissible learning rate required for convergence. An example of $\beta_{1,k}$ has already given in Section \ref{sec:gen_case}. An example of $b_k$ is $b_k=b_1\delta^{\lfloor k/E\rfloor}$. Examples of $\alpha_k$ are $\alpha_k =\alpha_{\max}$, the cosine annealing learning rate \citep{loshchilov2018decoupled}, and the polynomial decay learning rate \citep{liu2015parsenet}. Using these schedulers one can easily verify that the convergence rate is $O(1/K)$. In particular, since the cosine annealing learning rate and the polynomial decay learning rate include the total number of steps $\tilde{K}$ in their update rules, the convergence rates become $\mathcal{O}_{\tilde{K}}(1/\tilde{K})$.

\section{Conclusion}\label{sec:conclusion}
In this paper, we investigated how heavy-tailed stochastic gradient noise affects the convergence of vanilla Adam in the expected risk and the empirical risk minimization. For the expected risk minimization, we established convergence guarantees under bounded $\mathfrak{p}$-th moment assumptions and showed that the iteration-dependent convergence term deteriorates as the noise becomes heavier-tailed. We then analyzed the empirical risk minimization and showed that the effects of heavy-tailed population noise can appear through dataset-dependent constants, potentially becoming more pronounced as the dataset size increases. Taken together, these results clarify the relationship between the effects of heavy-tailed noise at the population and finite-dataset levels. Future work includes developing methods to reduce the noise-dependent residual term under heavy-tailed noise and establishing asymptotic and non-asymptotic convergence guarantees for vanilla Adam under weaker assumptions on the full gradient.

\subsubsection*{Acknowledgment}
We thank Kento Imaizumi for their assistance with some of the numerical experiments and for constructive discussions. This work was supported by the Japan Society for the Promotion of Science (JSPS) KAKENHI Grant Number 24K14846 awarded to Hideaki Iiduka.

\bibliography{biblio}
\bibliographystyle{tmlr}

\appendix
\section{Proof of Theorem \ref{thm:heavy_conv}}\label{appdix:prf_heavy_conv}
\begin{lemma}\label{lem:heavy_conv}
Let $\bar{\beta}_1\coloneq \sup_{k\in\mathbb{N}}\beta_{1,k}<1$, $\bar{\beta}_2\coloneq \sup_{k\in\mathbb{N}}\beta_{2,k}<1$, $\mathfrak{p}\in(1,2]$, and $\mathfrak{h}\in(0,1]$. Assume \ref{asm:bounded_gradient} and \ref{asm:boundedpthmoment_sg}. Then, for any $k\in\mathbb{N}$, we obtain
\begin{align*}
&\mathbb{E}_{\xi_k\sim\mathcal{D}}[\|\bm{g}(\bm{x}_k;\xi_k)\|]
\leq(\mathbb{E}_{\xi_k\sim\mathcal{D}}[\|\bm{g}(\bm{x}_k;\xi_k)\|^{\mathfrak{p}}])^{\frac{1}{\mathfrak{p}}}
\leq \sigma+G,\quad \mathbb{E}[\|\hat{\bm{m}}_k\|]\leq \sigma+G,
\quad\mathbb{E}[\|H_k^{-1}\hat{\bm{m}}_k\|]\leq \hat{Q}_2,\\
&\mathbb{E}[\|H_k\|_{\text{op}}^{\mathfrak{p}}]
\leq 2^{\mathfrak{p}-1}\epsilon^{\mathfrak{p}}+\frac{4^{\mathfrak{p}-1}d^{\frac{2-\mathfrak{p}}{2}}(\sigma^{\mathfrak{p}}+G^{\mathfrak{p}})}{1-\bar{\beta}_{2}^{\frac{\mathfrak{p}}{2}}}\exp\left(\frac{\mathfrak{p}\bar{\beta}_2}{2(1-\bar{\beta}_2)^2}\right)\eqcolon\hat{Q}_1,
\end{align*}
where 
\begin{align*}
\hat{Q}_2=
\begin{dcases}
\epsilon^{-1}\sigma+\epsilon^{-1}G & \text{(General case)},\\
\sqrt{\frac{d}{(1-\beta_2)(1-\bar{\beta}_1^2/\beta_2)}}& \text{(Selectable when $\beta_{2,k}=\beta_2$ and $\bar{\beta}_1^2<\beta_2$)}.
\end{dcases}
\end{align*}
Additionally, under $1+\mathfrak{h}\leq\mathfrak{p}$, it holds for any $k\in\mathbb{N}$ that
\begin{align*}
\mathbb{E}[\|\hat{\bm{m}}_k\|^{1+\mathfrak{h}}]
\leq 2^{1+\mathfrak{h}}(\sigma^{1+\mathfrak{h}}+G^{1+\mathfrak{h}}).
\end{align*}
\end{lemma}

\begin{proof}[Proof of Lemma \ref{lem:heavy_conv}]
Jensen's inequality, the Minkowski inequality, and \ref{asm:bounded_gradient} yield
\begin{align*}
\mathbb{E}_{\xi_k}[\|\bm{g}(\bm{x}_k;\xi_k)\|]
&\leq(\mathbb{E}_{\xi_k}[\|\bm{g}(\bm{x}_k;\xi_k)\|^{\mathfrak{p}}])^{\frac{1}{\mathfrak{p}}}
=(\mathbb{E}_{\xi_k}[\|(\bm{g}(\bm{x}_k;\xi_k)-\nabla f(\bm{x}_k))+\nabla f(\bm{x}_k)\|^{\mathfrak{p}}])^{\frac{1}{\mathfrak{p}}}\\
&\leq (\mathbb{E}_{\xi_k}[\|\bm{g}(\bm{x}_k;\xi_k)-\nabla f(\bm{x}_k)\|^{\mathfrak{p}}])^{\frac{1}{\mathfrak{p}}}+(\mathbb{E}_{\xi_k}[\|\nabla f(\bm{x}_k)\|^{\mathfrak{p}}])^{\frac{1}{\mathfrak{p}}}
\leq \sigma+G.
\end{align*}
From the definition of $\bm{m}_k=\beta_{1,k}\bm{m}_{k-1}+(1-\beta_{1,k})\bm{g}(\bm{x}_k;\xi_k),\bm{m}_0=0$ and $\prod_{j=k+1}^k\cdot\coloneq1$, 
\begin{align*}
\bm{m}_k=\left(\prod_{j=1}^k\beta_{1,j}\right)\bm{m}_0+\sum_{j=1}^k (1-\beta_{1,j})\left(\prod_{i=j+1}^k\beta_{1,i}\right)\bm{g}(\bm{x}_j;\xi_j)
=\sum_{j=1}^k (1-\beta_{1,j})\left(\prod_{i=j+1}^k\beta_{1,i}\right)\bm{g}(\bm{x}_j;\xi_j),
\end{align*}
which, together with $\sum_{j=1}^k (1-\beta_{1,j})(\prod_{i=j+1}^k\beta_{1,i})=\sum_{j=1}^k(\prod_{i=j+1}^k\beta_{1,i}-\prod_{i=j}^k\beta_{1,i})=(\prod_{i=k+1}^k\beta_{1,i}-\prod_{i=1}^k\beta_{1,i})=1-\prod_{i=1}^k\beta_{1,i}$ by telescoping, leads to
\begin{align*}
&\mathbb{E}[\|\hat{\bm{m}}_k\|]
\leq \frac{1}{1-\prod_{j=1}^k\beta_{1,j}}\mathbb{E}[\|\bm{m}_k\|]
=\frac{1}{1-\prod_{j=1}^k\beta_{1,j}}\mathbb{E}\left[\left\|\sum_{j=1}^k (1-\beta_{1,j})\left(\prod_{i=j+1}^k\beta_{1,i}\right)\bm{g}(\bm{x}_j;\xi_j)\right\|\right]\\
&\leq \frac{1}{1-\prod_{j=1}^k\beta_{1,j}}\sum_{j=1}^k (1-\beta_{1,j})\left(\prod_{i=j+1}^k\beta_{1,i}\right)\mathbb{E}[\mathbb{E}_{\xi_j}[\left\|\bm{g}(\bm{x}_j;\xi_j)\right\|]]
\leq  \sigma+G.
\end{align*}
Noting $H_k\succeq \epsilon I$ which leads to $\|H_k^{-1}\|_{\text{op}}\leq \epsilon^{-1}$, we have $\mathbb{E}[\|H_k^{-1}\hat{\bm{m}}_k\|]\leq \mathbb{E}[\|H_k^{-1}\|_{\text{op}}\|\hat{\bm{m}}_k\|]\leq \epsilon^{-1}\mathbb{E}[\|\hat{\bm{m}}_k\|]
\leq \epsilon^{-1}\sigma+\epsilon^{-1}G$, where the first inequality is followed by the definition of the operator norm. On the other hand, under $\bar{\beta}_1^2<\beta_2$, another upper bound can be obtained; from the definition of $\bm{v}_k=\beta_{2,k}\bm{v}_{k-1}+(1-\beta_{2,k})\bm{g}(\bm{x}_k;\xi_k) \odot \bm{g}(\bm{x}_k;\xi_k), \bm{v}_0=0$, it holds that
\begin{align*}
\bm{v}_k
&=\left(\prod_{j=1}^k\beta_{2,j}\right)\bm{v}_0+\sum_{j=1}^k (1-\beta_{2,j})\left(\prod_{i=j+1}^k\beta_{2,i}\right)\bm{g}(\bm{x}_j;\xi_j) \odot \bm{g}(\bm{x}_j;\xi_j)
=\sum_{j=1}^k (1-\beta_{2,j})\left(\prod_{i=j+1}^k\beta_{2,i}\right)\bm{g}(\bm{x}_j;\xi_j) \odot \bm{g}(\bm{x}_j;\xi_j).
\end{align*}
Denoting $\bm{g}(\bm{x}_j;\xi_j)=(g_{1,j},\ldots,g_{d,j})^{\top}, \bm{m}_k=(m_{1,k},\ldots,m_{d,k})^{\top}, \hat{\bm{m}}_k=(\hat{m}_{1,k},\ldots,\hat{m}_{d,k})^{\top}, \bm{v}_k=(v_{1,k},\ldots,v_{d,k})^{\top}$, and $\hat{\bm{v}}_k=(\hat{v}_{1,k},\ldots,\hat{v}_{d,k})^{\top}$, from the Cauchy-Schwarz inequality, we have 
\begin{align*}
\left(\frac{|\hat{m}_{i,k}|}{\sqrt{\hat{v}_{i,k}}+\epsilon}\right)^2
&\leq \left(\frac{|\hat{m}_{i,k}|}{\sqrt{\hat{v}_{i,k}}}\right)^2
=\frac{\hat{m}_{i,k}^2}{\hat{v}_{i,k}}
=\frac{1}{\hat{v}_{i,k}}\left\{\sum_{j=1}^k (1-\beta_{1,j})\left(\prod_{l=j+1}^k\beta_{1,l}\right)g_{j,k}\right\}^2\\
&=\frac{1}{\hat{v}_{i,k}}\left\{\sum_{j=1}^k \frac{(1-\beta_{1,j})\prod_{l=j+1}^k\beta_{1,l}}{\sqrt{(1-\beta_{2,j})\prod_{l=j+1}^k\beta_{2,l}}}\sqrt{ (1-\beta_{2,j})\prod_{l=j+1}^k\beta_{2,l}}g_{j,k}\right\}^2\\
&\leq\frac{1}{\hat{v}_{i,k}}\left\{\sum_{j=1}^k \frac{(1-\beta_{1,j})^2\prod_{l=j+1}^k\beta_{1,l}^2}{(1-\beta_{2,j})\prod_{l=j+1}^k\beta_{2,l}}\right\}\left\{\sum_{j=1}^k (1-\beta_{2,j})\left(\prod_{l=j+1}^k\beta_{2,l}\right)g_{j,k}^2\right\}\\
&=\sum_{j=1}^k \frac{(1-\beta_{1,j})^2}{(1-\beta_{2,j})}\prod_{l=j+1}^k\frac{\beta_{1,l}^2}{\beta_{2,l}}
\leq \frac{1}{1-\bar{\beta}_{2}}\sum_{j=1}^k \prod_{l=j+1}^k\frac{\beta_{1,l}^2}{\beta_{2,l}}
\leq \frac{1}{1-\bar{\beta}_{2}}\sum_{j=1}^k \prod_{l=j+1}^k\frac{\bar{\beta}_{1}^2}{\beta_{2,l}}.
\end{align*}
When one uses $\beta_{2,k}=\beta_2$, under $\bar{\beta}_1^2<\beta_2$, it follows from $\sum_{j=1}^k \prod_{l=j+1}^k\bar{\beta}_{1}^2/\beta_{2}= \sum_{j=1}^k (\bar{\beta}_{1}^2/\beta_{2})^{k-j}\leq (\bar{\beta}_{1}^2/\beta_{2})^{k-1}/(1-\bar{\beta}_{1}^2/\beta_{2})<1/(1-\bar{\beta}_{1}^2/\beta_{2})$ that
\begin{align*}
\|H_k^{-1}\hat{\bm{m}}_k\|
=\sqrt{\sum_{i=1}^d\left(\frac{|\hat{m}_{i,k}|}{\sqrt{\hat{v}_{i,k}}+\epsilon}\right)^2}
\leq \sqrt{\frac{d}{(1-\beta_2)(1-\bar{\beta}_1^2/\beta_2)}}.
\end{align*}
On the other hand, since $\|\bm{x}\|_p\leq d^{1/p-1/2}\|\bm{x}\|_2$ holds for $\bm{x}\in\mathbb{R}^d, p\in(1,2]$, we have $\mathbb{E}[\|\bm{g}(\bm{x}_k;\xi_k)\|_{\mathfrak{p}}^{\mathfrak{p}}]\leq d^{1-\mathfrak{p}/2}\mathbb{E}[\|\bm{g}(\bm{x}_k;\xi_k)\|_{2}^{\mathfrak{p}}]$. From this and $\hat{v}_{i,k}^{\mathfrak{p}/2}= (1-\prod_{j=1}^k\beta_{2,j})^{-\mathfrak{p}/2}v_{i,k}^{\mathfrak{p}/2}\leq (1-\prod_{j=1}^k\beta_{2,j})^{-\mathfrak{p}/2}\{\beta_{2,k}^{\mathfrak{p}/2}v_{i,k-1}^{\mathfrak{p}/2}+(1-\beta_{2,k})^{\mathfrak{p}/2}g_{i,k}^{\mathfrak{p}}\}$ yields
\begin{align*}
&\mathbb{E}\left[\sum_{i=1}^d\hat{v}_{i,k}^{\frac{\mathfrak{p}}{2}}\right]
\leq \frac{1}{(1-\prod_{j=1}^k\beta_{2,j})^{\frac{\mathfrak{p}}{2}}}\left\{\beta_{2,k}^{\frac{\mathfrak{p}}{2}}\mathbb{E}\left[\sum_{i=1}^dv_{i,k-1}^{\frac{\mathfrak{p}}{2}}\right]+(1-\beta_{2,k})^{\frac{\mathfrak{p}}{2}}\mathbb{E}[\|\bm{g}(\bm{x}_k;\xi_k)\|_{\mathfrak{p}}^{\mathfrak{p}}]\right\}\\
&\leq\frac{\beta_{2,k}^{\frac{\mathfrak{p}}{2}}}{(1-\prod_{j=1}^k\beta_{2,j})^{\frac{\mathfrak{p}}{2}}}\mathbb{E}\left[\sum_{i=1}^dv_{i,k-1}^{\frac{\mathfrak{p}}{2}}\right]+\frac{(1-\beta_{2,k})^{\frac{\mathfrak{p}}{2}}d^{\frac{2-\mathfrak{p}}{2}}}{(1-\prod_{j=1}^k\beta_{2,j})^{\frac{\mathfrak{p}}{2}}}\mathbb{E}[\|\bm{g}(\bm{x}_k;\xi_k)\|^{\mathfrak{p}}]\\
&\leq \frac{\beta_{2,k}^{\frac{\mathfrak{p}}{2}}}{(1-\prod_{j=1}^k\beta_{2,j})^{\frac{\mathfrak{p}}{2}}}\mathbb{E}\left[\sum_{i=1}^dv_{i,k-1}^{\frac{\mathfrak{p}}{2}}\right]+\frac{(1-\beta_{2,k})^{\frac{\mathfrak{p}}{2}}d^{\frac{2-\mathfrak{p}}{2}}}{(1-\prod_{j=1}^k\beta_{2,j})^{\frac{\mathfrak{p}}{2}}}\left(2^{\mathfrak{p}-1}\sigma^{\mathfrak{p}}+2^{\mathfrak{p}-1}G^{\mathfrak{p}}\right)\\
&\leq \prod_{j=1}^k\frac{\beta_{2,j}^{\frac{\mathfrak{p}}{2}}}{(1-\prod_{i=1}^j\beta_{2,j})^{\frac{\mathfrak{p}}{2}}}\sum_{i=1}^dv_{i,0}^{\frac{\mathfrak{p}}{2}}+2^{\mathfrak{p}-1}d^{\frac{2-\mathfrak{p}}{2}}\left(\sigma^{\mathfrak{p}}+G^{\mathfrak{p}}\right)\sum_{j=1}^k\left(\prod_{i=j+1}^k\frac{\beta_{2,i}^{\frac{\mathfrak{p}}{2}}}{(1-\prod_{l=1}^i\beta_{2,l})^{\frac{\mathfrak{p}}{2}}}\right)\frac{(1-\beta_{2,j})^{\frac{\mathfrak{p}}{2}}}{(1-\prod_{i=1}^j\beta_{2,i})^{\frac{\mathfrak{p}}{2}}}\\
&=2^{\mathfrak{p}-1}d^{\frac{2-\mathfrak{p}}{2}}\left(\sigma^{\mathfrak{p}}+G^{\mathfrak{p}}\right)\sum_{j=1}^k\left(\prod_{i=j+1}^k\frac{\beta_{2,i}^{\frac{\mathfrak{p}}{2}}}{(1-\prod_{l=1}^i\beta_{2,l})^{\frac{\mathfrak{p}}{2}}}\right)\left(\frac{1-\beta_{2,j}}{1-\prod_{i=1}^j\beta_{2,i}}\right)^{\frac{\mathfrak{p}}{2}}\\
&\leq \frac{2^{\mathfrak{p}-1}d^{\frac{2-\mathfrak{p}}{2}}}{1-\bar{\beta}_{2}^{\frac{\mathfrak{p}}{2}}}\left(\sigma^{\mathfrak{p}}+G^{\mathfrak{p}}\right)\exp\left(\frac{\mathfrak{p}\bar{\beta}_2}{2(1-\bar{\beta}_2)^2}\right),
\end{align*}
where we considered
\begin{align*}
&\sum_{j=1}^k\left(\prod_{i=j+1}^k\frac{\beta_{2,i}^{\frac{\mathfrak{p}}{2}}}{(1-\prod_{l=1}^i\beta_{2,l})^{\frac{\mathfrak{p}}{2}}}\right)\left(\frac{1-\beta_{2,j}}{1-\prod_{i=1}^j\beta_{2,i}}\right)^{\frac{\mathfrak{p}}{2}}
\leq\sum_{j=1}^k\prod_{i=j+1}^k\frac{\beta_{2,i}^{\frac{\mathfrak{p}}{2}}}{(1-\prod_{l=1}^i\beta_{2,l})^{\frac{\mathfrak{p}}{2}}}
\leq \sum_{j=1}^k\bar{\beta}_{2}^{\frac{\mathfrak{p}}{2}(k-j)}\prod_{i=j+1}^k\frac{1}{(1-\bar{\beta}_{2}^i)^{\frac{\mathfrak{p}}{2}}}\\
&\quad \leq \sum_{j=0}^{\infty}\bar{\beta}_{2}^{\frac{\mathfrak{p}}{2}j}\prod_{i=2}^{\infty}\frac{1}{(1-\bar{\beta}_{2}^i)^{\frac{\mathfrak{p}}{2}}}
=\frac{1}{1-\bar{\beta}_{2}^{\frac{\mathfrak{p}}{2}}}\exp\left(-\frac{\mathfrak{p}}{2}\sum_{i=2}^{\infty}\log(1-\bar{\beta}_{2}^i)\right)
\leq \frac{1}{1-\bar{\beta}_{2}^{\frac{\mathfrak{p}}{2}}}\exp\left(\frac{\mathfrak{p}}{2}\sum_{i=2}^{\infty}\frac{\bar{\beta}_2^i}{1-\bar{\beta}_2^i}\right)\\
&\quad\leq \frac{1}{1-\bar{\beta}_{2}^{\frac{\mathfrak{p}}{2}}}\exp\left(\frac{\mathfrak{p}}{2(1-\bar{\beta}_2)}\sum_{i=1}^{\infty}\bar{\beta}_2^i\right)
=\frac{1}{1-\bar{\beta}_{2}^{\frac{\mathfrak{p}}{2}}}\exp\left(\frac{\mathfrak{p}\bar{\beta}_2}{2(1-\bar{\beta}_2)^2}\right)
\end{align*}
due to $1-\beta_{2,j}\leq 1-\prod_{i=1}^j\beta_{2,i}$ and $-\log(1-x)\leq x/(1-x)$. Therefore, we obtain
\begin{align*}
\mathbb{E}[\|H_k\|_{\text{op}}^{\mathfrak{p}}]
&=\mathbb{E}\left[\left(\max_{1\leq j\leq d}\{\sqrt{\hat{v}_{j,k}}+\epsilon\}\right)^{\mathfrak{p}}\right]
\leq 2^{\mathfrak{p}-1}\left(\mathbb{E}\left[\max_{1\leq j\leq d}\hat{v}_{j,k}^{\frac{\mathfrak{p}}{2}}\right]+\epsilon^{\mathfrak{p}}\right)
\leq 2^{\mathfrak{p}-1}\left(\mathbb{E}\left[\sum_{i=1}^d\hat{v}_{i,k}^{\frac{\mathfrak{p}}{2}}\right]+\epsilon^{\mathfrak{p}}\right)\\
&\leq 2^{\mathfrak{p}-1}\epsilon^{\mathfrak{p}}+\frac{4^{\mathfrak{p}-1}d^{\frac{2-\mathfrak{p}}{2}}(\sigma^{\mathfrak{p}}+G^{\mathfrak{p}})}{1-\bar{\beta}_{2}^{\frac{\mathfrak{p}}{2}}}\exp\left(\frac{\mathfrak{p}\bar{\beta}_2}{2(1-\bar{\beta}_2)^2}\right).
\end{align*}
Finally, we estimate under $1+\mathfrak{h}\leq\mathfrak{p}$: since $\|\cdot\|^{\frac{1+\mathfrak{h}}{\mathfrak{p}}}$ is concave, Jensen's inequality yields
\begin{align*}
\mathbb{E}_{\xi_k}[\|\bm{g}(\bm{x}_k;\xi_k)\|^{1+\mathfrak{h}}]
=\mathbb{E}_{\xi_k}[\|\bm{g}(\bm{x}_k;\xi_k)\|^{\mathfrak{p}\frac{1+\mathfrak{h}}{\mathfrak{p}}}]
\leq \mathbb{E}_{\xi_k}[\|\bm{g}(\bm{x}_k;\xi_k)\|^{\mathfrak{p}}]^{\frac{1+\mathfrak{h}}{\mathfrak{p}}}
\leq \left(2^{\mathfrak{p}-1}\sigma^{\mathfrak{p}}+2^{\mathfrak{p}-1}G^{\mathfrak{p}}\right)^{\frac{1+\mathfrak{h}}{\mathfrak{p}}}
\leq 2^{1+\mathfrak{h}}(\sigma^{1+\mathfrak{h}}+G^{1+\mathfrak{h}}).
\end{align*}
Hence, the estimation of $\mathbb{E}[\|\hat{\bm{m}}_k\|]^{1+\mathfrak{h}}]$ is followed from
\begin{align*}
&\left(\mathbb{E}[\|\hat{\bm{m}}_k\|]^{1+\mathfrak{h}}]\right)^{\frac{1}{1+\mathfrak{h}}}
= \frac{1}{1-\prod_{j=1}^k\beta_{1,j}}\left(\mathbb{E}[\|\bm{m}_k\|^{1+\mathfrak{h}}]\right)^{\frac{1}{1+\mathfrak{h}}}\\
&=\frac{1}{1-\prod_{j=1}^k\beta_{1,j}}\left(\mathbb{E}\left[\left\|\sum_{j=1}^k (1-\beta_{1,j})\left(\prod_{i=j+1}^k\beta_{1,i}\right)\bm{g}(\bm{x}_j;\xi_j)\right\|^{1+\mathfrak{h}}\right]\right)^{\frac{1}{1+\mathfrak{h}}}\\
&\leq \frac{1}{1-\prod_{j=1}^k\beta_{1,j}}\sum_{j=1}^k (1-\beta_{1,j})\left(\prod_{i=j+1}^k\beta_{1,i}\right)\left(\mathbb{E}\left[\mathbb{E}_{\xi_j}\left[\left\|\bm{g}(\bm{x}_j;\xi_j)\right\|^{1+\mathfrak{h}}\right]\right]\right)^{\frac{1}{1+\mathfrak{h}}}
\leq 2(\sigma^{1+\mathfrak{h}}+G^{1+\mathfrak{h}})^{\frac{1}{1+\mathfrak{h}}}.
\end{align*}
\end{proof}

\begin{proof}[Proof of Theorem \ref{thm:heavy_conv}]
\ref{asm:holdersmooth_gradient} and the update rule of Algorithm \ref{algo:adam} yield
\begin{align*}
f(\bm{x}_{k+1}) 
&\leq f(\bm{x}_k) + \left\langle \nabla f(\bm{x}_k), \bm{x}_{k+1} - \bm{x}_k\right\rangle + \frac{L_{\mathfrak{h}}}{1+\mathfrak{h}}\| \bm{x}_{k+1} - \bm{x}_k \|^{1+\mathfrak{h}}\\
&\leq f(\bm{x}_k) - \alpha_k\left\langle \nabla f(\bm{x}_k), H_k^{-1}\hat{\bm{m}}_k\right\rangle + \frac{\alpha_k^{1+\mathfrak{h}}L_{\mathfrak{h}}}{1+\mathfrak{h}}\left\| H_k^{-1}\hat{\bm{m}}_k \right\|^{1+\mathfrak{h}} \\
&= f(\bm{x}_k) + \underbrace{\left\langle \nabla f(\bm{x}_k), - \alpha_{k-1}H_{k-1}^{-1}\hat{\bm{m}}_k\right\rangle}_{X_1} + \underbrace{\left\langle \nabla f(\bm{x}_k), (\alpha_{k-1}H_{k-1}^{-1} - \alpha_kH_k^{-1})\hat{\bm{m}}_k\right\rangle}_{X_2}+ \frac{\alpha_k^{1+\mathfrak{h}}L_{\mathfrak{h}}}{1+\mathfrak{h}}\underbrace{\left\| H_k^{-1}\hat{\bm{m}}_k \right\|^{1+\mathfrak{h}}}_{X_3}.
\end{align*}
$\|\nabla f(\bm{x}_k)\|^r=\{(\|\nabla f(\bm{x}_k)\|^2/\|H_{k-1}\|_{\text{op}})\|H_{k-1}\|_{\text{op}}\}^{r/2}=(\|\nabla f(\bm{x}_k)\|^2/\|H_{k-1}\|_{\text{op}})^{r/2}(\|H_{k-1}\|_{\text{op}}^{r/(2-r)})^{(2-r)/2}$ holds for $r\in\mathbb{R}$, and $h_1\coloneq 2/r$ and $h_2\coloneq 2/(2-r)$ satisfy $1/h_2+1/h_2=1$. By setting $r=2\mathfrak{p}/(1+\mathfrak{p})$ for satisfying $r/(2-r)=\mathfrak{p}$, $h_1$ and $h_2$ become h\"older conjugates due to $h_1=2/r=(1+\mathfrak{p})/\mathfrak{p}=1+1/\mathfrak{p}>1$ and $h_2=2/(2-r)=1+\mathfrak{p}>1$. Thus, H\"older's inequality yields
\begin{align*}
\mathbb{E}\left[\|\nabla f(\bm{x}_k)\|^{\frac{2\mathfrak{p}}{1+\mathfrak{p}}}\right]
&=\mathbb{E}\left[\|\nabla f(\bm{x}_k)\|^{r}\right]
=\mathbb{E}\left[\left(\frac{\|\nabla f(\bm{x}_k)\|^2}{\|H_{k-1}\|_{\text{op}}}\right)^{\frac{1}{h_1}}\left(\|H_{k-1}\|_{\text{op}}^{\mathfrak{p}}\right)^{\frac{1}{h_2}}\right]
\leq \mathbb{E}\left[\frac{\|\nabla f(\bm{x}_k)\|^2}{\|H_{k-1}\|_{\text{op}}}\right]^{\frac{1}{h_1}}\mathbb{E}\left[\|H_{k-1}\|_{\text{op}}^{\mathfrak{p}}\right]^{\frac{1}{h_2}}\\
&=\mathbb{E}\left[\frac{\|\nabla f(\bm{x}_k)\|^2}{\|H_{k-1}\|_{\text{op}}}\right]^{\frac{\mathfrak{p}}{1+\mathfrak{p}}}\mathbb{E}\left[\|H_{k-1}\|_{\text{op}}^{\mathfrak{p}}\right]^{\frac{1}{1+\mathfrak{p}}},
\end{align*}
which implies $\mathbb{E}[\|\nabla f(\bm{x}_k)\|^{2\mathfrak{p}/(1+\mathfrak{p})}]^{(1+\mathfrak{p})/\mathfrak{p}}\leq \mathbb{E}[\|\nabla f(\bm{x}_k)\|^2/\|H_{k-1}\|_{\text{op}}]\mathbb{E}[\|H_{k-1}\|_{\text{op}}^{\mathfrak{p}}]^{\mathfrak{p}}$. Thus, from Lemma \ref{lem:heavy_conv} and $\langle\nabla f(\bm{x}_k), H_{k-1}^{-1}\nabla f(\bm{x}_k)\rangle\geq \|\nabla f(\bm{x}_k)\|^2/\|H_{k-1}\|_{\text{op}}$ due to $H_{k-1}\succeq \|H_{k-1}\|_{\text{op}}I\succeq\epsilon I$, we have
\begin{align*}
\mathbb{E}[\langle\nabla f(\bm{x}_k), H_{k-1}^{-1}\nabla f(\bm{x}_k)\rangle]
&\geq \mathbb{E}\left[\frac{\|\nabla f(\bm{x}_k)\|^2}{\|H_{k-1}\|_{\text{op}}}\right]
\geq \frac{\mathbb{E}\left[\|\nabla f(\bm{x}_k)\|^{\frac{2\mathfrak{p}}{1+\mathfrak{p}}}\right]^{\frac{1+\mathfrak{p}}{\mathfrak{p}}}}{\mathbb{E}\left[\|H_{k-1}\|_{\text{op}}^{\mathfrak{p}}\right]^{\mathfrak{p}}}
\geq \frac{1}{\hat{Q}_1}\mathbb{E}\left[\|\nabla f(\bm{x}_k)\|^{\frac{2\mathfrak{p}}{1+\mathfrak{p}}}\right]^{\frac{1+\mathfrak{p}}{\mathfrak{p}}}\\
&=\frac{1}{\hat{Q}_1}\mathbb{E}\left[\frac{\|\nabla f(\bm{x}_k)\|^{2}}{\|\nabla f(\bm{x}_k)\|^{\frac{2}{1+\mathfrak{p}}}}\right]^{\frac{1+\mathfrak{p}}{\mathfrak{p}}}
\geq \frac{1}{\hat{Q}_1G^{\frac{2}{1+\mathfrak{p}}}}\mathbb{E}\left[\|\nabla f(\bm{x}_k)\|^{2}\right]^{\frac{1+\mathfrak{p}}{\mathfrak{p}}}.
\end{align*}
By taking the expectation of $X_1$, the above estimation and Lemma \ref{lem:heavy_conv} yield
\begin{align*}
\mathbb{E}[X_1] 
&= \frac{1-\beta_{1,k}}{1-\prod_{j=1}^k\beta_{1,j}}\mathbb{E}[\left\langle\nabla f(\bm{x}_k), -\alpha_{k-1}H_{k-1}^{-1}\mathbb{E}_{\xi_k}[\bm{g}(\bm{x}_k;\xi_k)]\right\rangle] + \beta_{1,k}\mathbb{E}\left[\left\langle\nabla f(\bm{x}_k), -\alpha_{k-1}H_{k-1}^{-1}\frac{\bm{m}_{k-1}}{1-\prod_{j=1}^k\beta_{1,j}}\right\rangle\right] \\
&= \frac{1-\beta_{1,k}}{1-\prod_{j=1}^k\beta_{1,j}}\mathbb{E}[\left\langle\nabla f(\bm{x}_k), -\alpha_{k-1}H_{k-1}^{-1}\nabla f(\bm{x}_k)\right\rangle] + \beta_{1,k}\mathbb{E}[\left\langle\nabla f(\bm{x}_k), -\alpha_{k-1}H_{k-1}^{-1}\hat{\bm{m}}_{k-1}\right\rangle] \\
&\leq -\frac{1-\beta_{1,k}}{1-\prod_{j=1}^k\beta_{1,j}}\frac{\alpha_{k-1}}{\hat{Q}_1G^{\frac{2}{1+\mathfrak{p}}}}\mathbb{E}[\|\nabla f(\bm{x}_k)\|^2]^{\frac{1+\mathfrak{p}}{\mathfrak{p}}} + \alpha_{k-1}\beta_{1,k}\mathbb{E}[\|\nabla f(\bm{x}_k)\|\|H_{k-1}^{-1}\hat{\bm{m}}_{k-1}\|] \\
&\leq -\frac{(1-\bar{\beta_{1}})\alpha_{k}}{\hat{Q}_1G^{\frac{2}{1+\mathfrak{p}}}}\mathbb{E}[\|\nabla f(\bm{x}_k)\|^2]^{\frac{1+\mathfrak{p}}{\mathfrak{p}}} +\alpha_{\max}G\beta_{1,k}\mathbb{E}[\|H_{k-1}^{-1}\hat{\bm{m}}_{k-1}\|]
\leq -\frac{(1-\bar{\beta_{1}})\alpha_{k}}{\hat{Q}_1G^{\frac{2}{1+\mathfrak{p}}}}\mathbb{E}[\|\nabla f(\bm{x}_k)\|^2]^{\frac{1+\mathfrak{p}}{\mathfrak{p}}} + \alpha_{\max}G\hat{Q}_2\beta_{1,k},
\end{align*}
where we considered $1-\bar{\beta}_1\leq 1 - \beta_{1,k} \leq 1 - \prod_{j=1}^k\beta_{1,j} \leq 1$. From Lemma \ref{lem:heavy_conv}, $H_k\succeq\epsilon I$, and $\mathfrak{h}\in(0,1]$, $X_3$ can be estimated as
\begin{align*}
\mathbb{E}[X_3]
\leq \mathbb{E}[\|H_k^{-1}\hat{\bm{m}}_k\|^{1+\mathfrak{h}}]
\leq \mathbb{E}[\|H_k^{-1}\|_{\text{op}}^{1+\mathfrak{h}}\|\hat{\bm{m}}_k\|^{1+\mathfrak{h}}]
\leq \epsilon^{-(1+\mathfrak{h})}\mathbb{E}[\|\hat{\bm{m}}_k\|^{1+\mathfrak{h}}]
\leq 2^{1+\mathfrak{h}}\epsilon^{-(1+\mathfrak{h})}\sigma^{1+\mathfrak{h}}+2^{1+\mathfrak{h}}\epsilon^{-(1+\mathfrak{h})}G^{1+\mathfrak{h}}.
\end{align*}
We proceed to estimate $\sum_{k=1}^K\mathbb{E}[X_2]$; Lemma \ref{lem:heavy_conv}, \ref{asm:holdersmooth_gradient}, Jensen's inequality, and $H_K^{-1}\succ O$ yield
\begin{align*}
&\sum_{k=1}^K\mathbb{E}[\left\langle \nabla f(\bm{x}_k), (\alpha_{k-1}H_{k-1}^{-1} - \alpha_kH_k^{-1})\nabla f(\bm{x}_k)\right\rangle]\\
&=\left\langle \nabla f(\bm{x}_0), \alpha_{0}H_0^{-1}\nabla f(\bm{x}_0)\right\rangle +\sum_{k=1}^{K-1}\mathbb{E}[\langle \nabla f(\bm{x}_k), \alpha_{k}H_{k}^{-1}\nabla f(\bm{x}_{k+1})\rangle]-\sum_{k=1}^{K-1}\mathbb{E}[\langle \nabla f(\bm{x}_k), \alpha_{k}H_{k}^{-1}\nabla f(\bm{x}_{k})\rangle]\\
&\quad-\langle \nabla f(\bm{x}_K), \alpha_{K}H_{K}^{-1}\nabla f(\bm{x}_{K})\rangle\\
&=\sum_{k=1}^{K-1}\alpha_{k}\mathbb{E}[\langle \nabla f(\bm{x}_k), H_{k}^{-1}(\nabla f(\bm{x}_{k+1})-\nabla f(\bm{x}_{k}))\rangle]
\leq G\sum_{k=1}^{K-1}\alpha_{k}\mathbb{E}[\|H_{k}^{-1}\|_{\text{op}}\|(\nabla f(\bm{x}_{k+1})-\nabla f(\bm{x}_{k}))\|]\\
&\leq \frac{GL_{\mathfrak{h}}}{\epsilon}\sum_{k=1}^{K-1}\alpha_{k}\mathbb{E}[\|\bm{x}_{k+1}-\bm{x}_{k}\|^{\mathfrak{h}}]
=\frac{GL_{\mathfrak{h}}}{\epsilon}\sum_{k=1}^{K-1}\alpha_{k}^{1+\mathfrak{h}}\mathbb{E}[\|H_k^{-1}\hat{\bm{m}}_{k}\|^{\mathfrak{h}}]
\leq \frac{GL_{\mathfrak{h}}}{\epsilon}\sum_{k=1}^{K-1}\alpha_{k}^{1+\mathfrak{h}}\mathbb{E}[\|H_k^{-1}\hat{\bm{m}}_{k}\|]^{\mathfrak{h}}
\leq \frac{GL_{\mathfrak{h}}\hat{Q}_2^{\mathfrak{h}}}{\epsilon}\sum_{k=1}^{K-1}\alpha_{k}^{1+\mathfrak{h}}.
\end{align*}
From $\|\alpha_{k-1}H_{k-1}^{-1}-\alpha_{k}H_{k}^{-1}\|_{\text{op}}\leq \alpha_{k-1}\|H_{k-1}^{-1}\|_{\text{op}}+\alpha_{k}\|H_{k}^{-1}\|_{\text{op}}\leq (\alpha_{k-1}+\alpha_k)/\epsilon$, the Cauchy-Schwarz inequality, Jensen's inequality, Lemma \ref{lem:heavy_conv}, and \ref{asm:holdersmooth_gradient}, we obtain
\begin{align*}
&\mathbb{E}[\left\langle \nabla f(\bm{x}_k), (\alpha_{k-1}H_{k-1}^{-1} - \alpha_kH_k^{-1})(\hat{\bm{m}}_{k-1}-\nabla f(\bm{x}_{k}))\right\rangle]\\
&\leq \frac{G(\alpha_{k-1}+\alpha_k)}{\epsilon}\{\mathbb{E}[\|\hat{\bm{m}}_{k-1}-\nabla f(\bm{x}_{k-1})\|] +\mathbb{E}[\|\nabla f(\bm{x}_{k})-\nabla f(\bm{x}_{k-1})\|]\}\\
&\leq \frac{G(\alpha_{k-1}+\alpha_k)}{\epsilon}\{\mathbb{E}[\|\hat{\bm{m}}_{k-1}\|]+G +L_{\mathfrak{h}}\mathbb{E}[\|\bm{x}_{k}-\bm{x}_{k-1}\|^{\mathfrak{h}}]\}
=\frac{G(\alpha_{k-1}+\alpha_k)}{\epsilon}\{\mathbb{E}[\|\hat{\bm{m}}_{k-1}\|]+G +L_{\mathfrak{h}}\alpha_{k-1}^{\mathfrak{h}}\mathbb{E}[\|H_{k-1}^{-1}\hat{\bm{m}}_{k-1}\|^{\mathfrak{h}}]\}\\
&\leq \frac{G(\alpha_{k-1}+\alpha_k)}{\epsilon}\left\{\sigma+2G +L_{\mathfrak{h}}\alpha_{k-1}^{\mathfrak{h}}\mathbb{E}[\|H_{k-1}^{-1}\hat{\bm{m}}_{k-1}\|]^{\mathfrak{h}}\right\}
\leq \frac{2G\alpha_{\max}(\sigma+2G+L_{\mathfrak{h}}\hat{Q}_2^{\mathfrak{h}}\alpha_{\max}^{\mathfrak{h}})}{\epsilon}.
\end{align*}
The component-wise definition of $\hat{\bm{v}}_{k}$
\begin{align*}
\hat{v}_{k,i}
&=\frac{1 - \beta_{2,k}}{1-\prod_{j=1}^k{\beta}_{2, j}}g_{k,i}^2 + \frac{\beta_{2,k}}{1-\prod_{j=1}^k{\beta}_{2, j}}v_{k-1,i}
=\frac{1 - \beta_{2,k}}{1-\prod_{j=1}^k{\beta}_{2, j}}g_{k,i}^2 + \frac{\beta_{2,k}(1-\prod_{j=1}^{k-1}{\beta}_{2, j})}{1-\prod_{j=1}^k{\beta}_{2, j}}\hat{v}_{k-1,i},
\end{align*}
for $a,b\geq0$, $|(a+\epsilon)^{-1}-(b+\epsilon)^{-1}|=|a-b|/\{(a+\epsilon)(b+\epsilon)\}\leq \min\{\epsilon^{-1},|a-b|/\epsilon^2\},\ |\sqrt{a+b}-\sqrt{a}|\leq \sqrt{b}$, and $\min\{a,b\}\leq a^{2-\mathfrak{p}}b^{\mathfrak{p}-1}$ yield
\begin{align*}
&\left|(\sqrt{\hat{v}_{k,i}}+\epsilon)^{-1}-(\sqrt{\frac{\beta_{2,k}(1-\prod_{j=1}^{k-1}{\beta}_{2, j})}{1-\prod_{j=1}^k{\beta}_{2, j}}\hat{v}_{k-1,i}}+\epsilon)^{-1}\right|
\leq \min\left\{\frac{1}{\epsilon},\frac{1}{\epsilon^2}\left|\sqrt{\hat{v}_{k,i}}-\sqrt{\frac{\beta_{2,k}(1-\prod_{j=1}^{k-1}{\beta}_{2, j})}{1-\prod_{j=1}^k{\beta}_{2, j}}\hat{v}_{k-1,i}}\right|\right\}\\
&\leq \min\left\{\frac{1}{\epsilon},\frac{|g_{k,i}|}{\epsilon^{2}}\sqrt{\frac{1 - \beta_{2,k}}{1-\prod_{j=1}^k{\beta}_{2, j}}}\right\}
=\frac{1}{\epsilon^{2-\mathfrak{p}}}\frac{|g_{k,i}|^{\mathfrak{p}-1}}{\epsilon^{2\mathfrak{p}-2}}\left(\frac{1 - \beta_{2,k}}{1-\prod_{j=1}^k{\beta}_{2, j}}\right)^{(\mathfrak{p}-1)/2}
\leq\frac{|g_{k,i}|^{\mathfrak{p}-1}}{\epsilon^{\mathfrak{p}}(1-\beta_{2,1})^{(\mathfrak{p}-1)/2}}(1 - \beta_{2,k})^{\mathfrak{h}/2}.
\end{align*}
Reducing the fractions to a common denominator and considering $1-a\leq\sqrt{1-a}$ for $0\leq a\leq 1$ lead to
\begin{align*}
&\left|(\sqrt{\frac{\beta_{2,k}(1-\prod_{j=1}^{k-1}{\beta}_{2, j})}{1-\prod_{j=1}^k{\beta}_{2, j}}\hat{v}_{k-1,i}}+\epsilon)^{-1}-(\sqrt{\hat{v}_{k-1,i}}+\epsilon)^{-1}\right|\\
&=\left(1-\sqrt{\frac{\beta_{2,k}(1-\prod_{j=1}^{k-1}{\beta}_{2, j})}{1-\prod_{j=1}^k{\beta}_{2, j}}}\right)\sqrt{\hat{v}_{k-1,i}}(\sqrt{\frac{\beta_{2,k}(1-\prod_{j=1}^{k-1}{\beta}_{2, j})}{1-\prod_{j=1}^k{\beta}_{2, j}}\hat{v}_{k-1,i}}+\epsilon)^{-1}(\sqrt{\hat{v}_{k-1,i}}+\epsilon)^{-1}\\
&\leq \frac{1}{\epsilon}\left(1-\sqrt{\frac{\beta_{2,k}(1-\prod_{j=1}^{k-1}{\beta}_{2, j})}{1-\prod_{j=1}^k{\beta}_{2, j}}}\right)
=\frac{1}{\epsilon}\left(1-\sqrt{1-\frac{1-{\beta}_{2, k}}{1-\prod_{j=1}^k{\beta}_{2, j}}}\right)
\leq \frac{1}{\epsilon}\frac{1-{\beta}_{2, k}}{1-\prod_{j=1}^k{\beta}_{2, j}}
\leq \frac{1}{\epsilon(1-\beta_{2,1})^{(\mathfrak{p}-1)/2}}(1-\beta_{2,k})^{\mathfrak{h}/2}
\end{align*}
Considering the triangle inequality, the monotonicity of $\ell^p$-norm, and the above evaluations, we have
\begin{align*}
\|H_k^{-1}-H_{k-1}^{-1}\|_{\text{op}}
&=\max_{i\in\mathbb{N}_d}|(\sqrt{\hat{v}_{k,i}}+\epsilon)^{-1}-(\sqrt{\hat{v}_{k-1,i}}+\epsilon)^{-1}|
\leq \frac{\|\bm{g}(\bm{x}_k;\xi_k)\|^{\mathfrak{p}-1}(\epsilon^{-\mathfrak{p}}+\epsilon^{-1})}{(1-\beta_{2,1})^{(\mathfrak{p}-1)/2}}(1 - \beta_{2,k})^{\mathfrak{h}/2}
\end{align*}
Since $\xi_k$ is independent of $H_{k-1}^{-1}$ and $\nabla f(\bm{x}_k)$, \ref{asm:unbiasedestimator_sg} yields 
\begin{align}\label{eq:thm1_delete_k-1}
&\mathbb{E}[\langle \nabla f(\bm{x}_k), \alpha_{k-1}H_{k-1}^{-1}(\bm{g}(\bm{x}_k;\xi_k)-\nabla f(\bm{x}_k))\rangle]
=\mathbb{E}\mathbb{E}_{\xi_k}[\langle \nabla f(\bm{x}_k), \alpha_{k-1}H_{k-1}^{-1}(\bm{g}(\bm{x}_k;\xi_k)-\nabla f(\bm{x}_k))\rangle]\nonumber\\
&=\mathbb{E}[\langle \nabla f(\bm{x}_k), \alpha_{k-1}H_{k-1}^{-1}(\mathbb{E}_{\xi_{k}}[\bm{g}(\bm{x}_k;\xi_k)]-\nabla f(\bm{x}_k))\rangle]
=0,
\end{align}
which, together with the H\"older inequality, implies
\begin{align*}
&\mathbb{E}[\langle \nabla f(\bm{x}_k), (\alpha_{k-1}H_{k-1}^{-1} - \alpha_kH_k^{-1})(\bm{g}(\bm{x}_k;\xi_k)-\nabla f(\bm{x}_k))\rangle]
=\alpha_k\mathbb{E}[\langle \nabla f(\bm{x}_k), (H_{k-1}^{-1} - H_k^{-1})(\bm{g}(\bm{x}_k;\xi_k)-\nabla f(\bm{x}_k))\rangle]\\
&\leq G\alpha_k \mathbb{E}[\|H_{k-1}^{-1} - H_k^{-1}\|_{\text{op}}\|\bm{g}(\bm{x}_k;\xi_k)-\nabla f(\bm{x}_k)\|]
\leq \frac{G(\epsilon^{-\mathfrak{p}}+\epsilon^{-1})\alpha_k}{(1-\beta_{2,1})^{(\mathfrak{p}-1)/2}}(1 - \beta_{2,k})^{\mathfrak{h}/2}\mathbb{E}[\|\bm{g}(\bm{x}_k;\xi_k)\|^{\mathfrak{p}-1}\|\bm{g}(\bm{x}_k;\xi_k)-\nabla f(\bm{x}_k)\|]\\
&\leq \frac{G(\epsilon^{-\mathfrak{p}}+\epsilon^{-1})\alpha_k}{(1-\beta_{2,1})^{(\mathfrak{p}-1)/2}}(1 - \beta_{2,k})^{\mathfrak{h}/2}\mathbb{E}[\|\bm{g}(\bm{x}_k;\xi_k)\|^{(\mathfrak{p}-1)\frac{\mathfrak{p}}{\mathfrak{p}-1}}]^{\frac{\mathfrak{p}-1}{\mathfrak{p}}}\mathbb{E}[\|\bm{g}(\bm{x}_k;\xi_k)-\nabla f(\bm{x}_k)\|^{\mathfrak{p}}]^{1/\mathfrak{p}}\\
&\leq \frac{G(\epsilon^{-\mathfrak{p}}+\epsilon^{-1})\sigma(\sigma+G)^{\mathfrak{p}-1}\alpha_k}{(1-\beta_{2,1})^{(\mathfrak{p}-1)/2}}(1 - \beta_{2,k})^{\mathfrak{h}/2}.
\end{align*}
Hence, by combining the above estimations and considering
\begin{align*}
\hat{\bm{m}}_k-\nabla f(\bm{x}_k)
&=\frac{1}{1-\prod_{j=1}^k\beta_{1,j}}\bm{m}_k-\nabla f(\bm{x}_k)
=\frac{\beta_{1,k}}{1-\prod_{j=1}^k\beta_{1,j}}\bm{m}_{k-1}+\frac{1-\beta_{1,k}}{1-\prod_{j=1}^k\beta_{1,j}}\bm{g}(\bm{x}_k;\xi_k)-\nabla f(\bm{x}_k)\\
&=\frac{\beta_{1,k}(1-\prod_{j=1}^{k-1}\beta_{1,j})}{1-\prod_{j=1}^k\beta_{1,j}}\hat{\bm{m}}_{k-1}+\frac{1-\beta_{1,k}}{1-\prod_{j=1}^k\beta_{1,j}}\bm{g}(\bm{x}_k;\xi_k)-\nabla f(\bm{x}_k)\\
&=\frac{\beta_{1,k}(1-\prod_{j=1}^{k-1}\beta_{1,j})}{1-\prod_{j=1}^k\beta_{1,j}}(\hat{\bm{m}}_{k-1}-\nabla f(\bm{x}_{k})) +\left(1-\frac{\beta_{1,k}(1-\prod_{j=1}^{k-1}\beta_{1,j})}{1-\prod_{j=1}^k\beta_{1,j}}\right)(\bm{g}(\bm{x}_k;\xi_k)-\nabla f(\bm{x}_k)),
\end{align*}
$\sum_{k=1}^K\mathbb{E}[X_2]$ can be estimated as
\begin{align}\label{eq:thm1_addi_1}
\sum_{k=1}^K\mathbb{E}[X_2]
&=\sum_{k=1}^K\mathbb{E}[\left\langle \nabla f(\bm{x}_k), (\alpha_{k-1}H_{k-1}^{-1} - \alpha_kH_k^{-1})\nabla f(\bm{x}_k)\right\rangle]\nonumber\\
&\quad +\sum_{k=1}^K\frac{\beta_{1,k}(1-\prod_{j=1}^{k-1}\beta_{1,j})}{1-\prod_{j=1}^k\beta_{1,j}}\mathbb{E}[\langle \nabla f(\bm{x}_k), (\alpha_{k-1}H_{k-1}^{-1} - \alpha_kH_k^{-1})(\hat{\bm{m}}_{k-1}-\nabla f(\bm{x}_{k}))\rangle]\nonumber\\
&\quad +\sum_{k=1}^K\left(1-\frac{\beta_{1,k}(1-\prod_{j=1}^{k-1}\beta_{1,j})}{1-\prod_{j=1}^k\beta_{1,j}}\right)\mathbb{E}[\langle \nabla f(\bm{x}_k), (\alpha_{k-1}H_{k-1}^{-1} - \alpha_kH_k^{-1})(\bm{g}(\bm{x}_k;\xi_k)-\nabla f(\bm{x}_k))\rangle]\\
&\leq \frac{GL_{\mathfrak{h}}\hat{Q}_2^{\mathfrak{h}}}{\epsilon}\sum_{k=1}^{K}\alpha_{k}^{1+\mathfrak{h}}+\frac{2G\alpha_{\max}(\sigma+2G+L_{\mathfrak{h}}\hat{Q}_2^{\mathfrak{h}}\alpha_{\max}^{\mathfrak{h}})}{\epsilon(1-\bar{\beta}_1)}\sum_{k=1}^{K}\beta_{1,k}+\frac{G(\epsilon^{-\mathfrak{p}}+\epsilon^{-1})\sigma(\sigma+G)^{\mathfrak{p}-1}}{(1-\beta_{2,1})^{(\mathfrak{p}-1)/2}}\sum_{k=1}^{K}\alpha_k(1 - \beta_{2,k})^{\mathfrak{h}/2}.\nonumber
\end{align}
From the evaluations of $X_1,X_2$ and $X_3$ and taking the expectation of the both sides of the descent lemma, we have
\begin{align*}
&\mathbb{E}[f(\bm{x}_{k+1})]
\leq \mathbb{E}[f(\bm{x}_k)] + \mathbb{E}[X_1]+\mathbb{E}[X_2]+\frac{\alpha_k^{1+\mathfrak{h}}L_{\mathfrak{h}}}{1+\mathfrak{h}}\mathbb{E}[X_3]\\
&\leq\mathbb{E}[f(\bm{x}_k)] +\left(-\frac{(1-\bar{\beta_{1}})\alpha_{k}}{\hat{Q}_1G^{\frac{2}{1+\mathfrak{p}}}}\mathbb{E}[\|\nabla f(\bm{x}_k)\|^2]^{\frac{1+\mathfrak{p}}{\mathfrak{p}}} + \alpha_{\max}G\hat{Q}_2\beta_{1,k}\right)+\mathbb{E}[X_2] +\frac{\alpha_k^{1+\mathfrak{h}} L_{\mathfrak{h}}}{1+\mathfrak{h}}\left(2^{1+\mathfrak{h}}\epsilon^{-(1+\mathfrak{h})}\sigma^{1+\mathfrak{h}}+2^{1+\mathfrak{h}}\epsilon^{-(1+\mathfrak{h})}G^{1+\mathfrak{h}}\right)\\
&\leq \mathbb{E}[f(\bm{x}_k)] -\frac{(1-\bar{\beta_{1}})\alpha_{k}}{\hat{Q}_1G^{\frac{2}{1+\mathfrak{p}}}}\mathbb{E}[\|\nabla f(\bm{x}_k)\|^2]^{\frac{1+\mathfrak{p}}{\mathfrak{p}}} + \alpha_{\max}G\hat{Q}_2\beta_{1,k}+\mathbb{E}[X_2]+\frac{2^{1+\mathfrak{h}}\epsilon^{-(1+\mathfrak{h})}L_{\mathfrak{h}}(\sigma^{1+\mathfrak{h}}+G^{1+\mathfrak{h}})}{1+\mathfrak{h}}\alpha_k^{1+\mathfrak{h}},
\end{align*}
which, together with \ref{asm:lowerbounded_objective}, leads to
\begin{align*}
&\sum_{k=1}^K\alpha_{k}\mathbb{E}[\|\nabla f(\bm{x}_k)\|^2]^{\frac{1+\mathfrak{p}}{\mathfrak{p}}}
\leq\frac{\hat{Q}_1G^{\frac{2}{1+\mathfrak{p}}}}{1-\bar{\beta}_1}\sum_{k=1}^K(\mathbb{E}[f(\bm{x}_k)] -\mathbb{E}[f(\bm{x}_{k+1})])+ \frac{\alpha_{\max}G^{\frac{3+\mathfrak{p}}{1+\mathfrak{p}}}\hat{Q}_1\hat{Q}_2}{1-\bar{\beta}_1}\sum_{k=1}^K\beta_{1,k}+\frac{\hat{Q}_1G^{\frac{2}{1+\mathfrak{p}}}}{1-\bar{\beta}_1}\sum_{k=1}^K\mathbb{E}[X_2]\\
&\quad +\frac{2^{1+\mathfrak{h}}\epsilon^{-(1+\mathfrak{h})}L_{\mathfrak{h}}\hat{Q}_1G^{\frac{2}{1+\mathfrak{p}}}(\sigma^{1+\mathfrak{h}}+G^{1+\mathfrak{h}})}{(1+\mathfrak{h})(1-\bar{\beta}_1)}\sum_{k=1}^K\alpha_k^{1+\mathfrak{h}}\\
&\leq \frac{\hat{Q}_1G^{\frac{2}{1+\mathfrak{p}}}(f(\bm{x}_1)-f_{\star})}{1-\bar{\beta}_1}+ \frac{2\alpha_{\max}\epsilon^{-1}G^{\frac{3+\mathfrak{p}}{1+\mathfrak{p}}}\hat{Q}_1(\hat{Q}_2\epsilon+\sigma+2G+L_{\mathfrak{h}}\alpha_{\max}^{\mathfrak{h}}\hat{Q}_2^{\mathfrak{h}})}{(1-\bar{\beta}_1)^2}\sum_{k=1}^K\beta_{1,k}\\
&\quad +\frac{L_{\mathfrak{h}}\epsilon^{-(1+\mathfrak{h})}G^{\frac{2}{1+\mathfrak{p}}}\hat{Q}_1\{2^{1+\mathfrak{h}}(\sigma^{1+\mathfrak{h}}+G^{1+\mathfrak{h}})+G(1+\mathfrak{h})\epsilon^{\mathfrak{h}}\hat{Q}_2^{\mathfrak{h}}\}}{(1+\mathfrak{h})(1-\bar{\beta}_1)}\sum_{k=1}^K\alpha_k^{1+\mathfrak{h}}\\
&\quad +\frac{\hat{Q}_1\sigma G^{\frac{3+\mathfrak{p}}{1+\mathfrak{p}}}(\epsilon^{-\mathfrak{p}}+\epsilon^{-1})(\sigma+G)^{\mathfrak{p}-1}}{(1-\bar{\beta}_1)(1-\beta_{2,1})^{(\mathfrak{p}-1)/2}}\sum_{k=1}^{K}\alpha_k(1 - \beta_{2,k})^{\mathfrak{h}/2}.
\end{align*}
Therefore, since $x\mapsto x^{(1+\mathfrak{p})/\mathfrak{p}}$ is convex due to $(1+\mathfrak{p})/\mathfrak{p}\geq1$, the last estimation and Jensen's inequality imply
\begin{align*}
&\left(\underset{1\leq k\leq K}{\min}\mathbb{E}[\|\nabla f(\bm{x}_k)\|^2]\right)^{\frac{1+\mathfrak{p}}{\mathfrak{p}}}
\leq \left(\frac{1}{\sum_{k=1}^{K}\alpha_k}\sum_{k=1}^K\alpha_{k}\mathbb{E}[\|\nabla f(\bm{x}_k)\|^2]\right)^{\frac{1+\mathfrak{p}}{\mathfrak{p}}}
\leq \frac{1}{\sum_{k=1}^{K}\alpha_k}\sum_{k=1}^K\alpha_{k}\mathbb{E}[\|\nabla f(\bm{x}_k)\|^2]^{\frac{1+\mathfrak{p}}{\mathfrak{p}}}
\end{align*}
which, together with $(u+v)^r\leq u^r+v^r (r\in(0,1])$, yields the upper bound. Finally, using the conditions $\alpha_{k}\leq \alpha_{k-1}$ and $\beta_{k}\leq \beta_{k-1}$, we modify the upper bound. From the similar calculation, we have
\begin{align*}
\mathbb{E}[X_1]
\leq -\frac{(1-\bar{\beta_{1}})\alpha_{k}}{\hat{Q}_1G^{\frac{2}{1+\mathfrak{p}}}}\mathbb{E}[\|\nabla f(\bm{x}_k)\|^2]^{\frac{1+\mathfrak{p}}{\mathfrak{p}}} + G\hat{Q}_2\alpha_{k-1}\beta_{1,k}
\leq -\frac{(1-\bar{\beta_{1}})\alpha_{k}}{\hat{Q}_1G^{\frac{2}{1+\mathfrak{p}}}}\mathbb{E}[\|\nabla f(\bm{x}_k)\|^2]^{\frac{1+\mathfrak{p}}{\mathfrak{p}}} + G\hat{Q}_2\alpha_{k-1}\beta_{1,k-1}
\end{align*}
and
\begin{align*}
&\sum_{k=1}^K\frac{\beta_{1,k}(1-\prod_{j=1}^{k-1}\beta_{1,j})}{1-\prod_{j=1}^k\beta_{1,j}}\mathbb{E}[\left\langle \nabla f(\bm{x}_k), (\alpha_{k-1}H_{k-1}^{-1} - \alpha_kH_k^{-1})(\hat{\bm{m}}_{k-1}-\nabla f(\bm{x}_{k}))\right\rangle]\\
&\leq \sum_{k=1}^K\frac{\beta_{1,k}(1-\prod_{j=1}^{k-1}\beta_{1,j})}{1-\prod_{j=1}^k\beta_{1,j}}\frac{G(\alpha_{k-1}+\alpha_k)}{\epsilon}(\sigma+2G +L_{\mathfrak{h}}\hat{Q}_2^{\mathfrak{h}}\alpha_{k-1}^{\mathfrak{h}})\\
&\leq \frac{2G}{\epsilon(1-\bar{\beta}_{1})}\sum_{k=1}^K\beta_{1,k}\alpha_{k-1}(\sigma+2G +L_{\mathfrak{h}}\hat{Q}_2^{\mathfrak{h}}\alpha_{k-1}^{\mathfrak{h}})
\leq \frac{2G(\sigma+2G)}{\epsilon(1-\bar{\beta}_{1})}\sum_{k=1}^K\beta_{1,k-1}\alpha_{k-1}+\frac{2GL_{\mathfrak{h}}\hat{Q}_2^{\mathfrak{h}}\bar{\beta}_1}{\epsilon(1-\bar{\beta}_{1})}\sum_{k=1}^K\alpha_{k-1}^{1+\mathfrak{h}}\\
&\leq \frac{2G(\sigma+2G)}{\epsilon(1-\bar{\beta}_{1})}\sum_{k=1}^K\beta_{1,k}\alpha_{k}+\frac{2GL_{\mathfrak{h}}\hat{Q}_2^{\mathfrak{h}}\bar{\beta}_1}{\epsilon(1-\bar{\beta}_{1})}\sum_{k=1}^K\alpha_{k}^{1+\mathfrak{h}}.
\end{align*}
By re-estimating \eqref{eq:thm1_addi_1} with the latter one, and using the former one and the estimation of $X_3$, we obtain another result as
\begin{align}\label{eq:alter_thm1}
&\left(\underset{1\leq k\leq K}{\min}\mathbb{E}[\|\nabla f(\bm{x}_k)\|^2]\right)^{\frac{1+\mathfrak{p}}{\mathfrak{p}}}
\leq \left(\frac{1}{\sum_{k=1}^{K}\alpha_k}\sum_{k=1}^K\alpha_{k}\mathbb{E}[\|\nabla f(\bm{x}_k)\|^2]\right)^{\frac{1+\mathfrak{p}}{\mathfrak{p}}}
\leq \frac{1}{\sum_{k=1}^{K}\alpha_k}\sum_{k=1}^K\alpha_{k}\mathbb{E}[\|\nabla f(\bm{x}_k)\|^2]^{\frac{1+\mathfrak{p}}{\mathfrak{p}}}\nonumber\\
&\leq \frac{\hat{Q}_1G^{\frac{2}{1+\mathfrak{p}}}(f(\bm{x}_1)-f_{\star})}{1-\bar{\beta}_1}\frac{1}{\sum_{k=1}^{K}\alpha_k}+ \underbrace{\frac{2\epsilon^{-1}G^{\frac{3+\mathfrak{p}}{1+\mathfrak{p}}}\hat{Q}_1(\hat{Q}_2\epsilon+\sigma+2G)}{(1-\bar{\beta}_1)^2}}_{\eqcolon {Q'}_2^{\frac{1+\mathfrak{p}}{\mathfrak{p}}}}\frac{\sum_{k=1}^K\alpha_k\beta_{1,k}}{\sum_{k=1}^{K}\alpha_k}\nonumber\\
&\quad +\frac{4L_{\mathfrak{h}}\epsilon^{-(1+\mathfrak{h})}G^{\frac{2}{1+\mathfrak{p}}}\hat{Q}_1\{\sigma^{1+\mathfrak{h}}+G^{1+\mathfrak{h}}+G\epsilon^{\mathfrak{h}}\hat{Q}_2^{\mathfrak{h}}\}}{(1+\mathfrak{h})(1-\bar{\beta}_1)^2}\frac{\sum_{k=1}^K\alpha_k^{1+\mathfrak{h}}}{\sum_{k=1}^{K}\alpha_k}+\frac{\hat{Q}_1\sigma G^{\frac{3+\mathfrak{p}}{1+\mathfrak{p}}}(\epsilon^{-\mathfrak{p}}+\epsilon^{-1})(\sigma+G)^{\mathfrak{p}-1}}{(1-\bar{\beta}_1)(1-\beta_{2,1})^{(\mathfrak{p}-1)/2}}\sum_{k=1}^{K}\alpha_k(1 - \beta_{2,k})^{\mathfrak{h}/2}.
\end{align}
\end{proof}

\begin{proof}[Proof of Corollary \ref{cor:heavy_convrate}]
For any $r\in(0,1)$, $\sum_{k=1}^{K}\alpha_k=\alpha_{\max}\sum_{k=1}^{K}k^{-r}\geq\alpha_{\max}KK^{-r}=\alpha_{\max}K^{1-r}$ holds. From $s>2(1-r)/\mathfrak{h}$, i.e., $r+s\mathfrak{h}/2>1$, we note that $\sum_{k=1}^K\alpha_k(1-\beta_{2,k})^{\mathfrak{h}/2}\leq \tilde{\beta}_2\sum_{k=1}^{\infty} k^{-(r+s\mathfrak{h}/2)}<\infty$. 

When $r\in(0,1/(1+\mathfrak{h}))$ and $s>2r$, $\sum_{k=1}^{K}\alpha_k^{1+\mathfrak{h}}\leq \alpha_{\max}^{1+\mathfrak{h}}\int_0^{K}t^{-r(1+\mathfrak{h})}dt=\alpha_{\max}^{1+\mathfrak{h}}(1-r(1+\mathfrak{h}))^{-1}(K^{1-r(1+\mathfrak{h})}-\lim_{\omega\downarrow0}\omega)=\alpha_{\max}^{1+\mathfrak{h}}K^{1-r(1+\mathfrak{h})}/(1-r(1+\mathfrak{h}))$ leads to $\sum_{k=1}^{K}\alpha_k^{1+\mathfrak{h}}/\sum_{k=1}^{K}\alpha_k=O(K^{1-r(1+\mathfrak{h})-(1-r)})=O(K^{-r\mathfrak{h}})$. The result is followed from this equation and $r+s\mathfrak{h}/2>r(1+\mathfrak{h})$ yielding $\alpha_k(1-\beta_{2,k})^{\mathfrak{h}/2}\leq \alpha_k^{1+\mathfrak{h}}$.

When $r\in(1/(1+\mathfrak{h}),1)$, $\int_1^{\infty}t^{-r(1+\mathfrak{h})}dt=(1-r(1+\mathfrak{h}))^{-1}(\lim_{t\to\infty}1/t^{r(1+\mathfrak{h})-1}-1)=(r(1+\mathfrak{h})-1)^{-1}$ leads to $\sum_{k=1}^{K}\alpha_k^{1+\mathfrak{h}}\leq \alpha_{\max}^{1+\mathfrak{h}}(1+\int_1^{\infty}t^{-r(1+\mathfrak{h})}dt)=\alpha_{\max}^{1+\mathfrak{h}}(1+\int_1^{\infty}t^{-r(1+\mathfrak{h})}dt)=\alpha_{\max}^{1+\mathfrak{h}}(1+(r(1+\mathfrak{h})-1)^{-1})<\infty$ and $\sum_{k=1}^{K}\alpha_k^{1+\mathfrak{h}}/\sum_{k=1}^{K}\alpha_k=O(1//\sum_{k=1}^{K}\alpha_k)=O(K^{-(1-r)})$. 

When $r=1$, $\int_1^{\infty}t^{-(1+\mathfrak{h})}dt=(-\mathfrak{h})^{-1}(\lim_{t\to\infty}1/t^{\mathfrak{h}}-1)=\mathfrak{h}^{-1}$ and $\sum_{k=1}^{K-1}\alpha_k\geq\alpha_{\max}\int_1^{K}dt/t=\alpha_{\max}\log K$ lead to $\sum_{k=1}^{K}\alpha_k^{1+\mathfrak{h}}/\sum_{k=1}^{K}\alpha_k=O(1/\sum_{k=1}^{K}\alpha_k)=O(1/\log K)$. 

When $r=1/(1+\mathfrak{h})$, $\sum_{k=1}^{K}\alpha_k^{1+\mathfrak{h}}=\alpha_{\max}^{1+\mathfrak{h}}\sum_{k=1}^{K}1/k\leq \alpha_{\max}^{1+\mathfrak{h}}(1+\int_{1}^{K}dt/t)=\alpha_{\max}^{1+\mathfrak{h}}(1+\log K)$ leads to $\sum_{k=1}^{K}\alpha_k^{1+\mathfrak{h}}/\sum_{k=1}^{K}\alpha_k=O(\log K/K^{1-1/(1+\mathfrak{h})})=O(\log K/K^{\mathfrak{h}/(1+\mathfrak{h})})$.
\qed
\end{proof}

\section{Proof of Theorem \ref{thm:holder_bcg_conv}}\label{appdix:prf_holder_bcg_conv}
\begin{lemma} \label{lem:appdix_boundgradnorm}
Under \ref{asm:asboundedsg}, for all $k \in \mathbb{N}$ and $q\in(1,2)$,
\begin{align*}
\|\bm{g}_{B_k}(\bm{x}_k)\|\leq G_N,\ \|\nabla f_N(\bm{x}_k) \|\leq G_N\text{ a.s.},\ \mathbb{E}_{\bm{\iota}_k}[\|\bm{g}_{B_k}(\bm{x}_k)-\nabla f(\bm{x}_k)\|^q]\leq \frac{2^{q}C_qG_N^q}{b_k^{q-1}}.
\end{align*}
Additionally, we have $\mathbb{E}_{\bm{\iota}_k}[\|\bm{g}_{B_k}(\bm{x}_k)-\nabla f(\bm{x}_k)\|^2]\leq G_N^2/b_k$ and $\|\hat{\bm{m}}_{k}\|\leq G_N$ a.s.
\end{lemma}

\begin{proof}
From \ref{asm:asboundedsg}, the most left one holds due to $\| \bm{g}_{B_k}(\bm{x}_k)\|
\leq b_k^{-1}\sum_{i=1}^{b_k}\| \bm{g}_{\iota_{k, i}}(\bm{x}_k)\|
\leq b_k^{-1}\sum_{i=1}^{b_k}G_N
=G_N$ a.s. From \ref{asm:asboundedsg} and Proposition \ref{prop:unbiasedestimator_sg_empi}, the middle one is followed as $\|\nabla f_N(\bm{x}_k) \|
=\|\mathbb{E}_{\iota_{k,i}}[\bm{g}_{\iota_{k, i}}(\bm{x}_k)]\|
\leq \mathbb{E}_{\iota_{k,i}}[\|\bm{g}_{\iota_{k, i}}(\bm{x}_k)\|]
\leq G_N$. Combining these results and Theorem 3.1 in \citet{pinelis2022multidim_vonbar_ineq} yield the most right one:
\begin{align*}
&\mathbb{E}_{\bm{\iota}_k}[\|\bm{g}_{B_k}(\bm{x}_k)-\nabla f(\bm{x}_k)\|^q]
=\mathbb{E}_{\bm{\iota}_k}\left[\left\|\frac{1}{b_k}\sum_{j=1}^{b_k}\bm{g}_{\iota_{k,j}}(\bm{x}_k)-\nabla f(\bm{x}_k)\right\|^q\right]
=\mathbb{E}_{\bm{\iota}_k}\left[\left\|\frac{1}{b_k}\sum_{j=1}^{b_k}\{\bm{g}_{\iota_{k,j}}(\bm{x}_k)-\nabla f(\bm{x}_k)\}\right\|^q\right]\\
&\leq\frac{C_q}{b_k^q}\sum_{j=1}^{b_k}\mathbb{E}_{\iota_{k,j}}[\left\|\bm{g}_{\iota_{k,j}}(\bm{x}_k)-\nabla f(\bm{x}_k)\right\|^q]
\leq\frac{2^{q-1}C_q}{b_k^q}\sum_{j=1}^{b_k}(\mathbb{E}_{\iota_{k,j}}[\|\bm{g}_{\iota_{k,j}}(\bm{x}_k)\|^q]+\mathbb{E}_{\iota_{k,j}}[\|\nabla f(\bm{x}_k)\|^q])
\leq \frac{2^{q}C_qG_N^q}{b_k^{q-1}},
\end{align*}
where $C_{\mathfrak{p}}\in[1,2]$ appears in \citet{pinelis2022multidim_vonbar_ineq}. In particular, when $q=2$, we have
\begin{align*}
&\mathbb{E}_{\bm{\iota}_k}[\|\bm{g}_{B_k}(\bm{x}_k)-\nabla f(\bm{x}_k)\|^2]
\leq\frac{1}{b_k^2}\sum_{j=1}^{b_k}\mathbb{E}_{\iota_{k,j}}[\left\|\bm{g}_{\iota_{k,j}}(\bm{x}_k)-\nabla f(\bm{x}_k)\right\|^2]\\
&=\frac{1}{b_k^2}\sum_{j=1}^{b_k}\{\mathbb{E}_{\iota_{k,j}}[\|\bm{g}_{\iota_{k,j}}(\bm{x}_k)\|^2]+\mathbb{E}_{\iota_{k,j}}[\|\nabla f(\bm{x}_k)\|^2]-2\mathbb{E}_{\iota_{k,j}}[\langle \bm{g}_{\iota_{k,j}}(\bm{x}_k),\nabla f(\bm{x}_k)\rangle]\}\\
&=\frac{1}{b_k^2}\sum_{j=1}^{b_k}\{\mathbb{E}_{\iota_{k,j}}[\|\bm{g}_{\iota_{k,j}}(\bm{x}_k)\|^2]-\mathbb{E}_{\iota_{k,j}}[\|\nabla f(\bm{x}_k)\|^2]\}
\leq \frac{1}{b_k^2}\sum_{j=1}^{b_k}\mathbb{E}_{\iota_{k,j}}[\|\bm{g}_{\iota_{k,j}}(\bm{x}_k)\|^2]
\leq \frac{G_N^2}{b_k}.
\end{align*}
By applying the similar estimation technique as in Lemma \ref{lem:heavy_conv}, we obtain $\|\hat{\bm{m}}_{k}\|\leq G_N$ a.s.
\end{proof}

\begin{proof}[Proof of \ref{thm:holder_bcg_upperbound_beta2}]
\ref{asm:holdersmooth_gradient} and the updating rule of Algorithm \ref{algo:adam} lead to
\begin{align*}
f_N(\bm{x}_{k+1}) 
&\leq f_N(\bm{x}_k) - \alpha_k\left\langle \nabla f_N(\bm{x}_k), H_k^{-1}\hat{\bm{m}}_k\right\rangle + \frac{\alpha_k^{1+\mathfrak{h}}L}{1+\mathfrak{h}}\left\| H_k^{-1}\hat{\bm{m}}_k \right\|^{1+\mathfrak{h}} \\
&=f_N(\bm{x}_k) - \alpha_k\underbrace{\left\langle \nabla f_N(\bm{x}_k), H_k^{-1}\nabla f_N(\bm{x}_k)\right\rangle}_{X_1} - \alpha_k\underbrace{\left\langle \nabla f_N(\bm{x}_k), H_k^{-1}(\hat{\bm{m}}_k-\bm{g}_{B_k}(\bm{x}_k)\right\rangle}_{X_2}\\
&\quad - \alpha_k\underbrace{\left\langle \nabla f_N(\bm{x}_k), H_k^{-1}(\bm{g}_{B_k}(\bm{x}_k)-\nabla f_N(\bm{x}_k))\right\rangle}_{X_3} + \frac{\alpha_k^{1+\mathfrak{h}}L}{1+\mathfrak{h}}\underbrace{\left\| H_k^{-1}\hat{\bm{m}}_k \right\|^{1+\mathfrak{h}}}_{X_4}.
\end{align*}
Proposition \ref{prop:bounds_paramofadam} gives a lower bound of $H_k^{-1}$, which yields $\mathbb{E}[X_1]
\geq \mu \mathbb{E}[\|\nabla f_N(\bm{x}_k)\|^2]$. From Lemma \ref{lem:appdix_boundgradnorm} and Proposition \ref{prop:bounds_paramofadam}, we have $\mathbb{E}[X_4]
\leq \mathbb{E}[\| H_k^{-1}\|_{\text{op}}^{1+\mathfrak{h}}\|\hat{\bm{m}}_k \|^{1+\mathfrak{h}}]
\leq \nu^{1+\mathfrak{h}}\mathbb{E}[\|\hat{\bm{m}}_k \|^{1+\mathfrak{h}}]
\leq \nu^{1+\mathfrak{h}}G_N^{1+\mathfrak{h}}$., where $1/(G_N+\epsilon)$ should be used as $\mu$ in the first result; indeed, the definition of $\bm{v}_k=\beta_{2,k}\bm{v}_{k-1}+(1-\beta_{2,k})\bm{g}_{B_k}(\bm{x}_k)\odot\bm{g}_{B_k}(\bm{x}_k), \bm{v}_0=0$ and $\sum_{j=1}^k (1-\beta_{2,j})\left(\prod_{i=j+1}^k\beta_{2,i}\right)=1-\prod_{j=1}^k\beta_{2,k}$ yield
\begin{align*}
\|\hat{\bm{v}}_k\|
&=\frac{1}{1-\prod_{j=1}^k\beta_{2,k}}\left\|\left(\prod_{j=1}^k\beta_{2,j}\right)\bm{v}_0+\sum_{j=1}^k (1-\beta_{2,j})\left(\prod_{i=j+1}^k\beta_{2,i}\right)\bm{g}_{B_j}(\bm{x}_j)\odot\bm{g}_{B_j}(\bm{x}_j)\right\|\\
&=\frac{1}{1-\prod_{j=1}^k\beta_{2,k}}\left\|\sum_{j=1}^k (1-\beta_{2,j})\left(\prod_{i=j+1}^k\beta_{2,i}\right)\bm{g}_{B_j}(\bm{x}_j)\odot\bm{g}_{B_j}(\bm{x}_j)\right\|\\
&\leq \frac{G_N^2}{1-\prod_{j=1}^k\beta_{2,k}}\sum_{j=1}^k (1-\beta_{2,j})\left(\prod_{i=j+1}^k\beta_{2,i}\right)
=G_N^2 \text{ a.s.}
\end{align*}
Furthermore, a similar calculation as in \eqref{eq:m_minus_g_sqr} leads to $\|\hat{\bm{m}}_k-\bm{g}_{B_k}(\bm{x}_k)\|\leq 2G_N\beta_{1,k}$ a.s., which, together with Young's inequality $ab\leq \mu a^2/4+b^2/4\mu$ and Jensen's inequality, implies
\begin{align*}
\mathbb{E}[X_2]
&\geq -\mathbb{E}[|X_2|]
\geq -\mathbb{E}[\| \nabla f_N(\bm{x}_k)\|\|H_k^{-1}\|_{\text{op}}\|\hat{\bm{m}}_k-\bm{g}_{B_k}(\bm{x}_k)\|]
\geq -2\nu G_N\beta_{1,k}\mathbb{E}[\| \nabla f_N(\bm{x}_k)\|]\\
&\geq -\frac{4\nu^2 G_N^2\beta_{1,k}^2}{\mu}-\frac{\mu}{4}\mathbb{E}[\| \nabla f_N(\bm{x}_k)\|^2].
\end{align*}
The component-wise definition of $\hat{\bm{v}}_{k}$ leads to
\begin{align*}
|\hat{v}_{k,i}-\hat{v}_{k-1,i}|
=\frac{1 - \beta_{2,k}}{1-\prod_{j=1}^k{\beta}_{2, j}}|g_{k,i}^2-\hat{v}_{k-1,i}|
\leq \frac{1 - \beta_{2,k}}{1-\prod_{j=1}^k{\beta}_{2, j}}\|\bm{g}_{\iota_{k,i}}(\bm{x}_k)\odot\bm{g}_{\iota_{k,i}}(\bm{x}_k)-\hat{\bm{v}}_{k-1}\|
\leq 2G_N^2\frac{1 - \beta_{2,k}}{1-\prod_{j=1}^k{\beta}_{2, j}}\text{ a.s.}
\end{align*}
Noting that $|\sqrt{a}-\sqrt{b}|\leq \sqrt{|a-b|}$ and $|(\sqrt{a}+\epsilon)^{-1}-(\sqrt{b}+\epsilon)^{-1}|=|(\sqrt{b}-\sqrt{a})/\{(\sqrt{a}+\epsilon)(\sqrt{b}+\epsilon)\}|\leq |\sqrt{a}-\sqrt{b}|/\epsilon^2$ for $a,b\geq 0$, we have
\begin{align*}
\|H_k^{-1}-H_{k-1}^{-1}\|_{\text{op}}
&=\max_{i\in\mathbb{N}_d}\left|(\sqrt{\hat{v}_{k,i}}+\epsilon)^{-1}-(\sqrt{\hat{v}_{k-1,i}}+\epsilon)^{-1}\right|
\leq \epsilon^{-2}\max_i\left|\sqrt{\hat{v}_{k,i}}-\sqrt{\hat{v}_{k-1,i}}\right|\\
&\leq \epsilon^{-2}\max_i\sqrt{\left|\hat{v}_{k,i}-\hat{v}_{k-1,i}\right|}
\leq \frac{\sqrt{2}G_N}{\epsilon^2}\sqrt{\frac{1 - \beta_{2,k}}{1-\prod_{j=1}^k{\beta}_{2, j}}} \text{ a.s.}
\end{align*}
Since \eqref{eq:thm1_delete_k-1} yields $\mathbb{E}_{\bm{\iota}_k}[X_3]=\mathbb{E}_{\bm{\iota}_k}[\langle \nabla f_N(\bm{x}_k), (H_{k}^{-1}-H_{k-1}^{-1})(\bm{g}_{B_k}(\bm{x}_k)-\nabla f_N(\bm{x}_k))\rangle]$, the Cauchy-Schwartz inequality, Young's inequality, and $1-{\beta}_{2, 1}\leq1-\prod_{j=1}^k{\beta}_{2, j}$ imply
\begin{align*}
&\mathbb{E}_{\bm{\iota}_k}[X_3]
\geq -\mathbb{E}_{\bm{\iota}_k}[|X_3|]
= -\mathbb{E}_{\bm{\iota}_k}[|\langle \nabla f_N(\bm{x}_k), (H_{k}^{-1}-H_{k-1}^{-1})(\bm{g}_{B_k}(\bm{x}_k)-\nabla f_N(\bm{x}_k))\rangle|]\\
&\geq -\mathbb{E}_{\bm{\iota}_k}[\|\nabla f_N(\bm{x}_k)\|\|H_{k}^{-1}-H_{k-1}^{-1}\|_{\text{op}}\|\|\bm{g}_{B_k}(\bm{x}_k)-\nabla f_N(\bm{x}_k)\|]\\
&\geq -\mathbb{E}_{\bm{\iota}_k}\left[\|\nabla f_N(\bm{x}_k)\|\left(\frac{\sqrt{2}G_N}{\epsilon^2}\sqrt{\frac{1 - \beta_{2,k}}{1-\prod_{j=1}^k{\beta}_{2, j}}}\|\bm{g}_{B_k}(\bm{x}_k)-\nabla f_N(\bm{x}_k)\|\right)\right]\\
&\geq -\frac{\mu}{4}\mathbb{E}_{\bm{\iota}_k}[\|\nabla f_N(\bm{x}_k)\|^2]-\frac{G_N^2}{2\epsilon^4\mu}\frac{1 - \beta_{2,k}}{1-\prod_{j=1}^k{\beta}_{2, j}}\mathbb{E}_{\bm{\iota}_k}[\|\bm{g}_{B_k}(\bm{x}_k)-\nabla f_N(\bm{x}_k)\|]\\
&\geq -\frac{\mu}{4}\mathbb{E}_{\bm{\iota}_k}[\|\nabla f_N(\bm{x}_k)\|^2]-\frac{G_N^2}{2\epsilon^4\mu}\frac{1 - \beta_{2,k}}{1-{\beta}_{2, 1}}\frac{G_N^2}{b_k}
\geq -\frac{\mu}{4}\mathbb{E}_{\bm{\iota}_k}[\|\nabla f_N(\bm{x}_k)\|^2]-\frac{G_N^4}{2\epsilon^4\mu}\frac{1 - \beta_{2,k}}{1-{\beta}_{2, 1}}.
\end{align*}
Combining the estimations of $X_1,X_2,X_3$ and $X_4$ and taking the total expectation, we have
\begin{align*}
0
&\leq \sum_{k=1}^{K}(\mathbb{E}[f_N(\bm{x}_k)]-\mathbb{E}[f_N(\bm{x}_{k+1})]) - \sum_{k=1}^{K}\alpha_k\mathbb{E}[X_1]-\sum_{k=1}^{K}\alpha_k\mathbb{E}[X_2]-\sum_{k=1}^{K}\alpha_k\mathbb{E}[X_3]+\sum_{k=1}^{K}\frac{\alpha_k^{1+\mathfrak{h}}L_{\mathfrak{h},N}}{1+\mathfrak{h}}\mathbb{E}[X_4]\\
&\leq f_N(\bm{x}_1)-f_{N,\star} -\mu\sum_{k=1}^{K}\alpha_k \mathbb{E}[\|\nabla f_N(\bm{x}_k)\|^2]+\sum_{k=1}^{K}\alpha_k\left(\frac{4\nu^2 G_N^2\beta_{1,k}^2}{\mu}+\frac{\mu}{4}\mathbb{E}[\| \nabla f_N(\bm{x}_k)\|^2]\right)\\
&\quad+\sum_{k=1}^{K}\alpha_k\left(\frac{\mu}{4}\mathbb{E}_{\bm{\iota}_k}[\|\nabla f_N(\bm{x}_k)\|^2]+\frac{G_N^4}{2\epsilon^4\mu}\frac{1 - \beta_{2,k}}{1-{\beta}_{2, 1}}\right)+\sum_{k=1}^{K}\frac{\alpha_k^{1+\mathfrak{h}}L_{\mathfrak{h},N}}{1+\mathfrak{h}}\nu^{1+\mathfrak{h}}G_N^{1+\mathfrak{h}}\\
&= -\frac{\mu}{2}\sum_{k=1}^{K}\alpha_k \mathbb{E}[\|\nabla f_N(\bm{x}_k)\|^2]+f_N(\bm{x}_1)-f_{N,\star}+\frac{4\nu^2 G_N^2}{\mu}\sum_{k=1}^{K}\alpha_k\beta_{1,k}^2+\frac{G_N^4}{2\epsilon^4\mu(1-{\beta}_{2, 1})}\sum_{k=1}^{K}\alpha_k(1 - \beta_{2,k})\\
&\quad +\frac{L_{\mathfrak{h},N}\nu^{1+\mathfrak{h}}G_N^{1+\mathfrak{h}}}{1+\mathfrak{h}}\sum_{k=1}^{K}\alpha_k^{1+\mathfrak{h}},
\end{align*}
which proves the first result. Next, we proceed the latter result and  start with 
\begin{align*}
f_N(\bm{x}_{k+1}) 
&\leq f_N(\bm{x}_k) - \alpha_k\underbrace{\left\langle \nabla f_N(\bm{x}_k), H_k^{-1}\nabla f_N(\bm{x}_k)\right\rangle}_{X_1} - \alpha_k\underbrace{\left\langle \nabla f_N(\bm{x}_k), H_k^{-1}(\hat{\bm{m}}_k-\nabla f_N(\bm{x}_k))\right\rangle}_{X_2} + \frac{\alpha_k^{1+\mathfrak{h}}L}{1+\mathfrak{h}}\underbrace{\left\| H_k^{-1}\hat{\bm{m}}_k \right\|^{1+\mathfrak{h}}}_{X_3}.
\end{align*}
For $\mathbb{E}[X_1]$ and $\mathbb{E}[X_3]$, we adopt the same estimation with the first result. Proposition \ref{prop:bounds_paramofadam} gives an upper bound of $H_k^{-1}$, which, together with Young's inequality and Lemma \ref{lem:appdix_boundgradnorm}, implies
\begin{align*}
\mathbb{E}[X_2]
&\geq -\mathbb{E}[|X_2|]
\geq -\mathbb{E}[\|\nabla f_N(\bm{x}_k)\|\|H_k^{-1}\|_{\text{op}}\|\hat{\bm{m}}_k-\nabla f_N(\bm{x}_k)\|]
\geq -\mathbb{E}[\|\nabla f_N(\bm{x}_k)\|(\nu\|\hat{\bm{m}}_k-\nabla f_N(\bm{x}_k)\|)]\\
&\geq -\frac{\mu}{2}\mathbb{E}[\|\nabla f_N(\bm{x}_k)\|^2]-\frac{\nu}{2\mu}\mathbb{E}[\|\hat{\bm{m}}_k-\nabla f_N(\bm{x}_k)\|^2]\\
&\geq -\frac{\mu}{2}\mathbb{E}[\|\nabla f_N(\bm{x}_k)\|^2]-\frac{\nu}{\mu}\mathbb{E}[\|\hat{\bm{m}}_k-\bm{g}_{B_k}(\bm{x}_k)\|^2+\|\bm{g}_{B_k}(\bm{x}_k)-\nabla f_N(\bm{x}_k)\|^2]\\
&\geq -\frac{\mu}{2}\mathbb{E}[\|\nabla f_N(\bm{x}_k)\|^2]-\frac{\nu}{\mu}\mathbb{E}[\|\hat{\bm{m}}_k-\bm{g}_{B_k}(\bm{x}_k)\|^2]-\frac{\nu G_N^2}{\mu}\frac{1}{b_k}.
\end{align*}
Considering the updating rule of $\hat{\bm{m}}_k$ and $1-\prod_{j=1}^{k-1}{\beta}_{1, j}\leq 1-\prod_{j=1}^{k}{\beta}_{1, j}$, we have
\begin{align}\label{eq:m_minus_g_sqr}
&\mathbb{E}[\|\hat{\bm{m}}_k-\bm{g}_{B_k}(\bm{x}_k)\|^2]
=\mathbb{E}\left[\left\|\frac{1 - \beta_{1,k}}{1-\prod_{j=1}^k{\beta}_{1, j}}\bm{g}_{B_k}(\bm{x}_k) + \frac{\beta_{1,k}}{1-\prod_{j=1}^k{\beta}_{1, j}}\bm{m}_{k-1}-\bm{g}_{B_k}(\bm{x}_k)\right\|^2\right]\nonumber\\
&=\mathbb{E}\left[\left\|\frac{-\beta_{1,k}(1-\prod_{j=1}^{k-1}{\beta}_{1, j})}{1-\prod_{j=1}^k{\beta}_{1, j}}\bm{g}_{B_k}(\bm{x}_k) + \frac{\beta_{1,k}}{1-\prod_{j=1}^k{\beta}_{1, j}}\bm{m}_{k-1}\right\|^2\right]\nonumber\\
&=\mathbb{E}\left[\left\|\frac{-\beta_{1,k}(1-\prod_{j=1}^{k-1}{\beta}_{1, j})}{1-\prod_{j=1}^k{\beta}_{1, j}}\bm{g}_{B_k}(\bm{x}_k) + \frac{\beta_{1,k}(1-\prod_{j=1}^{k-1}{\beta}_{1, j})}{1-\prod_{j=1}^k{\beta}_{1, j}}\hat{\bm{m}}_{k-1}\right\|^2\right]
\leq \frac{\beta_{1,k}^2(1-\prod_{j=1}^{k-1}{\beta}_{1, j})^2}{(1-\prod_{j=1}^k{\beta}_{1, j})^2}\mathbb{E}[\|\hat{\bm{m}}_{k-1}-\bm{g}_{B_k}(\bm{x}_k)\|^2]\nonumber\\
&\leq \beta_{1,k}^2\mathbb{E}[\|\hat{\bm{m}}_{k-1}-\bm{g}_{B_k}(\bm{x}_k)\|^2]
\leq 2\beta_{1,k}^2\mathbb{E}[\|\hat{\bm{m}}_{k-1}\|^2+\|\bm{g}_{B_k}(\bm{x}_k)\|^2]
\leq 4G_N^2\beta_{1,k}^2,
\end{align}
where the last inequality is obtained from Lemma \ref{lem:appdix_boundgradnorm}. Hence, $X_2$ is evaluated as $\mathbb{E}[X_2]
\geq -2^{-1}\mu\mathbb{E}[\|\nabla f_N(\bm{x}_k)\|^2]-4\nu G_N^2\mu^{-1}\beta_{1,k}^2-\nu G_N^2\mu^{-1}/b_k$. Combining the estimations of $X_1,X_2$ and $X_3$ and taking the total expectation, the latter result holds from
\begin{align*}
0
&\leq \sum_{k=1}^{K}(\mathbb{E}[f_N(\bm{x}_k)]-\mathbb{E}[f_N(\bm{x}_{k+1})]) + \sum_{k=1}^{K}\mathbb{E}[X_1]+\sum_{k=1}^{K}\mathbb{E}[X_2]+\sum_{k=1}^{K}\frac{\alpha_k^{1+\mathfrak{h}}L_{\mathfrak{h},N}}{1+\mathfrak{h}}\mathbb{E}[X_3]\\
&= -\frac{\mu}{2}\sum_{k=1}^{K}\alpha_k \mathbb{E}[\|\nabla f_N(\bm{x}_k)\|^2]+f_N(\bm{x}_1)-f_{N,\star}+\frac{4\nu G_N^2}{\mu}\sum_{k=1}^{K}\alpha_k\beta_{1,k}^2+\frac{\nu G_N^2}{\mu}\sum_{k=1}^{K}\frac{\alpha_k}{b_k}+\frac{L_{\mathfrak{h},N}\nu^{1+\mathfrak{h}}G_N^{1+\mathfrak{h}}}{1+\mathfrak{h}}\sum_{k=1}^{K}\alpha_k^{1+\mathfrak{h}}.
\end{align*}
\end{proof}

\begin{proof}[Proof of \ref{thm:holder_bcg_small_o}]
From $\sum_{k=1}^{\infty}\alpha_k^{1+\mathfrak{h}}<\infty,\ \sum_{k=1}^{\infty}\beta_{1,k}^2<\infty,\sum_{k=1}^{\infty}\alpha_k(1 - \beta_{2,k})<\infty$, and \ref{thm:holder_bcg_upperbound_beta2}, we have
\begin{align*}
\sum_{k=1}^{\infty}\alpha_{k}\mathbb{E}[\|\nabla f_N(\bm{x}_k)\|^2]
&=\lim_{K\to\infty}\sum_{k=1}^{K}\alpha_{k}\mathbb{E}[\|\nabla f_N(\bm{x}_k)\|^2]
\leq \lim_{K\to\infty}\left(Q_1+Q_2\alpha_{\max}\sum_{k=1}^{K}\beta_{1,k}^2+Q_3\sum_{k=1}^{K}\alpha_{k}(1 - \beta_{2,k})+Q_4\sum_{k=1}^{K}\alpha_k^{1+\mathfrak{h}}\right)\\
&=Q_1+Q_2\alpha_{\max}\sum_{k=1}^{\infty}\beta_{1,k}^2+Q_3\sum_{k=1}^{\infty}\alpha_{k}(1 - \beta_{2,k})+Q_4\sum_{k=1}^{\infty}\alpha_k^{1+\mathfrak{h}}<\infty.
\end{align*}
Similarly, $\sum_{k=1}^{\infty}\alpha_k^{1+\mathfrak{h}}<\infty,\ \sum_{k=1}^{\infty}\beta_{1,k}^2<\infty$, and \ref{thm:holder_bcg_upperbound_beta2} imply $\sum_{k=1}^{\infty}\alpha_{k}\mathbb{E}[\|\nabla f_N(\bm{x}_k)\|^2]<\infty$. Therefore, the assumption $\sum_{k=1}^{\infty}\alpha_k=\infty$ and Lemma 2 in \citet{liu2022smallo} with $a_k\coloneq\alpha_k$ and $x_k\coloneq\mathbb{E}[\|\nabla f_N(\bm{x}_k)\|^2]$ complete the proof.
\end{proof}

\section{Proof of Theorem \ref{thm:bvariance_conv}}\label{appdix:prf_variance_conv}
\begin{proof}[Proof of Theorem \ref{thm:bvariance_conv}]
From \ref{asmmodi:lipsmooth_gradient}, we start with
\begin{align*}
f_N(\bm{x}_{k+1}) 
&\leq f_N(\bm{x}_k) - \alpha_k\left\langle \nabla f_N(\bm{x}_k), H_k^{-1}\hat{\bm{m}}_k\right\rangle + \frac{\alpha_k^{2}L_N}{2}\left\| H_k^{-1}\hat{\bm{m}}_k \right\|^{2} \\
&=f_N(\bm{x}_k) - \alpha_k\underbrace{\left\langle \nabla f_N(\bm{x}_k), H_k^{-1}\nabla f_N(\bm{x}_k)\right\rangle}_{X_1} - \alpha_k\underbrace{\left\langle \nabla f_N(\bm{x}_k), H_k^{-1}(\hat{\bm{m}}_k-\nabla f_N(\bm{x}_k))\right\rangle}_{X_2} + \frac{\alpha_k^{2}L_N}{2}\underbrace{\left\| H_k^{-1}\hat{\bm{m}}_k \right\|^{2}}_{X_3}.
\end{align*}
For $\mathbb{E}[X_1]$ and $\mathbb{E}[X_2]$, we adopt the same estimations with \ref{thm:holder_bcg_upperbound_beta2}. \eqref{eq:m_minus_g_sqr} and Lemma \ref{lem:appdix_boundgradnorm} yield
\begin{align*}
\mathbb{E}[\left\|\hat{\bm{m}}_k -\nabla f_N(\bm{x}_k)\right\|^{2}]
\leq 2\mathbb{E}[\left\|\hat{\bm{m}}_k -\bm{g}_{B_k}(\bm{x}_k)\right\|^{2}]+2\mathbb{E}[\left\|\bm{g}_{B_k}(\bm{x}_k) -\nabla f_N(\bm{x}_k)\right\|^{2}]
\leq 8G_N^2\beta_{1,k}^2+\frac{2G_N^2}{b_k},
\end{align*}
which, together with Proposition \ref{prop:bounds_paramofadam}, leads to
\begin{align*}
\mathbb{E}[X_3]
&=\mathbb{E}[\left\| H_k^{-1}\hat{\bm{m}}_k \right\|^{2}]
\leq \nu^2\mathbb{E}[\left\|\hat{\bm{m}}_k \right\|^{2}]
\leq 2\nu^2\mathbb{E}[\left\|\hat{\bm{m}}_k -\nabla f_N(\bm{x}_k)\right\|^{2}]+2\nu^2\mathbb{E}[\left\|\nabla f_N(\bm{x}_k) \right\|^{2}]\\
&\leq 2\nu^2\left(8G_N^2\beta_{1,k}^2+\frac{2G_N^2}{b_k}\right)+2\nu^2\mathbb{E}[\left\|\nabla f_N(\bm{x}_k) \right\|^{2}]
=16\nu^2G_N^2\beta_{1,k}^2+\frac{4\nu^2G_N^2}{b_k}+2\nu^2\mathbb{E}[\left\|\nabla f_N(\bm{x}_k) \right\|^{2}].
\end{align*}
Substituting the estimations for $X_1, X_2,$ and $X_3$, we obtain
\begin{align*}
0
&\leq \sum_{k=1}^{K}(\mathbb{E}[f_N(\bm{x}_k)]-\mathbb{E}[f_N(\bm{x}_{k+1})]) - \sum_{k=1}^{K}\alpha_k\mathbb{E}[X_1]-\sum_{k=1}^{K}\alpha_k\mathbb{E}[X_2]+\sum_{k=1}^{K}\frac{\alpha_k^{2}L_N}{2}\mathbb{E}[X_3]\\
&\leq f_N(\bm{x}_1)-f_{N,\star} - \mu\sum_{k=1}^{K}\alpha_k\mathbb{E}[\|\nabla f_N(\bm{x}_k)\|^2]+\sum_{k=1}^{K}\alpha_k\left(\frac{\mu}{2}\mathbb{E}[\|\nabla f_N(\bm{x}_k)\|^2]+\frac{4\nu G_N^2}{\mu}\beta_{1,k}^2+\frac{\nu G_N^2}{\mu}\frac{1}{b_k}\right)\\
&\quad +\sum_{k=1}^{K}\frac{\alpha_k^{2}L_N}{2}\left(16\nu^2G_N^2\beta_{1,k}^2+\frac{4\nu^2G_N^2}{b_k}+2\nu^2\mathbb{E}[\left\|\nabla f_N(\bm{x}_k) \right\|^{2}]\right)\\
&\leq-\left(\frac{\mu}{2}-L_N\nu^2 \alpha_{\max}\right)\sum_{k=1}^{K}\alpha_k\mathbb{E}[\|\nabla f_N(\bm{x}_k)\|^2]+f_N(\bm{x}_1)-f_{N,\star}+4\nu G_N^2(\mu^{-1}+2L\nu\alpha_{\max})\sum_{k=1}^{K}\alpha_k\beta_{1,k}^2\\
&\quad +\nu G_N^2(\mu^{-1}+2L\nu\alpha_{\max})\sum_{k=1}^{K}\frac{\alpha_k}{b_k},
\end{align*}
which, together with $\alpha_{\max} < \mu/(2L\nu^2)$, completes the proof.
\end{proof}

\section{Proof of Proposition \ref{prop:bound_grad} and Objective Function in Section \ref{sec:num_heavy}}\label{appdix:prf_boundedgrad_andobj}
\begin{proof}
If $\bm{x}_k$ is an optimal solution, then the gradient norm of the objective function $f$ equals to zero, which leads \ref{asm:bounded_gradient}. Hence, it is enough to prove under the case $\|\nabla f(\bm{x}_k)\|\not=0$. Let $\bar{\bm{x}}_k := \bm{x}_k - \frac{\|\nabla f(\bm{x}_k)\|^{1/\mathfrak{h}}}{L_{\mathfrak{h}}^{1/\mathfrak{h}}}\frac{\nabla f(\bm{x}_k)}{\|\nabla f(\bm{x}_k)\|}$. From \ref{asm:holdersmooth_gradient}, we have
\begin{align*}
f(\bar{\bm{x}}_k)
&\leq f(\bm{x}_k)+\langle\nabla f(\bm{x}_k),\bar{\bm{x}}_k-\bm{x}_k\rangle+\frac{L_{\mathfrak{h}}}{1+\mathfrak{h}}\|\bar{\bm{x}}_k-\bm{x}_k\|^{1+\mathfrak{h}}\\
&\leq f(\bm{x}_k)- \frac{\|\nabla f(\bm{x}_k)\|^{\frac{1}{\mathfrak{h}}}}{L_{\mathfrak{h}}^{\frac{1}{\mathfrak{h}}}}\|\nabla f(\bm{x}_k)\|+\frac{L_{\mathfrak{h}}}{1+\mathfrak{h}}\frac{\|\nabla f(\bm{x}_k)\|^{\frac{1+\mathfrak{h}}{\mathfrak{h}}}}{L^{\frac{1+\mathfrak{h}}{\mathfrak{h}}}}
=f(\bm{x}_k)-\frac{\mathfrak{h}}{(1+\mathfrak{h})L_{\mathfrak{h}}^{\frac{1}{\mathfrak{h}}}}\|\nabla f(\bm{x}_k)\|^{\frac{1+\mathfrak{h}}{\mathfrak{h}}}.
\end{align*}
Thus, \ref{asm:lowerbounded_objective} and $f(\bm{x}_k)\leq f^{\star}$ complete the proof as $\|\nabla f(\bm{x}_k)\|^{\frac{1+\mathfrak{h}}{\mathfrak{h}}}
\leq (1+\mathfrak{h})L_{\mathfrak{h}}^{\frac{1}{\mathfrak{h}}}(f(\bm{x}_k)-f(\bar{\bm{x}}_k))/\mathfrak{h}
\leq (1+\mathfrak{h})L_{\mathfrak{h}}^{\frac{1}{\mathfrak{h}}}(f^{\star}-f_{\star})/\mathfrak{h}$.
\qed
\end{proof}
Next, we check that the objective function in Section \ref{sec:num_heavy} satisfies \ref{asm:lowerbounded_objective}, \ref{asm:holdersmooth_gradient}, \ref{asm:bounded_gradient}, \ref{asm:unbiasedestimator_sg}, and \ref{parprop:hev_exp}. Let $\bm{x}_k=(x_{k,1},\ldots,x_{k,d})^{\top}$. Since $s_{j,k}$ is zero-mean, and $s_{j,k}, u_{j,k}$ are independent samples, the expected risk is reduced to
\begin{align*}
f(\boldsymbol{x}_k)
&= \mathbb{E}_{\bm{\xi}\sim\mathcal{D}}[F(\bm{x}_k;\bm{\xi}_k)]
= \mathbb{E}_{\bm{\xi}_k\sim\mathcal{D}}\left[\sum_{j=1}^{d}
\left\{\int_{0}^{x_{k,j}}\text{sign}(s)\min\left\{|s|^{\mathfrak{h}},1\right\}ds
-\frac{1}{2}\cos(2\pi x_{k,j})+\frac{1}{2}\right\}+\lambda\langle\bm{x}_k,\bm{\xi}_k\rangle\right]\\
&=\sum_{j=1}^{d}
\left\{\int_{0}^{x_{k,j}}\text{sign}(s)\min\left\{|s|^{\mathfrak{h}},1\right\}ds
-\frac{1}{2}\cos(2\pi x_{k,j})+\frac{1}{2}\right\}+\lambda\langle\bm{x}_k,\mathbb{E}_{\bm{\xi}_k\sim\mathcal{D}}[\bm{\xi}_k]\rangle\\
&=\sum_{j=1}^{d}
\left\{\int_{0}^{x_{k,j}}\text{sign}(s)\min\left\{|s|^{\mathfrak{h}},1\right\}ds
-\frac{1}{2}\cos(2\pi x_{k,j})+\frac{1}{2}\right\},
\end{align*}
which leads to $F(\bm{x}_k;\bm{\xi}_k)=f(\bm{x}_k)+\lambda\langle\bm{x}_k,\bm{\xi}_k\rangle$. Hence, \ref{asm:unbiasedestimator_sg} is followed from 
\begin{align*}
\mathbb{E}_{\bm{\xi}_k\sim\mathcal{D}}[\nabla F(\bm{x}_k;\bm{\xi}_k)]
=\mathbb{E}_{\bm{\xi}_k\sim\mathcal{D}}[\nabla f(\bm{x}_k)+\lambda\nabla\langle\bm{x}_k,\bm{\xi}_k\rangle]
=\nabla f(\bm{x}_k)+\lambda\mathbb{E}_{\bm{\xi}_k\sim\mathcal{D}}[\bm{\xi}_k]
=\nabla f(\bm{x}_k).
\end{align*}
Since $\text{sign}(s)\min\left\{|s|^{\mathfrak{h}},1\right\}$ is continuous, the integral is differentiable, and $\partial f(\bm{x})/\partial x_i =\text{sign}(x_i)\min\left\{|x_i|^{\mathfrak{h}},1\right\}+\pi\sin(2\pi x_i)$ holds. Since these partial derivatives are also continuous, $f\in C^1(\mathbb{R}^d)$ holds. \ref{asm:bounded_gradient} is followed from 
\begin{align*}
\|\nabla f(\bm{x}_k)\|^2
&\leq \sum_{j=1}^d\left|\frac{\partial f(\bm{x}_k)}{\partial x_{k,i}}\right|^2 
=\sum_{j=1}^d\left|\text{sign}(x_{k,i})\min\left\{|x_{k,i}|^{\mathfrak{h}},1\right\}+\pi\sin(2\pi x_{k,i})\right|^2\\
&\leq 2\sum_{j=1}^d(|\text{sign}(x_{k,i})\min\left\{|x_{k,i}|^{\mathfrak{h}},1\right\}|^2+|\pi\sin(2\pi x_{k,i})|^2)
\leq 2d(1+\pi^2)
<\infty.
\end{align*}
Next, we confirm \ref{asm:holdersmooth_gradient}; let us start with 
\begin{align*}
\left|\frac{\partial f(\bm{x})}{\partial x_j}-\frac{\partial f(\bm{y})}{\partial y_j}\right|
\leq \left|\text{sign}(x_j)\min\left\{|x_j|^{\mathfrak{h}},1\right\}-\text{sign}(y_j)\min\left\{|y_j|^{\mathfrak{h}},1\right\}\right|+\pi|\sin(2\pi x_j)-\sin(2\pi y_j)|. 
\end{align*}
For $a,b\geq 0$, and $0\leq c\leq 1$, $\min\{a,b\}\leq a^cb^{1-c}$ holds; indeed, we may assume $a\leq b$ without loss of generality due to $0\leq c\leq 1$, and then $a^cb^{1-c}\geq a^ca^{1-c}=a=\min\{a,b\}$. This inequality and $|\sin(2\pi x_j)-\sin(2\pi y_j)|\leq 2$ yields $|\sin(2\pi x_j)-\sin(2\pi y_j)|\leq\min\{2,2\pi|x_j-y_j|\}\leq 2\pi^{\mathfrak{h}}|x_j-y_j|^{\mathfrak{h}}$.
We proceed to estimate the first term in the case of $\text{sign}(x_j)=\text{sign}(y_j)$. Let $l$ be the first term. If $|x_j|^{\mathfrak{h}},|y_j|^{\mathfrak{h}}\geq 1$, then $l=|1-1|=0\leq |x_j-y_j|^{\mathfrak{h}}$; if $|x_j|^{\mathfrak{h}}\geq1$ and $|y_j|^{\mathfrak{h}}< 1$, then $l=|1-|y_j|^{\mathfrak{h}}|=1-|y_j|^{\mathfrak{h}}\leq|x_j|^{\mathfrak{h}}-|y_j|^{\mathfrak{h}}\leq |x_j-y_j|^{\mathfrak{h}}$; if $|x_j|^{\mathfrak{h}}<1$ and $|y_j|^{\mathfrak{h}}\geq 1$, then $l=||x_j|^{\mathfrak{h}}-1|=1-|x_j|^{\mathfrak{h}}\leq |y_j|^{\mathfrak{h}}-|x_j|^{\mathfrak{h}}\leq |x_j-y_j|^{\mathfrak{h}}$; if $|x_j|^{\mathfrak{h}},|y_j|^{\mathfrak{h}}< 1$, then $l=||x_j|^{\mathfrak{h}}-|y_j|^{\mathfrak{h}}|\leq |x_j-y_j|^{\mathfrak{h}}$. Thus, $l\leq |x_j-y_j|^{\mathfrak{h}}$ holds when $\text{sign}(x_j)=\text{sign}(y_j)$. When $\text{sign}(x_j)\neq\text{sign}(y_j)$, we have $l=\min\left\{|x_j|^{\mathfrak{h}},1\right\}+\min\left\{|y_j|^{\mathfrak{h}},1\right\}\leq |x_j|^{\mathfrak{h}}+|y_j|^{\mathfrak{h}}$. Since $f(x)=x^{\mathfrak{h}}\ (x>0)$ is concave due to $f''(x)=-x^{\mathfrak{h}-2}/(\mathfrak{h}(1-\mathfrak{h}))<0$, it follows that $(|x_j|^{\mathfrak{h}}+|y_j|^{\mathfrak{h}})/2\leq \{(|x_j|+|y_j|)/2\}^{\mathfrak{h}}$. Hence, together with $|x_j|+|y_j|=|x_j-y_j|$ in the case of $\text{sign}(x_j)\neq\text{sign}(y_j)$, we obtain $l\leq |x_j|^{\mathfrak{h}}+|y_j|^{\mathfrak{h}}\leq 2^{1-\mathfrak{h}}(|x_j|+|y_j|)^{\mathfrak{h}}=2^{1-\mathfrak{h}}|x_j-y_j|^{\mathfrak{h}}$. Thus, $l\leq (1\vee 2^{1-\mathfrak{h}})|x_j-y_j|^{\mathfrak{h}}$ holds for any $x_j, y_j\in\mathbb{R}^d$. Combining the evaluations, we have $|\frac{\partial f(\bm{x})}{\partial x_j}-\frac{\partial f(\bm{y})}{\partial y_j}|
\leq(1\vee 2^{1-\mathfrak{h}}+2\pi^{\mathfrak{h}})|x_j-y_j|^{\mathfrak{h}}
\leq2(1+\pi)|x_j-y_j|^{\mathfrak{h}}$. Since $t\mapsto t^{\mathfrak{h}}$ is concave on $t\geq 0$ due to $\mathfrak{h}\in(0,1]$, Jensen's inequality yields $d^{-\mathfrak{h}}\|\bm{x}-\bm{y}\|^{2\mathfrak{h}}
=(d^{-1}\|\bm{x}-\bm{y}\|^{2})^{\mathfrak{h}}
= (d^{-1}\sum_{j=1}^d|x_j-y_j|^{2})^{\mathfrak{h}}
\geq d^{-1}\sum_{j=1}^d|x_j-y_j|^{2\mathfrak{h}}$. Thus, \ref{asm:holdersmooth_gradient} is followed from
\begin{align*}
\|\nabla f(\bm{x})-\nabla f(\bm{y})\|^2
=\sum_{j=1}^d\left|\frac{\partial f(\bm{x})}{\partial x_j}-\frac{\partial f(\bm{y})}{\partial y_j}\right|^2
\leq 4(1+\pi)^2\sum_{j=1}^d|x_j-y_j|^{2\mathfrak{h}}
\leq \underbrace{4(1+\pi)^2d^{1-\mathfrak{h}}}_{L_{\mathfrak{h}}^2}\|\bm{x}-\bm{y}\|^{2\mathfrak{h}}.
\end{align*}
We confirm \ref{asm:lowerbounded_objective}; if $x_j\geq 0$, then
\begin{align*}
\int_{0}^{x_j}\text{sign}(s)\min\left\{|s|^{\mathfrak{h}},1\right\}ds
=\int_{0}^{x_j}\min\left\{s^{\mathfrak{h}},1\right\}ds=
\begin{dcases}
\int_0^{x_j}s^{\mathfrak{h}}ds\\
\int_0^1 s^{\mathfrak{h}}ds+\int_1^{x_j}ds
\end{dcases}
=
\begin{dcases}
\frac{x_j^{1+\mathfrak{h}}}{1+\mathfrak{h}}\geq 0&(0\leq x_j\leq 1),\\
x_j-\frac{\mathfrak{h}}{1+\mathfrak{h}}\geq 0&(x_j>1).
\end{dcases}
\end{align*}
Here, let $g$ be the integrand. Even if $x_j<0$, since the integrand is odd, it holds from the above calculation that $\int_0^{x_j}g(s) ds=\int_{x_j}^0-g(s) ds=\int_{x_j}^0g(-s) ds=\int_{-x_j}^0g(s) (-ds)=\int_0^{-x_j}g(s) ds\geq 0$. Therefore, \ref{asm:lowerbounded_objective} is followed from $f(\bm{x})
\geq \sum_{j=1}^d (-2^{-1}\cos(2\pi x_j)+2^{-1})
\geq \sum_{j=1}^d (-2^{-1}+2^{-1})
=0$. Finally, for $\alpha\in(1,2)$, we confirm that the $\mathfrak{p}$-th moment of stochastic gradient is bounded for any $\mathfrak{p}\in(1,\alpha)$, but not for any $\mathfrak{p}\in[\alpha,2)$. Recall $\bm{\xi}_k=(\xi_{k,1},\ldots,\xi_{k,d})^{\top}$ and $\xi_{k,j}\coloneq s_{k,j} u_{k,j}^{-1/\alpha}$, where $s_{k,j}\sim \mathcal{U}(\{\pm1\})$ and $u_{k,j}\sim \mathcal{U}(0,1)$. Then,
\begin{align*}
\mathbb{E}_{\bm{\xi}_k\sim\mathcal{D}}[|\xi_{k,j}|^\mathfrak{p}]
&=\mathbb{E}_{\xi_{k,j}\sim\mathcal{U}(\{\pm1\})\otimes\mathcal{U}(0,1)}\left[\left|s_{k,j} u_{k,j}^{-\frac{1}{\alpha}}\right|^\mathfrak{p}\right]
=\mathbb{E}_{\xi_{k,j}\sim\mathcal{U}(\{\pm1\})\otimes\mathcal{U}(0,1)}\left[u_{k,j}^{-\frac{\mathfrak{p}}{\alpha}}\right]
=\mathbb{E}_{u_{k,j}\sim\mathcal{U}(0,1)}\left[u_{k,j}^{-\frac{\mathfrak{p}}{\alpha}}\right]\\
&=\int_0^1u^{-\frac{\mathfrak{p}}{\alpha}} \frac{1}{1-0}du
=\frac{\alpha}{\alpha-\mathfrak{p}}\left[u^{\frac{\alpha-\mathfrak{p}}{\alpha}}\right]_0^1
=
\begin{dcases}
\frac{\alpha}{\alpha-\mathfrak{p}}<\infty & \text{if}\quad \mathfrak{p}\in(1,\alpha),\\
\infty & \text{if}\quad \mathfrak{p}\in[\alpha,2).
\end{dcases}
\end{align*}
Therefore, the properties are followed from
\begin{align*}
\mathbb{E}_{\bm{\xi}_k\sim\mathcal{D}}[\|\nabla F(\bm{x}_k;\bm{\xi}_k)-\nabla f(\bm{x}_k)\|^{\mathfrak{p}}]
&=\mathbb{E}_{\bm{\xi}_k\sim\mathcal{D}}[\|\lambda\nabla\langle\bm{x},\bm{\xi}_k\rangle \|^{\mathfrak{p}}]
=\lambda^{\mathfrak{p}}\mathbb{E}_{\bm{\xi}_k\sim\mathcal{D}}[\|\bm{\xi}_k\|^{\mathfrak{p}}]
=\lambda^{\mathfrak{p}}\mathbb{E}_{\bm{\xi}_k\sim\mathcal{D}}\left[\left(\sum_{j=1}^d|\xi_{k,j}|^2\right)^{\frac{\mathfrak{p}}{2}}\right]\\
&\leq \lambda^{\mathfrak{p}}\mathbb{E}_{\bm{\xi}_k\sim\mathcal{D}}\left[\sum_{j=1}^d|\xi_{k,j}|^{\mathfrak{p}}\right]
=\frac{d\lambda^{\mathfrak{p}}\alpha}{\alpha-\mathfrak{p}}<\infty,
\end{align*}
and $\mathbb{E}_{\bm{\xi}_k\sim\mathcal{D}}[\|\nabla F(\bm{x}_k;\bm{\xi}_k)-\nabla f(\bm{x}_k)\|^{\mathfrak{p}}]=\lambda^{\mathfrak{p}}\mathbb{E}_{\bm{\xi}_k\sim\mathcal{D}}[\|\bm{\xi}_k\|^{\mathfrak{p}}]
\geq \lambda^{\mathfrak{p}}\mathbb{E}_{\bm{\xi}_k\sim\mathcal{D}}[\|\xi_{k,1}\|^{\mathfrak{p}}]
=\infty$.\end{document}